\documentclass[letterpaper]{article} 
\usepackage{aaai24}  
\usepackage{times}  
\usepackage{helvet}  
\usepackage{courier}  
\usepackage[hyphens]{url}  
\usepackage{graphicx} 
\usepackage{natbib}  
\usepackage{caption} 
\usepackage{amsfonts,amssymb}
\usepackage{amsmath}
\usepackage{multirow}
\usepackage{makecell}
\usepackage{booktabs}
\usepackage{subfigure}

\title{ViTEraser: Harnessing the Power of Vision Transformers for Scene Text Removal with SegMIM Pretraining}
\author{
    Dezhi Peng\textsuperscript{\rm 1,3}, Chongyu Liu\textsuperscript{\rm 1}, Yuliang Liu\textsuperscript{\rm 4}, Lianwen Jin\textsuperscript{\rm 1,2,3,}\thanks{Corresponding author}
}
\affiliations{
    \textsuperscript{\rm 1}South China University of Technology\\
    \textsuperscript{\rm 2}SCUT-Zhuhai Institute of Modern Industrial Innovation\\
    \textsuperscript{\rm 3}INTSIG-SCUT Joint Lab of Document Image Analysis and Recognition\\
    \textsuperscript{\rm 4}Huazhong University of Science and Technology

    pengdzscut@foxmail.com, eelwjin@scut.edu.cn
}

\usepackage{bibentry}

\begin{document}

\maketitle

\begin{abstract}
Scene text removal (STR) aims at replacing text strokes in natural scenes with visually coherent backgrounds.
Recent STR approaches rely on iterative refinements or explicit text masks, resulting in high complexity and sensitivity to the accuracy of text localization.
Moreover, most existing STR methods adopt convolutional architectures while the potential of vision Transformers (ViTs) remains largely unexplored.
In this paper, we propose a simple-yet-effective ViT-based text eraser, dubbed ViTEraser. 
Following a concise encoder-decoder framework, ViTEraser can easily incorporate various ViTs to enhance long-range modeling.
Specifically, the encoder hierarchically maps the input image into the hidden space through ViT blocks and patch embedding layers, while the decoder gradually upsamples the hidden features to the text-erased image with ViT blocks and patch splitting layers.
As ViTEraser implicitly integrates text localization and inpainting, we propose a novel end-to-end pretraining method, termed SegMIM, which focuses the encoder and decoder on the text box segmentation and masked image modeling tasks, respectively. 
Experimental results demonstrate that ViTEraser with SegMIM achieves state-of-the-art performance on STR by a substantial margin and exhibits strong generalization ability when extended to other tasks, \textit{e.g.}, tampered scene text detection.
Furthermore, we comprehensively explore the architecture, pretraining, and scalability of the ViT-based encoder-decoder for STR, which provides deep insights into the application of ViT to the STR field.
Code is available at https://github.com/shannanyinxiang/ViTEraser.
\end{abstract}

\section{Introduction}

Scene text removal (STR) aims to realistically erase the text strokes in the wild by replacing them with visually plausible background, which has been widely applied to privacy protection \cite{inai2014selective}, image editing \cite{wu2019editing}, and image retrieval \cite{tursun2019component}.
Existing approaches to STR have evolved from the one-stage paradigm which implicitly integrates the text localization and background inpainting into a single network without the guidance of explicit text masks \cite{nakamura2017scene,zhang2019ensnet,liu2020erasenet}, to the two-stage framework which contains explicit text localizing processes and uses the resulting text masks to facilitate background inpainting \cite{tang2021stroke,lee2022surprisingly,wang2023pert,du2022progressive}.

\begin{figure}
    \centering
    \includegraphics[width=1\columnwidth]{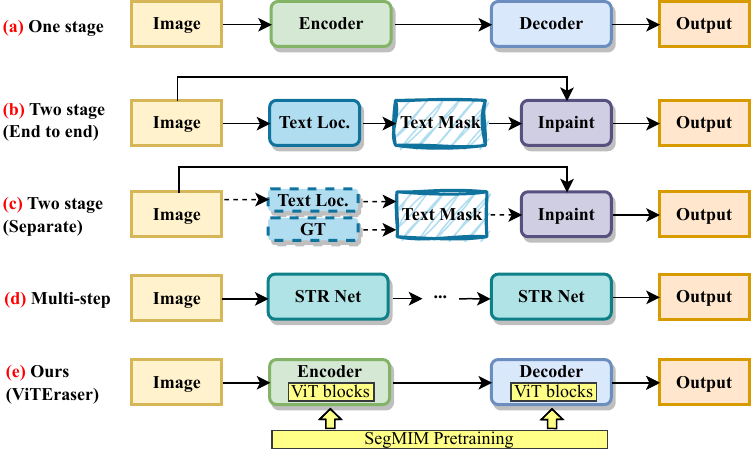}
    \caption{Comparison of ViTEraser with existing STR paradigms. Our method revisits the conventional single-step one-stage framework and improves it with ViTs for feature modeling and the proposed SegMIM pretraining. Dashed arrows indicate cutting off gradient flow. (Loc.: Localization)}
    \label{fig:intro}
\end{figure}
 
Despite the great success achieved by previous methods, there still remain two critical issues. 
\textbf{(1)}
The dominant two-stage methods suffer from the complex system design with two sub-tasks. 
The sequential text localizing and background inpainting pipeline introduces additional parameters, decreases the inference speed, and, more importantly, breaks the integrity of the entire model.
The error of text localization can be easily propagated to the background inpainting, especially for the methods that require pre-supplied text detectors \cite{tang2021stroke,liu2022don,lee2022surprisingly}.
\textbf{(2)}
Recent advances \cite{liu2020erasenet,lyu2022psstrnet,du2022progressive,wang2023pert} tend to employ a multi-step paradigm in a coarse-to-fine or progressive fashion, which significantly undermines efficiency and makes it difficult to balance the parameters involved in multiple steps. 
 
To this end, we revisit the one-stage paradigm and propose a novel simple-yet-effective ViT-based method for STR, termed as ViTEraser. Fig.~\ref{fig:intro} compares our method with existing STR approaches.
The ViTEraser follows the conventional one-stage framework which comprises a single-step encoder-decoder network and is free of text mask input or text localizing processes.
This concise pipeline perfectly gets rid of the drawbacks of the two-stage and multi-step approaches mentioned above, but has been discarded in recent advances because of the unexpected artifacts and inexhaustive erasure issues caused by the implicit text localization mechanism.
However, we argue that these limitations are actually due to the insufficient capacity of previous CNN-based architectures.
Recently, vision Transformers (ViTs) \cite{dosovitskiyimage} have achieved incredible success on diverse computer vision tasks \cite{han2022survey} but are rarely investigated for STR.
Nevertheless, ViT is perfect for STR since global information is indispensable for determining text locations and background textures, especially for large texts that existing STR systems still struggle with. 
Therefore, for the first time in the STR field, the proposed ViTEraser thoroughly utilizes ViTs for feature representation in both the encoder and decoder. 
Specifically, the encoder hierarchically maps the input image into the hidden space through ViT blocks and patch embedding layers, while the decoder gradually upsamples the hidden features to the text-erased image with ViT blocks and patch splitting layers.
Thanks to its high generality, ViTEraser can be effortlessly integrated with various ViTs, \textit{e.g.}, Swin Transformer (v2) \cite{liu2021swin,liu2022swin}, PVT \cite{wang2021pyramid,wang2022pvt}.

Despite the powerful ViT-based structure, the implicit integration of text localizing and background inpainting still significantly challenges the model capacity of ViTEraser, requiring both high-level text perception and fine-grained pixel infilling abilities.
However, the insufficient scale of existing STR datasets \cite{liu2020erasenet} limits the full learning of these abilities and makes the large-capacity ViT-based model prone to overfitting.
To solve similar issues, pretraining plays a crucial role in a variety of fields \cite{kenton2019bert,xu2020layoutlm,yang2022reading} but is quite under-explored in the STR realm.
Moreover, with the rapid development of large-scale scene text detection datasets and commercial optical character recognition (OCR) APIs, numerous real-world images with text bounding boxes are easily available.
Therefore, we propose SegMIM which fully pretrains STR models using large-scale scene text detection data.
Concretely, by assigning two pretraining tasks of text box segmentation and mask image modeling (MIM) \cite{he2022masked,xie2022simmim} to the output features of the encoder and decoder, respectively, the STR performance can be effectively boosted with enhanced text localizing, inpainting, and global reasoning abilities. 

Extensive experiments are conducted on two STR benchmarks including SCUT-EnsText \cite{liu2020erasenet} and SCUT-Syn \cite{zhang2019ensnet}. 
Furthermore, we comprehensively explore the architecture, pretraining, and scalability of the ViT-based encoder-decoder for STR.
The experimental results demonstrate the clear superiority of ViTEraser with and without the SegMIM pretraining.
Additionally, ViTEraser also achieves state-of-the-art performance on tampered scene text detection using the Tampered-IC13 \cite{wang2022detecting} dataset, exhibiting strong generalization ability.

In summary, the contributions of this paper are as follows.
\begin{itemize}
    \setlength{\itemsep}{1pt}
    \setlength{\parsep}{0pt}    
    \setlength{\parskip}{0pt}
    \item We propose a novel ViT-based one-stage method for STR, termed as ViTEraser. The ViTEraser adopts a concise single-step encoder-decoder paradigm, thoroughly integrating ViTs for feature representation in both the encoder and decoder.
    \item We propose SegMIM, a new pretraining scheme for STR. 
    With SegMIM, ViTEraser acquires enhanced global reasoning capabilities, enabling it to effectively distinguish and generate text and background regions.
    \item We conduct a comprehensive investigation into the architecture, pretraining, and scalability of the ViT-based encoder-decoder for STR, which provides deep insights into the application of ViT to the STR field.
    \item The experiments demonstrate that ViTEraser achieves state-of-the-art performance on STR, and its potential for extension to other domains is also highlighted.
\end{itemize}

\section{Related Work}
\subsection{Scene Text Removal}
Scene text removal aims at realistically erasing the texts in natural scenes.
Existing methods can be divided into one-stage and two-stage categories based on whether there are explicit text localizing processes.

\textbf{One-stage methods} follow a concise image-to-image translation pipeline, implicitly integrating text localizing and background inpainting procedures into a single network. 
\citet{nakamura2017scene} pioneered in erasing texts at patch level using a convolution-to-deconvolution encoder-decoder structure.
Inspired by Pix2Pix \cite{isola2017image}, \citet{zhang2019ensnet} proposed an end-to-end cGAN-based \cite{mirza2014conditional} EnsNet which directly erases texts at image level.
EraseNet \cite{liu2020erasenet} further improved EnsNet following a coarse-to-refine pipeline.
From a data perspective, \citet{jiang2022self} proposed a controllable synthesis module based on EraseNet.

\textbf{Two-stage methods} decompose STR into the text localizing and background inpainting processes. 
The text localizing component produces explicit text masks which are fed into subsequent modules to facilitate background inpainting. 
The two-stage methods can be further divided into separate and end-to-end categories. 
The \textbf{separate two-stage methods} depend on separately trained text detectors \cite{zdenek2020erasing,conrad2021two,liu2022don} or ground truth (GT) \cite{qinautomatic,tursun2019mtrnet,tang2021stroke,lee2022surprisingly} to obtain text masks.
In contrast, \textbf{end-to-end two-stage methods} end-to-end optimize the text localizing modules with other components \cite{keserwani2021text}.
Under this paradigm, recent advances tended to devise coarse-to-refine \cite{tursun2020mtrnet++,du2023modeling} or progressive frameworks \cite{lyu2022psstrnet,bian2022scene,du2022progressive,wang2023pert} with text segmentation modules.
On the contrary, \citet{hou2022multi} expanded the width of the network in a multi-branch fashion.
Additionally, \citet{lyu2023fetnet} incorporated text segmentation maps at feature level using the proposed FET mechanism.
Although the two-stage methods have dominated the STR field, they suffer from the high complexity caused by multiple modules and progressive erasing and are prone to text localizing accuracy. 

\begin{figure*}[t]
    \centering
    \includegraphics[width=0.95\linewidth]{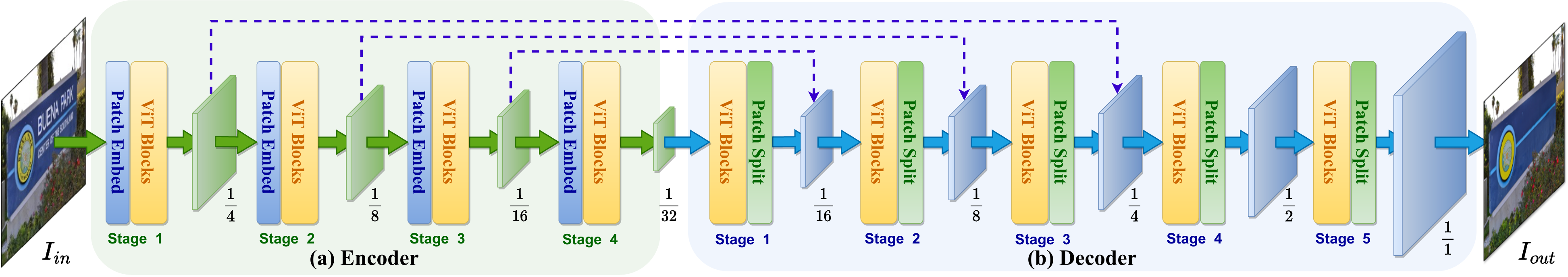}
    \caption{Overall architecture of ViTEraser. 
    The ViTEraser follows the one-stage paradigm but is thoroughly equipped with ViTs, yielding a simple-yet-effective STR approach that is free of progressive refinements and text localizing processes.}
    \label{fig:viteraser}
\end{figure*}

\subsection{Vision Transformer}
The Transformer \cite{vaswani2017attention} was first proposed for natural language processing \cite{kenton2019bert} but rapidly swept the computer vision field \cite{han2022survey}.
Early ViTs \cite{dosovitskiyimage,touvron2021training} first tokenized images with large window sizes and then kept the feature size throughout all Transformer layers. 
Recently, the research on ViTs has focused on producing pyramid feature maps, \textit{e.g.}, PVT \cite{wang2021pyramid,wang2022pvt}, HVT \cite{pan2021scalable}, Swin Transformer \cite{liu2021swin,liu2022swin}, and PiT \cite{heo2021rethinking}.
Nowadays, ViTs have played an important role in many tasks, such as object detection \cite{carion2020end}, semantic segmentation \cite{xie2021segformer,cao2022swin}, text spotting~\cite{peng2022spts,liu2023sptsv2}, and document understanding \cite{xu2020layoutlm}.

\section{ViTEraser}
As shown in Fig.~\ref{fig:viteraser}, we revisit the conventional single-step one-stage paradigm, getting rid of the complicated iterative refinement and the susceptibility to text localizing accuracy.
The proposed ViTEraser pioneers in thoroughly employing ViTs instead of CNN in both the encoder and decoder, yielding a simple-yet-effective pipeline.
Concretely, the encoder hierarchically maps the input image into the hidden space through successive ViT blocks and patch embedding layers, while the decoder gradually upsamples the hidden features to the text-erased image with successive ViT blocks and patch splitting layers.
ViT blocks throughout the encoder-decoder provide sufficient global context information, enabling the implicit integration of text localization and background inpainting into a single network within a single forward pass.
Moreover, lateral connections are devised between the encoder and decoder to preserve the input details.

\subsection{Encoder}
\label{sec:viteraser_encoder}

As shown in Fig. \ref{fig:viteraser}(a), the encoder of ViTEraser consists of four stages. Given an input image $I_{in} \in \mathbb{R}^{H \times W \times 3}$, the encoder hierarchically produces four feature maps $\{f_{i}^{enc}\}_{i=1}^{4}$ with strides of $\{2^{i+1}\}_{i=1}^4$ \textit{w.r.t} the input image and channels of $\{C_{i}^{enc}\}_{i=1}^{4}$, respectively. 
Specifically, the i\textit{-th} stage first downsamples the spatial size using a patch embedding layer and then captures global correlation through a stack of $N^{enc}_i$ ViT blocks.

\subsubsection{Patch Embedding Layer}
\label{sec:viteraser_encoder_patchembed}
Given an input feature map $f_{in} \in \mathbb{R}^{h \times w \times c_{in}}$, a patch embedding layer with a downsampling ratio of $d$ and an output channel of $c_{out}$ first flattens each $d\times d$ patch, yielding a $\frac{h}{d} \times \frac{w}{d} \times (d^2\times c_{in})$ feature map.
Then a $1 \times 1$ convolution layer is applied to transform this intermediate feature map into the output $f_{out} \in \mathbb{R}^{\frac{h}{d} \times \frac{w}{d} \times c_{out}}$.

\subsection{Decoder}
\label{sec:viteraser_decoder}
The decoder contains five stages as illustrated in Fig.~\ref{fig:viteraser}(b).
Based on the final feature $f_4^{enc}$ of the encoder, the decoder hierarchically generates five feature maps $\{f_{i}^{dec} \in \mathbb{R}^{\frac{H}{2^{5-i}} \times \frac{W}{2^{5-i}} \times C_{i}^{dec}}\}_{i=1}^5$. 
Concretely, in each stage, the feature is first processed with $N^{dec}_i$ ViT blocks and then upsampled by 2 via a patch splitting layer.
Moreover, lateral connections \cite{liu2020erasenet} are built between the features $\{f_{i}^{enc}\}_{i=1}^{3}$ of the encoder and the features $\{f_{4-i}^{dec}\}_{i=1}^{3}$ of the decoder.
Finally, the text-erased image is predicted via a $3 \times 3$ convolution based on the feature $f_{5}^{dec} \in \mathbb{R}^{H \times W \times C_{5}^{dec}}$.

\subsubsection{Patch Splitting Layer}
Patch splitting is designed as the inverse operation of the patched embedding to upsample the spatial size of features.
Fed with an input feature map $f_{in} \in \mathbb{R}^{h \times w \times c_{in}}$, the patch splitting layer first decomposes each $c_{in}$-dimensional token into a $2 \times 2$ patch with $\frac{c_{in}}{4}$ dimension, expanding the input feature map to a shape of $2h \times 2w \times \frac{c_{in}}{4}$. 
After that, a $1 \times 1$ convolution layer is adopted to produce the output feature map $f_{out} \in \mathbb{R}^{2h \times 2w \times c_{out}}$.

\begin{figure}[t]
    \centering 
    \includegraphics[width=1\columnwidth]{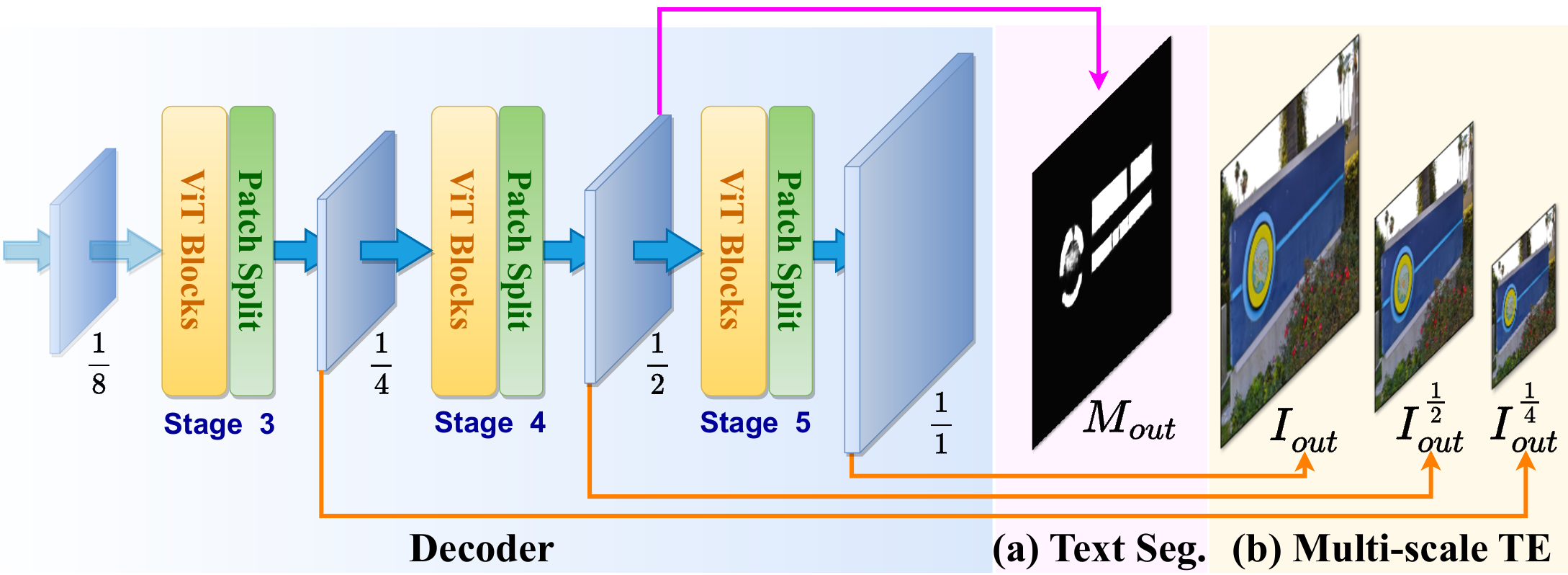}
    \caption{Auxiliary outputs of ViTEraser during training, including (a) text box segmentation map and (b) multi-scale text erasing results. (TE: text erasing)}
    \label{fig:aux_output}
\end{figure}

\begin{figure*}[t]
    \centering
    \includegraphics[width=0.95\linewidth]{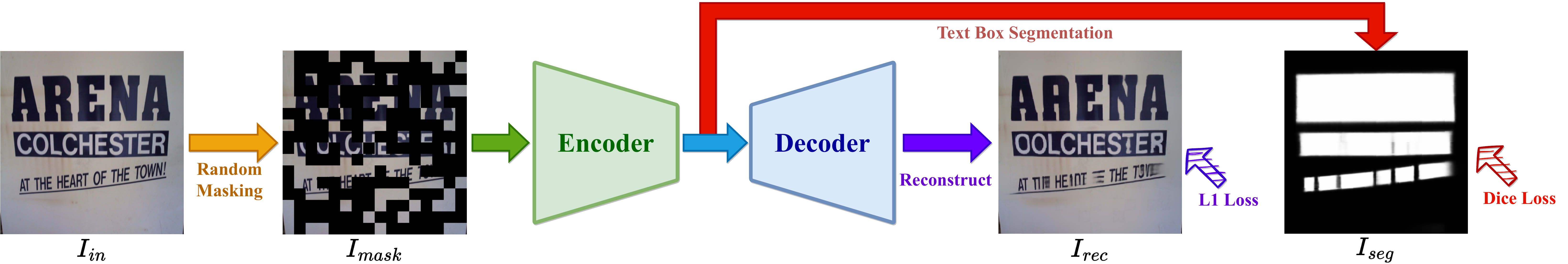}
    \caption{Pipeline of the proposed SegMIM pretraining.
    Given a randomly masked image, the text box segmentation and masked image modeling tasks are accomplished on top of the encoder and decoder, respectively.}
    \label{fig:segmim}
\end{figure*}

\subsection{Training}
\label{sec:viteraser_training}
As depicted in Fig.~\ref{fig:aux_output}, auxiliary outputs are produced during only training, including a text box segmentation map $M_{out}$ and multi-scale text erasing results $\{I^{\frac{1}{2}}_{out}, I^{\frac{1}{4}}_{out}\}$.
Specifically, $M_{out}$ is predicted based on $f_4^{dec}$ via a $3 \times 3$ deconvolution for $2\times$ upsampling and a $3 \times 3$ convolution with Sigmoid activation. 
Besides, $I^{\frac{1}{2}}_{out} \in \mathbb{R}^{\frac{H}{2} \times \frac{W}{2} \times 3}$ and $I^{\frac{1}{4}}_{out} \in \mathbb{R}^{\frac{H}{4} \times \frac{W}{4} \times 3}$ are generated from features $f_4^{dec}$ and $f_3^{dec}$, respectively, each through a $3 \times 3$ convolution.

The model training adopts a GAN-based paradigm with a local-global discriminator \cite{liu2020erasenet}.
Given an input image with corresponding text-erased image and groud-truth (GT) text box mask, the losses comprise multi-scale reconstruction loss, perceptual loss, style loss, segmentation loss, and adversarial loss, following EraseNet \cite{liu2020erasenet}.
\textit{See appendix for training details.}

\section{SegMIM Pretraining}
Unlike two-stage methods that utilize task-specific modules and training objectives, ViTEraser implicitly integrates text localizing and background inpainting tasks into a single encoder-decoder, thus facing challenges in fully learning to handle both tasks and susceptible to overfitting when scaled up. 
This limitation arises from the scarcity of training samples in STR and the high costs associated with annotating them. 
Moreover, enormous natural scene images with text bounding boxes are easily available as the advancing of scene text detection datasets and OCR APIs.
To this end, leveraging the availability of extensive scene text detection datasets, we exploit large-scale pretraining techniques which have recently shown significant advancements \cite{yang2022reading,xu2020layoutlm} but rarely been investigated for STR.

Because STR can be decomposed into text localizing and background inpainting sub-tasks, we intuitively propose a new pretraining paradigm for STR, termed SegMIM, which focuses the encoder on text box segmentation and the decoder on masked image modeling (MIM) as shown in Fig.~\ref{fig:segmim}.
Despite its simplicity, the clear advantages and interpretability of SegMIM are manifold.
\textbf{(1)}
The model learns the discriminative representation of texts and backgrounds through the text box segmentation task, which is crucial to STR. 
\textbf{(2)} 
The model learns the generative features of texts and backgrounds via MIM, enhancing the text perception and background recovery.
\textbf{(3)} 
The global reasoning capacity is significantly improved due to the high mask ratio (0.6).

\subsection{Architecture}
The network architecture during pretraining inherits the encoder-decoder structure as shown in Fig.~\ref{fig:viteraser} but adds two extra heads for text box segmentation and image reconstruction, respectively.
Given an input image $I_{in} \in \mathbb{R}^{H \times W \times 3}$, a binary mask $M_{mim} \in \mathbb{R}^{H \times W \times 1}$ is randomly generated following SimMIM \cite{xie2022simmim}.
Then, the masked image $I_{mask}$ combining $I_{in}$ and $M_{mim}$ is fed into the network.

\subsubsection{Text Box Segmentation Head}
Based on the final feature $f_{4}^{enc} \in \mathbb{R}^{\frac{H}{32} \times \frac{W}{32} \times C_{4}^{enc}}$ of the encoder, a $1 \times 1$ convolution layer changes its dimension from $C_{4}^{enc}$ to $1024$.
Subsequently, after transforming each $1024$-dimension vector to a $32 \times 32$ patch and activating using a sigmoid function, a text box segmentation map $I_{seg} \in \mathbb{R}^{H \times W \times 1}$ is obtained.

\subsubsection{Image Reconstruction Head}
Through a $3 \times 3$ convolution, a reconstructed image $I_{rec} \in \mathbb{R}^{H \times W \times 3}$ is predicted using the final feature $f_5^{dec} \in \mathbb{R}^{H \times W \times C_{5}^{dec}}$ of the decoder.

\subsection{Optimization}
The loss $\mathcal{L}_{pre}$ for pretraining is the sum of a text box segmentation loss $\mathcal{L}_{dice}$ and a MIM loss $\mathcal{L}_{mim}$ as follows.
\begin{align}
    \mathcal{L}_{pre}  = & \mathcal{L}_{dice} + \mathcal{L}_{mim}, \\
    \mathcal{L}_{dice} = & 1 - \frac{2\sum_{i,j}I_{seg(i,j)} \times S_{gt(i,j)}}{\sum_{i,j}(I_{seg(i,j)})^2 + \sum_{i,j}(S_{gt(i,j)})^2}, \\
    \mathcal{L}_{mim}  = &|| \Psi(I_{rec}, M_{mim}) - \Psi(I_{in}, M_{mim}) ||_1, 
\end{align}
where $S_{gt} \in \mathbb{R}^{H \times W \times 1}$ is the GT text box mask and the function $\Psi$ fetches the image pixels at masked positions.

\section{Experiments}
\subsection{Datasets}
\textbf{Scene Text Removal Datasets} include \textbf{SCUT-EnsText} \cite{liu2020erasenet} and \textbf{SCUT-Syn} \cite{zhang2019ensnet}. 
Specifically, SCUT-EnsText is a real-world dataset containing 2,749 samples for training and 813 samples for testing.
SCUT-Syn is a synthetic dataset with 8,000 and 800 samples for training and testing, respectively.

\noindent\textbf{Pretraining Datasets} include the training sets of \textbf{ICDAR2013} \cite{karatzas2013icdar}, \textbf{ICDAR2015} \cite{karatzas2015icdar}, \textbf{MLT2017} \cite{nayef2017icdar2017}, \textbf{ArT} \cite{chng2019icdar2019}, \textbf{LSVT} \cite{sun2019icdar}, and \textbf{ReCTS} \cite{zhang2019icdar}, as well as the training and validating sets of \textbf{TextOCR} \cite{singh2021textocr}. 
After removing the overlapping samples with the test set of SCUT-EnsText \cite{liu2020erasenet}, there are totally 88,340 valid samples for pretraining.

\begin{table*}[t]
    \centering
    \resizebox{1.5\columnwidth}{!}{
        \begin{tabular}{llcccccccc}
        \toprule
        \multirow{2}*{Encoder} & \multirow{2}*{Decoder} & \multicolumn{7}{c}{SCUT-EnsText} & \multirow{2}*{\thead{Params$\downarrow$ \\ (M)}} \\
        \cmidrule{3-9}
        & & PSNR$\uparrow$ & MSSIM $\uparrow$ & MSE $\downarrow$ & AGE$\downarrow$ & pEPs$\downarrow$ & pCEPs$\downarrow$ & FID$\downarrow$ & \\
        \midrule
        Conv & Deconv & 35.05 & 97.20 & 0.0893 & 2.14 & 0.0111 & 0.0069 & 13.98 & 131.45 \\ 
        \midrule
        Conv+TE & Deconv & 34.85 & 97.13 & 0.1043 & 2.22 & 0.0120 & 0.0076 & 14.20 & 133.04 \\ 
        Conv+TE & TD+Deconv & 34.89 & 97.14 & 0.1007 & 2.20 & 0.0118 & 0.0074 & 14.12 & 139.43 \\ 
        \midrule
        Swinv2-Tiny & Deconv & \underline{36.06} & \underline{97.40} & \underline{0.0573} & \underline{1.88} & \underline{0.0079} & \underline{0.0043} & \underline{12.17} & 65.83 \\ 
        Swinv2-Tiny & TD+Deconv & 35.92 & 97.39 & 0.0591 & 1.89 & 0.0082 & 0.0046 & 12.52 & 71.37 \\ 
        Swinv2-Tiny & MLP & 26.18 & 81.07 & 0.3532 & 7.32 & 0.0852 & 0.0157 & 38.90 & \textbf{28.21} \\ 
        \midrule
        \multicolumn{2}{c}{ViTEraser-Swinv2-Tiny} & \textbf{36.32} & \textbf{97.48} & \textbf{0.0569} & \textbf{1.81} & \textbf{0.0073} & \textbf{0.0040} & \textbf{11.77} & \underline{65.39} \\ 
        \bottomrule        
    \end{tabular}
    }
    \caption{Comparison of different Transformer-based STR architectures.}
    \label{tab:exp_arch}
\end{table*}

\subsection{Implementation Details} 
\subsubsection{Network Architecture}
We explore three types of ViT blocks, \textit{i.e.}, Pyramid Vision Transformer block (\textbf{PVT}), Swin Transformer block (\textbf{Swin}), and Swin Transformer v2 block (\textbf{Swinv2}), to implement the proposed ViTEraser.
Based on the original scale settings of these ViTs \cite{wang2021pyramid,liu2021swin,liu2022swin}, we obtain four scales of PVT-based ViTEraser (\textbf{ViTEraser-PVT-Tiny/Small/Medium/Large}), three scales of Swin-based ViTEraser (\textbf{ViTEraser-Swin-Tiny/Small/Base}), and three scales of Swinv2-based ViTEraser (\textbf{ViTEraser-Swinv2-Tiny/Small/Base}).
For conciseness, \textbf{ViTEraser refers to the Swinv2-based ViTEraser by default}.
\textit{See appendix for detailed network architectures.}

\subsubsection{Pretraining}
The input image is resized to $512 \times 512$.
Random masking is performed on the input image with a ratio of 0.6 and a patch size of 32.
Besides, a mask token is added to the encoder to represent the masked patches.
Using 4 NVIDIA A6000 GPUs with 48GB memory, the network is pretrained for 100 epochs with an AdamW optimizer, a batch size of 64, and learning rates of 0.0001 before the 80\textit{th} epoch and 0.00001 afterward.
Because the mask token can negatively affect the encoder, the encoder will be finetuned solely with the text box segmentation task after end-to-end pretraining, following the training strategy of SimMIM.
The finetuning lasts for 20 epochs with an initial learning rate of 0.00125 and a cosine decay learning rate schedule.

\subsubsection{Training}
The training procedure on SCUT-EnsText or SCUT-Syn uses only its corresponding training set.
The input size of the images is set to $512 \times 512$.
The network is trained with an AdamW optimizer for 300 epochs using 2 NVIDIA A6000 GPUs with 48GB memory.
The learning rate is initialized as 0.0005 and linearly decayed to 0.00001 at the last epoch.
The training batch size is set to 16.

\subsection{Evaluation Metrics}
Following previous studies \cite{liu2020erasenet,liu2022don}, the image-eval metrics include PSNR, MSSIM, MSE, AGE, pEPs, pCEPs, and FID, while the detection-eval metrics involve the precision (P), recall (R), and f-measure (F) using the pretrained CRAFT \cite{baek2019character} for text detection. 

\subsection{Experiments on Architecture}
\textit{Which architecture is the best for the integration of Transformer into STR models?}
To answer this question, we first introduce the encoders and decoders compared in Tab.~\ref{tab:exp_arch}. 

\noindent\textbf{Encoder}
\textbf{(1)} 
\textit{Conv} represents a ResNet50 \cite{he2016deep}.
\textbf{(2)} 
\textit{Conv+TE} indicates the concatenation of a ResNet50 and a 6-layer Transformer encoder with 256 channels.
\textbf{(3)} 
\textit{Swinv2-Tiny} is the tiny version of Swin Transformer v2 \cite{liu2022swin}. 
The ResNet50 and Swinv2-Tiny are pretrained using ImageNet-1k \cite{deng2009imagenet}.

\begin{table*}
    \centering 
    \normalsize
    \resizebox{1.5\columnwidth}{!}{
        \begin{tabular}{lllccccccc}
        \toprule 
        \multirow{2}*{Dataset} & \multirow{2}*{Encoder} & \multirow{2}*{Decoder} & \multicolumn{7}{c}{SCUT-EnsText} \\
        \cmidrule{4-10}
        & & & PSNR$\uparrow$ & MSSIM$\uparrow$ & MSE$\downarrow$ & AGE$\downarrow$ & pEPs$\downarrow$ & pCEPs$\downarrow$ & FID$\downarrow$ \\
        \midrule 
        × & × & × & 33.34 & 96.70 & 0.1854 & 2.52 & 0.0161 & 0.0109 & 18.38 \\ 
        \midrule    
        \multirow{4}*{ImageNet-1k} & CLS & × & 36.55 & 97.56 & 0.0497 & \underline{1.73} & 0.0072 & 0.0039 & 11.46 \\ 
        & SimMIM & × & 36.38 & 97.51 & 0.0622 & 1.79 & 0.0076 & 0.0042 & 11.92 \\ 
        & CLS & CLS & 36.54 & 97.55 & 0.0517 & \underline{1.73} & 0.0073 & 0.0039 & 11.48 \\ 
        & CLS & SimMIM & 36.45 & 97.55 & 0.0508 & 1.75 & 0.0082 & 0.0039 & 11.57 \\ 
        \midrule 
        \multirow{5}*{\thead{Scene Text \\Detection Dataset}} & Text Seg. & × & \underline{36.89} & \underline{97.59} & 0.0490 & \underline{1.73} & 0.0070 & 0.0038 & 10.98 \\ 
        & SimMIM & × & 36.43 & 97.49 & 0.0554 & 1.77 & 0.0074 & 0.0040 & 11.82 \\ 
        & \multicolumn{2}{c}{Text Seg.} & 36.78 & 97.57 & 0.0487 & 1.75 & 0.0070 & 0.0039 & \underline{10.81} \\ 
        & \multicolumn{2}{c}{SimMIM} & 36.68 & 97.58 & \underline{0.0480} & 1.77 & \underline{0.0067} & \underline{0.0036} & 11.09 \\ 
        \cmidrule{2-10} 
        & \multicolumn{2}{c}{SegMIM} & \textbf{37.08} & \textbf{97.62} & \textbf{0.0447} & \textbf{1.69} & \textbf{0.0064} & \textbf{0.0034} & \textbf{10.16} \\ 
        \bottomrule 
    \end{tabular}}
    \caption{Comparison of different pretraining strategies of ViTEraser-Swinv2-Small. (CLS: Classification)}
    \label{tab:exp_pretrain}
\end{table*}

\begin{figure}[t]
    \centering
    \subfigtopskip=2pt 
    \subfigbottomskip=2pt 
    \subfigcapskip=1pt 
    \renewcommand{\subcapsize}{\scriptsize}
    \subfigure{
        \begin{minipage}[c]{0.18\columnwidth}
            \centering
            \includegraphics[width=0.95\columnwidth]{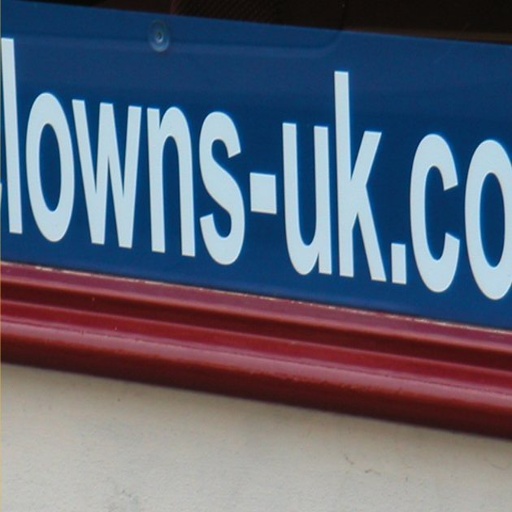}
        \end{minipage}%
    }%
    \subfigure{
        \begin{minipage}[c]{0.18\columnwidth}
            \centering
            \includegraphics[width=0.95\columnwidth]{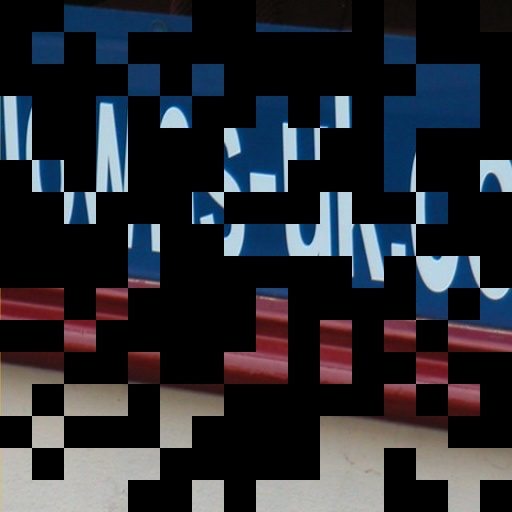}
        \end{minipage}%
    }%
    \subfigure{
        \begin{minipage}[c]{0.18\columnwidth}
            \centering
            \includegraphics[width=0.95\columnwidth]{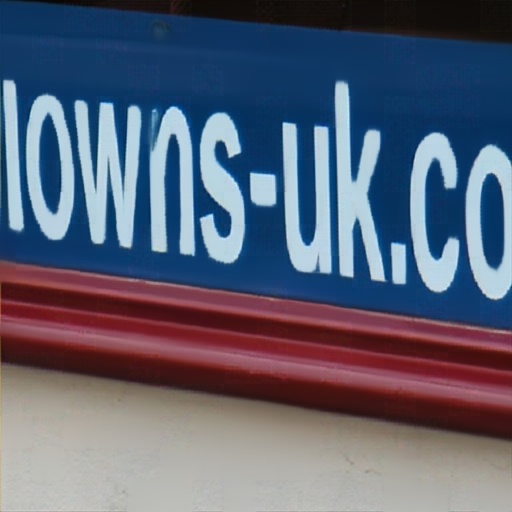}
        \end{minipage}%
    }%
    \subfigure{
        \begin{minipage}[c]{0.18\columnwidth}
            \centering
            \includegraphics[width=0.95\columnwidth]{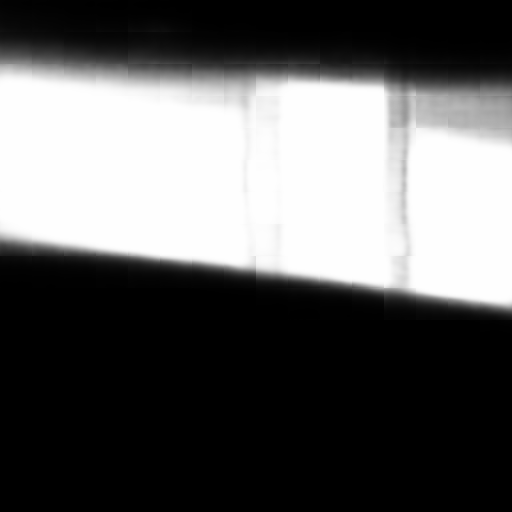}
        \end{minipage}%
    }%
    \subfigure{
        \begin{minipage}[c]{0.18\columnwidth}
            \centering
            \includegraphics[width=0.95\columnwidth]{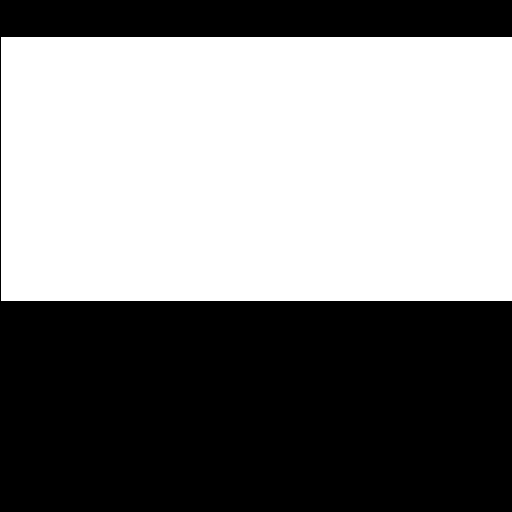}
        \end{minipage}%
    }
    \subfigure{
        \begin{minipage}[c]{0.18\columnwidth}
            \centering
            \includegraphics[width=0.95\columnwidth]{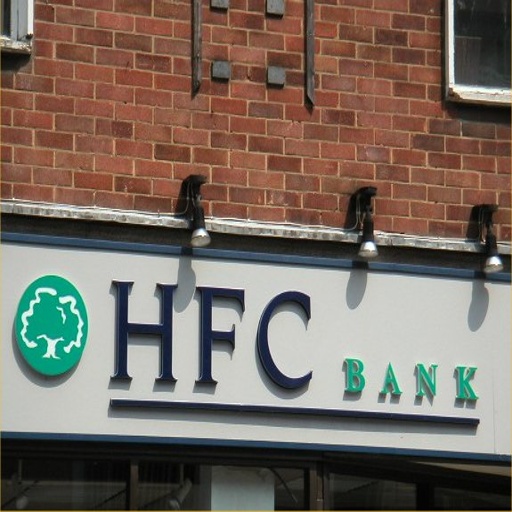}
        \end{minipage}%
    }%
    \subfigure{
        \begin{minipage}[c]{0.18\columnwidth}
            \centering
            \includegraphics[width=0.95\columnwidth]{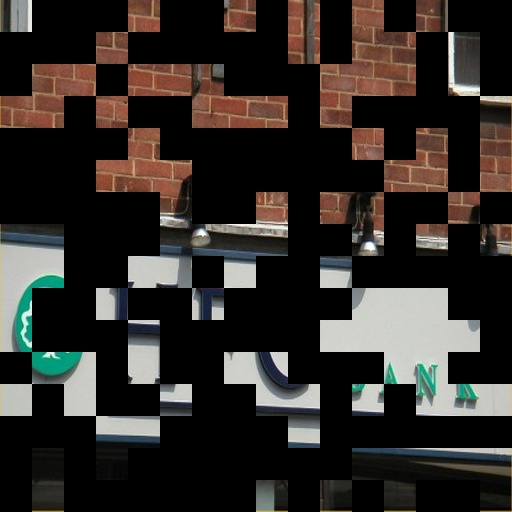}
        \end{minipage}%
    }%
    \subfigure{
        \begin{minipage}[c]{0.18\columnwidth}
            \centering
            \includegraphics[width=0.95\columnwidth]{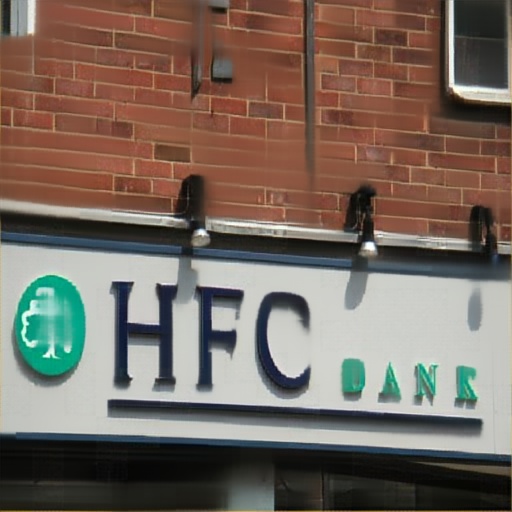}
        \end{minipage}%
    }%
    \subfigure{
        \begin{minipage}[c]{0.18\columnwidth}
            \centering
            \includegraphics[width=0.95\columnwidth]{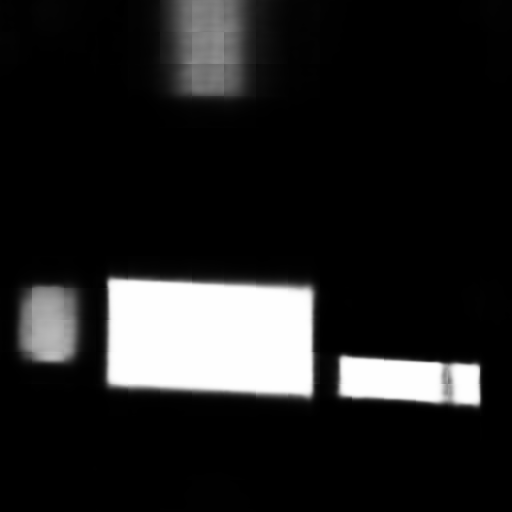}
        \end{minipage}%
    }%
    \subfigure{
        \begin{minipage}[c]{0.18\columnwidth}
            \centering
            \includegraphics[width=0.95\columnwidth]{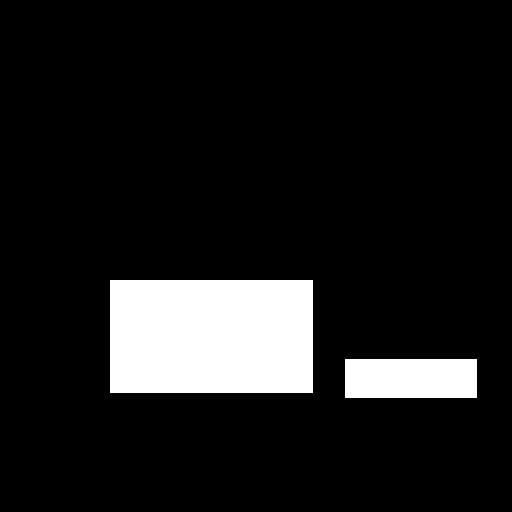}
        \end{minipage}%
    }
    \setcounter{subfigure}{0}
    \subfigure[Input]{
        \begin{minipage}[c]{0.18\columnwidth}
            \centering
            \includegraphics[width=0.95\columnwidth]{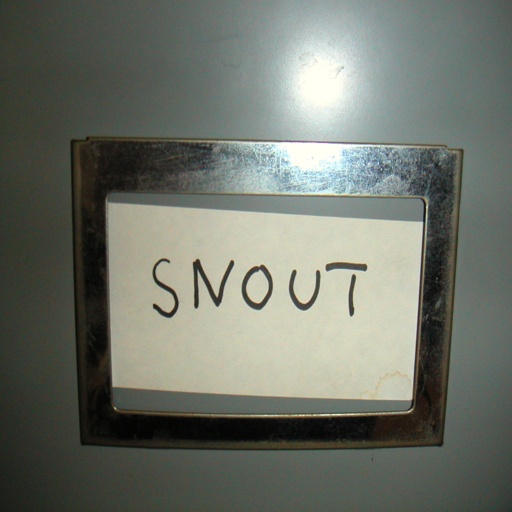}
        \end{minipage}%
    }%
    \subfigure[Masked]{
        \begin{minipage}[c]{0.18\columnwidth}
            \centering
            \includegraphics[width=0.95\columnwidth]{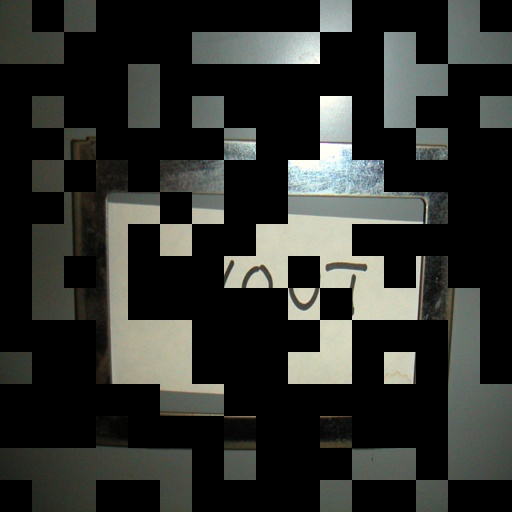}
        \end{minipage}%
    }%
    \subfigure[Reconstruct]{
        \begin{minipage}[c]{0.18\columnwidth}
            \centering
            \includegraphics[width=0.95\columnwidth]{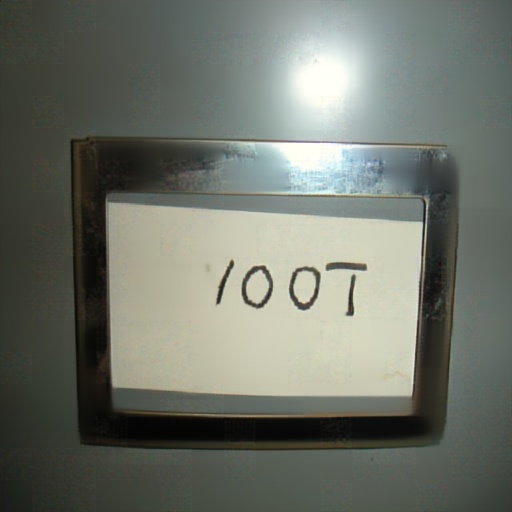}
        \end{minipage}%
    }%
    \subfigure[Pred. Mask]{
        \begin{minipage}[c]{0.18\columnwidth}
            \centering
            \includegraphics[width=0.95\columnwidth]{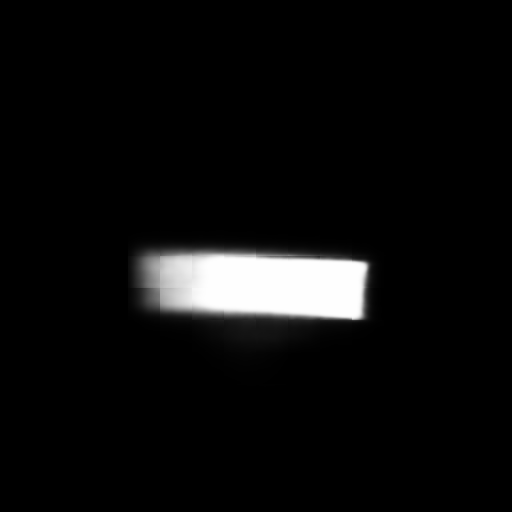}
        \end{minipage}%
    }%
    \subfigure[GT Mask]{
        \begin{minipage}[c]{0.18\columnwidth}
            \centering
            \includegraphics[width=0.95\columnwidth]{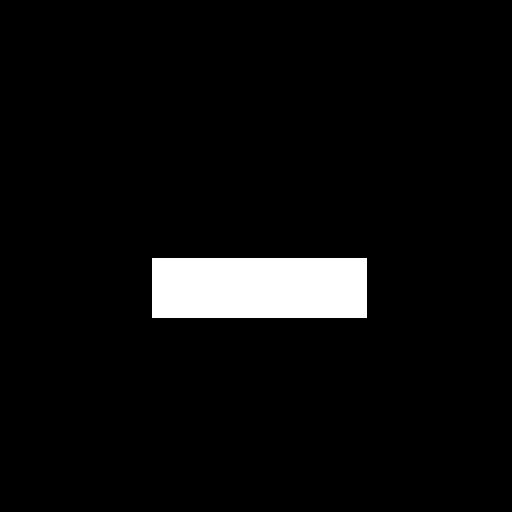}
        \end{minipage}%
    }
    \caption{Visualizations of SegMIM. (Pred.: Predicted)}
    \label{fig:exp_vis_segmim}
\end{figure}

\begin{table*}[t]
    \centering 
    \normalsize
    \resizebox{1.5\columnwidth}{!}{
        \begin{tabular}{llccccccccc}
        \toprule
        \multirow{2}*{Architecture} & \multirow{2}*{Scale} & \multicolumn{7}{c}{SCUT-EnsText} & \multirow{2}*{\thead{Params$\downarrow$ \\ (M)}} \\
        \cmidrule{3-9}
        & & PSNR$\uparrow$ & MSSIM$\uparrow$ & MSE$\downarrow$ & AGE$\downarrow$ & pEPs$\downarrow$ & pCEPs$\downarrow$ & FID$\downarrow$ & \\
        \midrule 
        \multirow{4}*{\thead{ViTEraser \\ (PVT)}} & Tiny & 34.48 & 96.98 & 0.1251 & 2.26 & 0.0124 & 0.0080 & 15.21 & \textbf{37.85} \\ 
        & Small & 35.03 & 97.15 & 0.1019 & 2.13 & 0.0112 & 0.0072 & 14.26 & \underline{60.36} \\ 
        & Medium & 35.09 & 97.16 & 0.1089 & 2.12 & 0.0105 & 0.0066 & 13.95 & 99.80 \\ 
        & Large & 34.91 & 97.08 & 0.1183 & 2.17 & 0.0116 & 0.0074 & 14.48 & 134.13\\ 
        \midrule 
        \multirow{3}*{\thead{ViTEraser \\ (Swin)}} & Tiny & 35.95 & 97.41 & 0.0647 & 1.87 & 0.0080 & 0.0045 & 12.38 & 65.26 \\ 
        & Small & 35.81 & 97.44 & 0.0589 & 2.01 & 0.0079 & 0.0044 & 11.94 & 107.90 \\ 
        & Base & 36.17 & 97.47 & 0.0637 & \underline{1.80} & 0.0078 & 0.0044 & 12.11 & 191.66 \\ 
        \midrule
        \multirow{3}*{\thead{ViTEraser \\ (Swinv2)}} & Tiny & \underline{36.32} & 97.48 & 0.0569 & 1.81 & \underline{0.0073} & \underline{0.0040} & 11.77 & 65.39 \\ 
        & Small & \textbf{36.55} & \textbf{97.56} & \textbf{0.0497} & \textbf{1.73} & \textbf{0.0072} & \textbf{0.0039} & \textbf{11.46} & 108.15 \\  
        & Base & \underline{36.32} & \underline{97.51} & \underline{0.0565} & 1.86 & 0.0074 & 0.0041 & \underline{11.68} & 191.97 \\ 
        \bottomrule
    \end{tabular}}
    \caption{Comparison of different scales of ViTEraser.}
    \label{tab:exp_scale}
\end{table*}

\noindent\textbf{Decoder}
\textbf{(1)}   
\textit{Deconv} decoder hierarchically upsamples a $16\times16$ feature map with 2048 channels to sizes of $\{32, 64, 128, 256, 512\}$ and channels of $\{1024, 512, 256, 64, 64\}$ through five deconvolution layers.
\textbf{(2)} 
\textit{TD+Deconv} decoder adds a 6-layer Transformer decoder with 256 channels before a \textit{Deconv} decoder.
With 256 learnable queries, the Transformer decoder produces a hidden feature of 256 tokens and 256 channels which is subsequently resized to a $16 \times 16 \times 256$ feature map. 
This feature map then undergoes a $1 \times 1$ convolution with 2048 channels before being processed by the \textit{Deconv} decoder. 
\textbf{(3)} 
\textit{MLP} follows the decoder of SegFormer \cite{xie2021segformer}. 
The multi-scale features produced by the encoder are transformed to 256 channels, then interpolated to a size of $512 \times 512$, and finally fused via a $1 \times 1$ convolution.

Based on the results in Tab.~\ref{tab:exp_arch}, the discussions are as follows.
\textbf{(1)} Inserting Transformer into CNNs cannot effectively improve the STR results (2\textit{nd} \& 3\textit{rd} rows \textit{v.s.} 1\textit{st} row).
The Transformer encoder only performs global attention on high-level features produced by CNN, omitting fine-grained correlations such as detailed textures.
Moreover, the learnable queries adopted in the Transformer decoder may cause spatial misalignment.
\textbf{(2)} 
Pure ViT-based encoder (4\textit{th} to 6\textit{th} rows) makes a significant improvement. 
The window-based Swinv2-Tiny can effectively capture local and global dependencies at both low- and high-level feature spaces.
\textbf{(3)} 
\textit{The ViTEraser, which thoroughly utilizes ViTs in both the encoder and decoder, provides the best architecture for applying Transformer to STR with a substantial margin.}
The Swinv2 blocks enable the decoder to fill the background considering both surrounding and long-distance context.

\begin{table*}[t]
    \centering 
    \normalsize
    \setlength{\tabcolsep}{1.5mm}
    \resizebox{2\columnwidth}{!}{
        \begin{tabular}{llcccccccccccc}
        \toprule 
        \multirow{2}*{Method} & \multirow{2}*{Venue} & \multicolumn{7}{c}{Image-Eval} & \multicolumn{3}{c}{Detection-Eval$\downarrow$} & \multirow{2}*{\thead{Params$\downarrow$ \\ (M)}} & \multirow{2}*{\thead{Speed$\uparrow$\\(fps)}} \\
        \cmidrule(r){3-9} \cmidrule(r){10-12}
        & & PSNR$\uparrow$ & MSSIM$\uparrow$ & MSE$\downarrow$ & AGE$\downarrow$ & pEPs$\downarrow$ & pCEPs$\downarrow$ & FID$\downarrow$ & R & P & F \\
        \midrule 
        Original & - & - & - & - & - & - & - & - & 69.5 & 79.4 & 74.1 & - & -\\ 
        Pix2pix \cite{isola2017image} & CVPR'17 & 26.70 & 88.56 & 0.37 & 6.09 & 0.0480 & 0.0227 & 46.88 & 35.4 & 69.7 & 47.0 & 54.42 & \underline{133}\\
        STE \cite{nakamura2017scene} &  ICDAR'17 &25.47 & 90.14 & 0.47 & 6.01 & 0.0533 & 0.0296 & 43.39 & 5.9 & 40.9 & 10.2 & - & - \\
        EnsNet \cite{zhang2019ensnet} & AAAI'19 & 29.54 & 92.74 & 0.24 & 4.16 & 0.0307 & 0.0136 & 32.71 & 32.8 & 68.7 & 44.4 & 12.40 & \textbf{199} \\
        MTRNet++ \cite{tursun2020mtrnet++} & CVIU'20 & 29.63 & 93.71 & 0.28 & 3.51 & 0.0305 & 0.0168 & 35.68 & 15.1 & 63.8 & 24.4 & 18.67 & 53 \\
        EraseNet \cite{liu2020erasenet} & TIP'20 &32.30 & 95.42 & 0.15 & 3.02 & 0.0160 & 0.0090 & 19.27 & 4.6 & 53.2 & 8.5 & 17.82 & 71 \\
        SSTE \cite{tang2021stroke} & TIP'21 & 35.34 & 96.24 & 0.09 & - & - & - & - & 3.6 & - & - & 30.75 & 7.8\\ 
        PSSTRNet \cite{lyu2022psstrnet} & ICME'22 & 34.65 & 96.75 & 0.14 & 1.72 & 0.0135 & 0.0074 & - & 5.1 & 47.7 & 9.3 & \textbf{4.88} & 56 \\
        CTRNet \cite{liu2022don} & ECCV'22 & 35.20 & 97.36 & 0.09 & 2.20 & 0.0106 & 0.0068 & 13.99 & 1.4 & 38.4 & 2.7 & 159.81 & 5.1 \\
        GaRNet$\S$ \cite{lee2022surprisingly} & ECCV'22 & 35.45 & 97.14 & 0.08 & 1.90 & 0.0105 & 0.0062 & 15.50 & 1.6 & 42.0 & 3.0 & 33.18 & 22 \\
        MBE \cite{hou2022multi} & ACCV'22 & 35.03 & 97.31 & - & 2.06 & 0.0128 & 0.0088 & - & - & - & - & - & - \\
        PEN \cite{du2022progressive} & CVIU'23 & 35.21 & 96.32 & 0.08 & 2.14 & 0.0097 & 0.0037 & - & 2.6 & 33.5 & 4.8 & - & - \\
        PEN* \cite{du2022progressive} & CVIU'23 & 35.72 & 96.68 & 0.05 & 1.95 & 0.0071 & 0.0020 & - & 2.1 & \textbf{26.2} & 3.9 & - & -  \\
        PERT \cite{wang2023pert} & TIP'23 & 33.62 & 97.00 & 0.13 & 2.19 & 0.0135 & 0.0088 & - & 4.1 & 50.5 & 7.6 & 14.00 & 24 \\ 
        SAEN \cite{du2023modeling} & WACV'23 & 34.75 & 96.53 & 0.07 & 1.98 & 0.0125 & 0.0073 & - & - & - & - & 19.79 & 62 \\
        FETNet \cite{lyu2023fetnet} & PR'23 & 34.53 & 97.01 & 0.13 & 1.75 & 0.0137 & 0.0080 & - & 5.8 & 51.3 & 10.5 & \underline{8.53} & 77 \\
        \midrule 
        ViTEraser-Tiny & - & 36.32 & 97.48 & 0.0569 & 1.81 & 0.0073 & 0.0040 & 11.77 & 0.717 & 32.7 & 1.403 & 65.39 & 24 \\ 
        ViTEraser-Tiny + SegMIM & - & 36.80 & 97.55 & 0.0491 & 1.79 & 0.0067 & 0.0036 & 10.79 & \underline{0.430} & \underline{27.3} & \underline{0.847} & 65.39 & 24 \\ 
        ViTEraser-Small & - & 36.55 & 97.56 & 0.0497 & 1.73 & 0.0072 & 0.0039 & 11.46 & 0.778 & 42.2 & 1.528 & 108.15 & 17 \\ 
        ViTEraser-Small + SegMIM & - & \underline{37.08} & \textbf{97.62} & \textbf{0.0447} & \textbf{1.69} & \textbf{0.0064} & \textbf{0.0034} & \underline{10.16} & \underline{0.430} & 30.9 & 0.848 & 108.15 & 17 \\ 
        ViTEraser-Base & - & 36.32 & 97.51 & 0.0565 & 1.86 & 0.0074 & 0.0041 & 11.68 & 0.635 & 37.8 & 1.248 & 191.97 & 15 \\ 
        ViTEraser-Base + SegMIM & - & \textbf{37.11} & \underline{97.61} & \underline{0.0474} & \underline{1.70} & \underline{0.0066} & \underline{0.0035} & \textbf{10.15} & \textbf{0.389} & 29.7 & \textbf{0.768} & 191.97 & 15 \\ 
        \bottomrule
    \end{tabular}}
    \caption{Comparison with state of the arts on SCUT-EnsText. (Bold: state of the art, underline: the second best)}
    \label{tab:exp_scut_enstext}
\end{table*}

\begin{figure*}[!h]
    \centering
    \subfigtopskip=0pt 
    \subfigbottomskip=2pt 
    \subfigcapskip=1pt 
    \renewcommand{\subcapsize}{\scriptsize}
    \subfigure{
        \begin{minipage}[c]{0.1\linewidth}
            \centering
            \includegraphics[width=\linewidth]{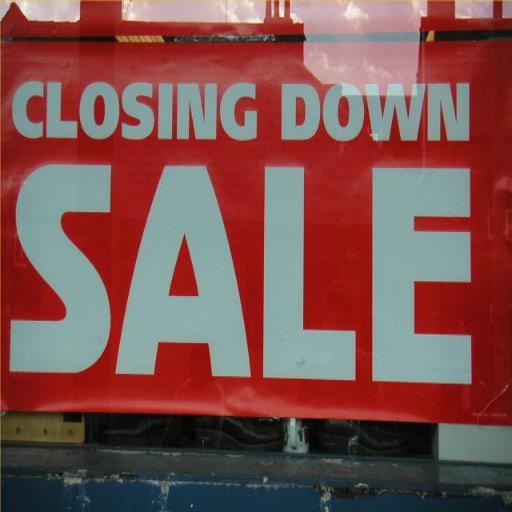}
        \end{minipage}%
    }%
    \subfigure{
        \begin{minipage}[c]{0.1\linewidth}
            \centering
            \includegraphics[width=\linewidth]{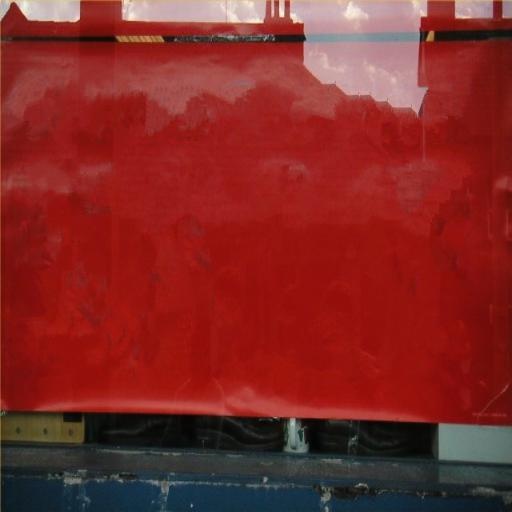}
        \end{minipage}%
    }%
    \subfigure{
        \begin{minipage}[c]{0.1\linewidth}
            \centering
            \includegraphics[width=\linewidth]{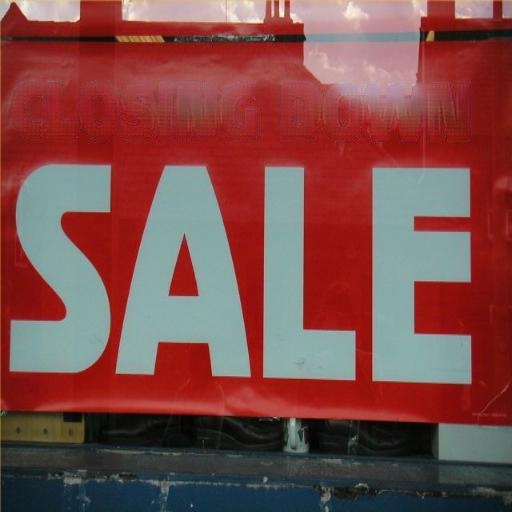}
        \end{minipage}%
    }%
    \subfigure{
        \begin{minipage}[c]{0.1\linewidth}
            \centering
            \includegraphics[width=\linewidth]{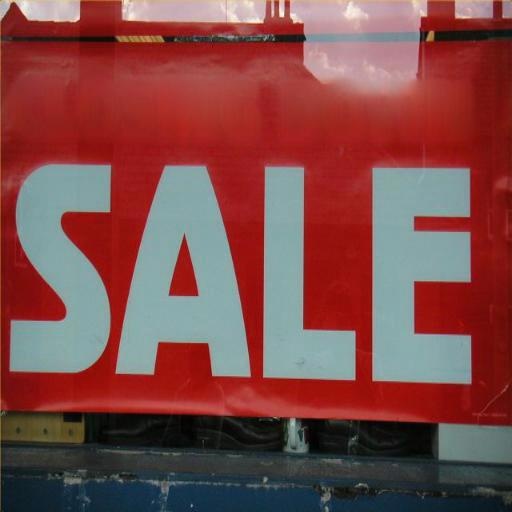}
        \end{minipage}%
    }%
    \subfigure{
        \begin{minipage}[c]{0.1\linewidth}
            \centering
            \includegraphics[width=\linewidth]{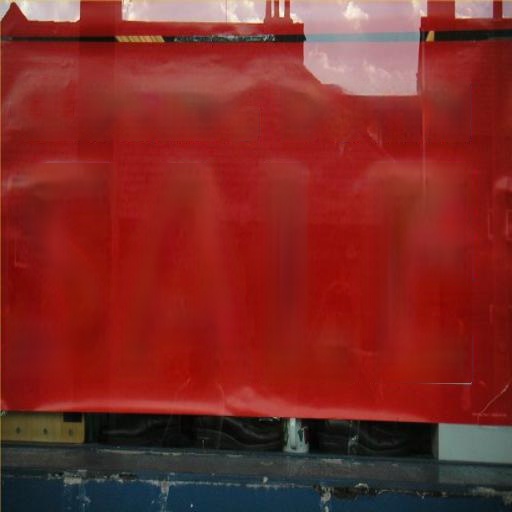}
        \end{minipage}%
    }%
    \subfigure{
        \begin{minipage}[c]{0.1\linewidth}
            \centering
            \includegraphics[width=\linewidth]{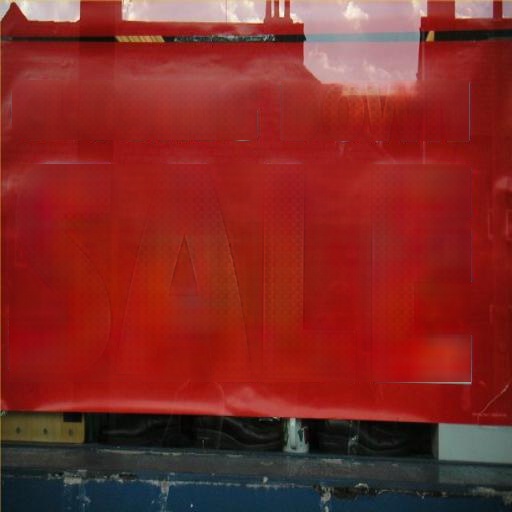}
        \end{minipage}%
    }%
    \subfigure{
        \begin{minipage}[c]{0.1\linewidth}
            \centering
            \includegraphics[width=\linewidth]{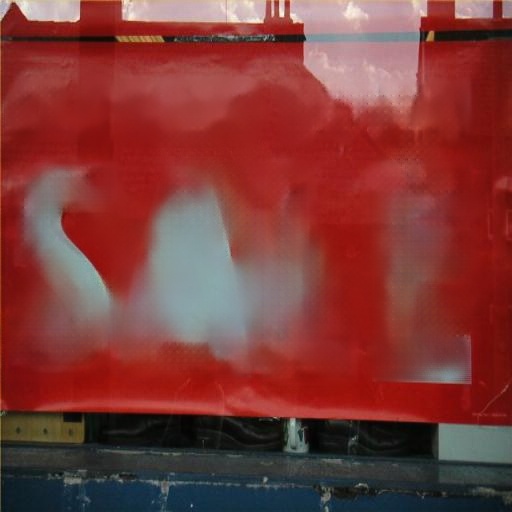}
        \end{minipage}%
    }%
    \subfigure{
        \begin{minipage}[c]{0.1\linewidth}
            \centering
            \includegraphics[width=\linewidth]{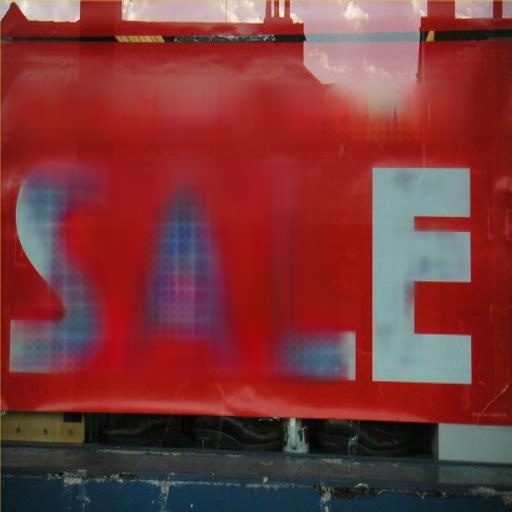}
        \end{minipage}%
    }%
    \subfigure{
        \begin{minipage}[c]{0.1\linewidth}
            \centering
            \includegraphics[width=\linewidth]{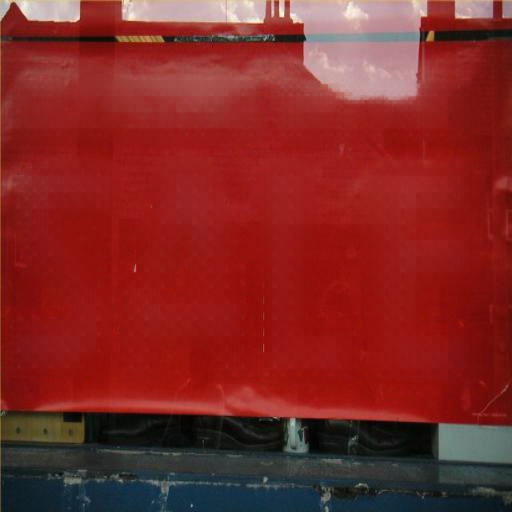}
        \end{minipage}%
    }
    \subfigure{
        \begin{minipage}[c]{0.1\linewidth}
            \centering
            \includegraphics[width=\linewidth]{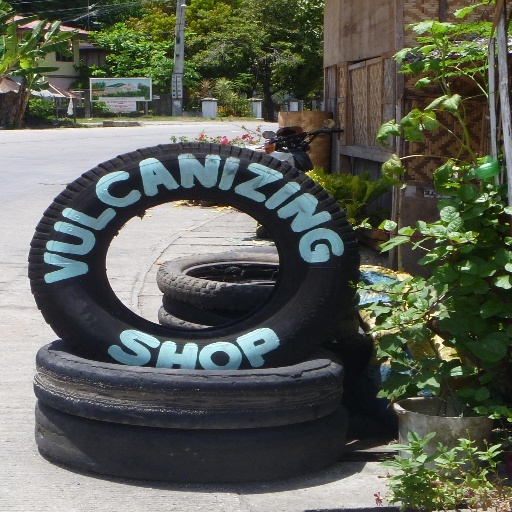}
        \end{minipage}%
    }%
    \subfigure{
        \begin{minipage}[c]{0.1\linewidth}
            \centering
            \includegraphics[width=\linewidth]{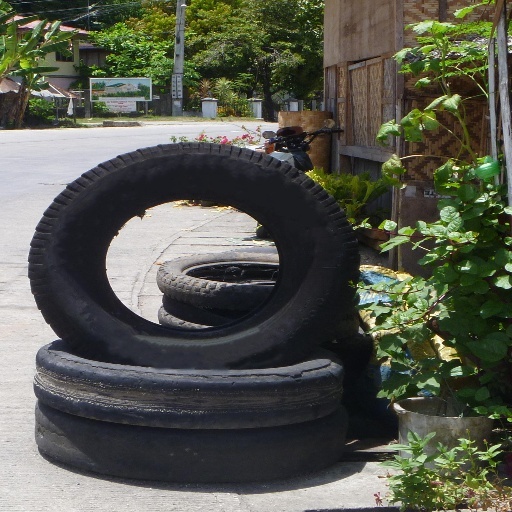}
        \end{minipage}%
    }%
    \subfigure{
        \begin{minipage}[c]{0.1\linewidth}
            \centering
            \includegraphics[width=\linewidth]{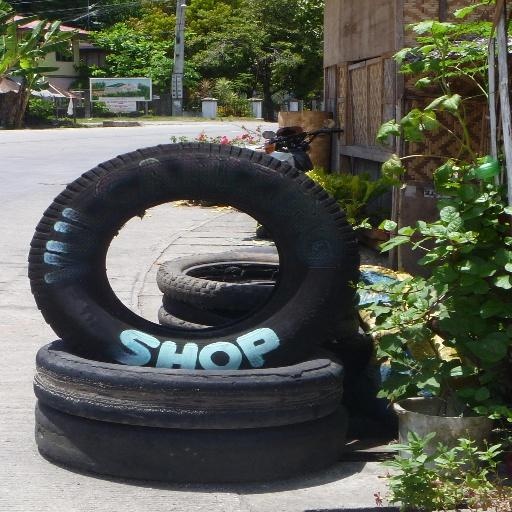}
        \end{minipage}%
    }%
    \subfigure{
        \begin{minipage}[c]{0.1\linewidth}
            \centering
            \includegraphics[width=\linewidth]{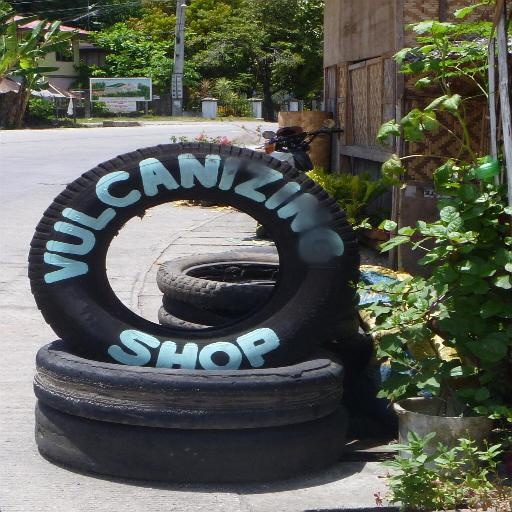}
        \end{minipage}%
    }%
    \subfigure{
        \begin{minipage}[c]{0.1\linewidth}
            \centering
            \includegraphics[width=\linewidth]{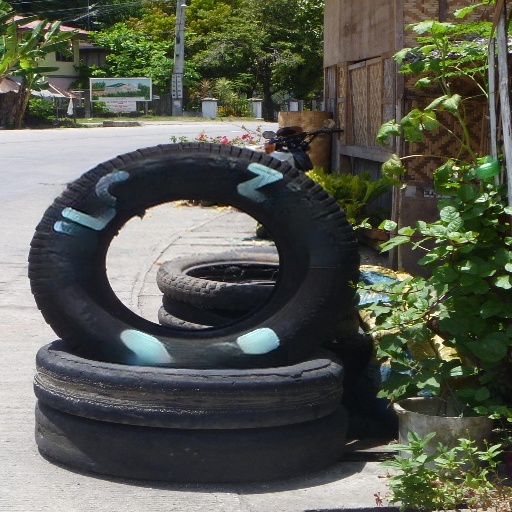}
        \end{minipage}%
    }%
    \subfigure{
        \begin{minipage}[c]{0.1\linewidth}
            \centering
            \includegraphics[width=\linewidth]{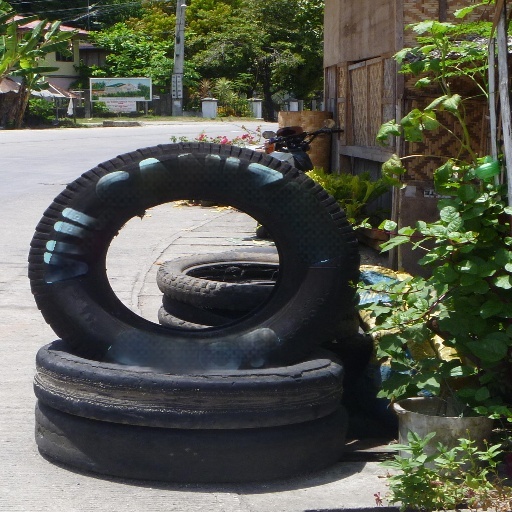}
        \end{minipage}%
    }%
    \subfigure{
        \begin{minipage}[c]{0.1\linewidth}
            \centering
            \includegraphics[width=\linewidth]{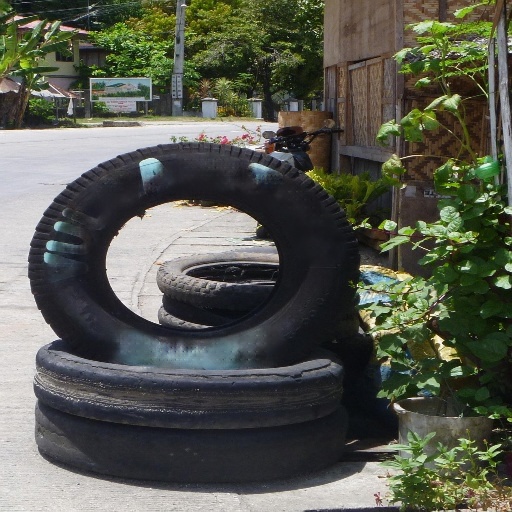}
        \end{minipage}%
    }%
    \subfigure{
        \begin{minipage}[c]{0.1\linewidth}
            \centering
            \includegraphics[width=\linewidth]{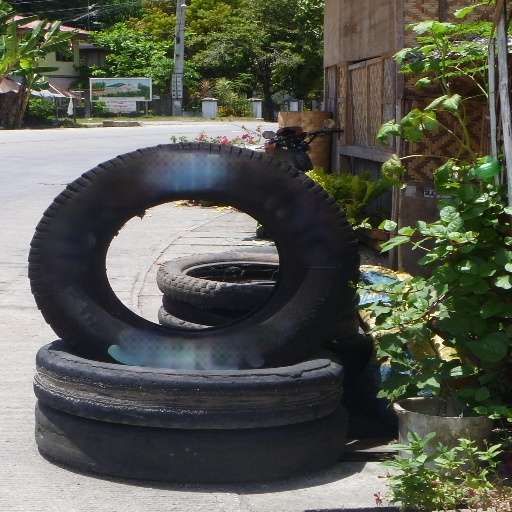}
        \end{minipage}%
    }%
    \subfigure{
        \begin{minipage}[c]{0.1\linewidth}
            \centering
            \includegraphics[width=\linewidth]{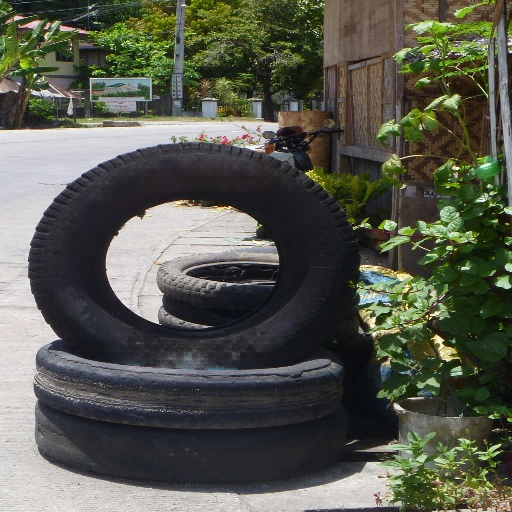}
        \end{minipage}%
    }
    \subfigure{
        \begin{minipage}[c]{0.1\linewidth}
            \centering
            \includegraphics[width=\linewidth]{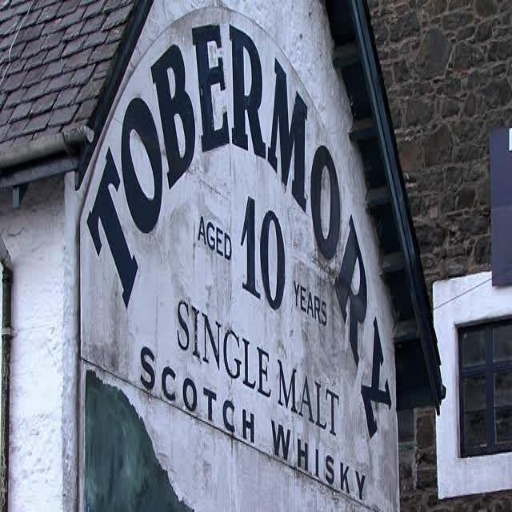}
        \end{minipage}%
    }%
    \subfigure{
        \begin{minipage}[c]{0.1\linewidth}
            \centering
            \includegraphics[width=\linewidth]{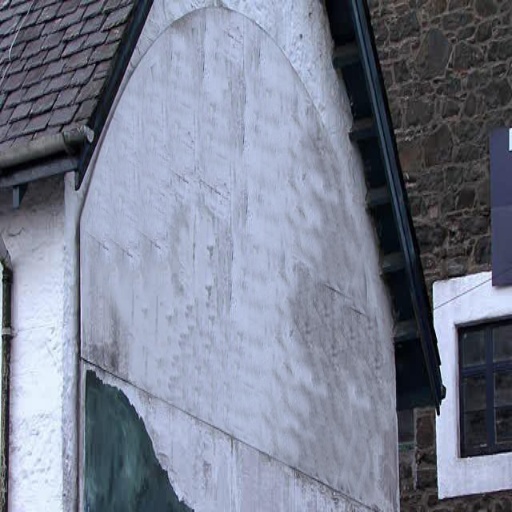}
        \end{minipage}%
    }%
    \subfigure{
        \begin{minipage}[c]{0.1\linewidth}
            \centering
            \includegraphics[width=\linewidth]{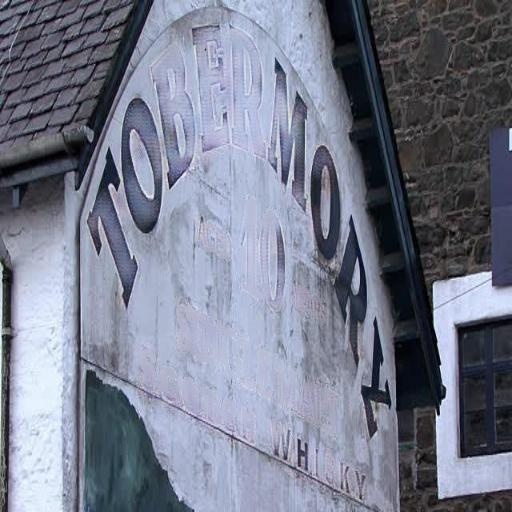}
        \end{minipage}%
    }%
    \subfigure{
        \begin{minipage}[c]{0.1\linewidth}
            \centering
            \includegraphics[width=\linewidth]{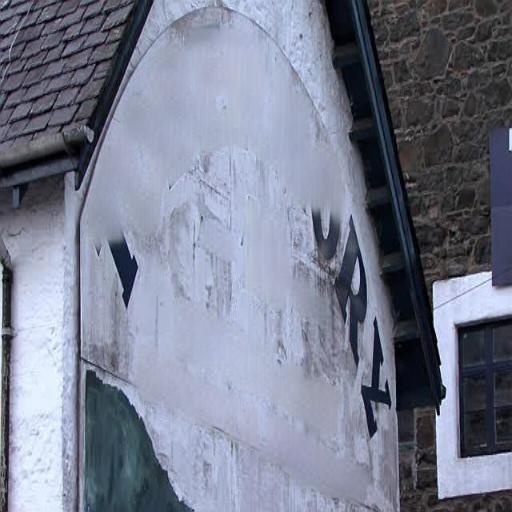}
        \end{minipage}%
    }%
    \subfigure{
        \begin{minipage}[c]{0.1\linewidth}
            \centering
            \includegraphics[width=\linewidth]{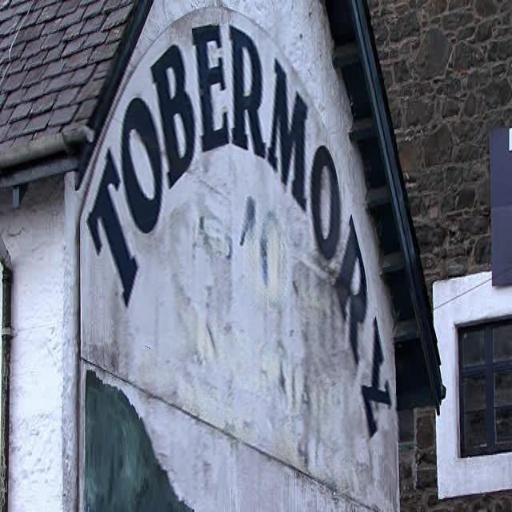}
        \end{minipage}%
    }%
    \subfigure{
        \begin{minipage}[c]{0.1\linewidth}
            \centering
            \includegraphics[width=\linewidth]{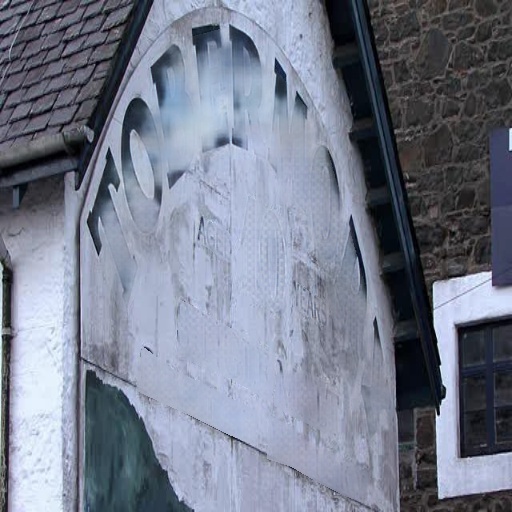}
        \end{minipage}%
    }%
    \subfigure{
        \begin{minipage}[c]{0.1\linewidth}
            \centering
            \includegraphics[width=\linewidth]{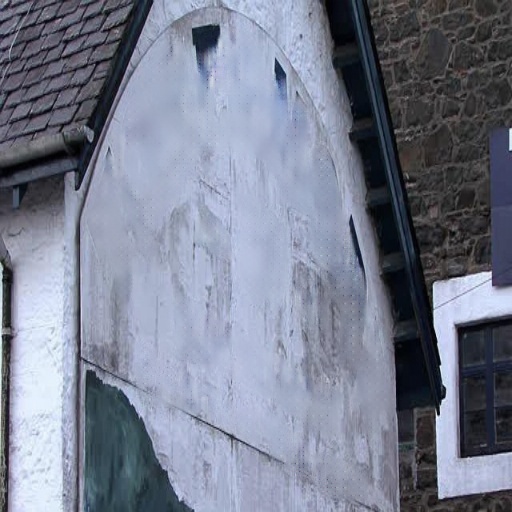}
        \end{minipage}%
    }%
    \subfigure{
        \begin{minipage}[c]{0.1\linewidth}
            \centering
            \includegraphics[width=\linewidth]{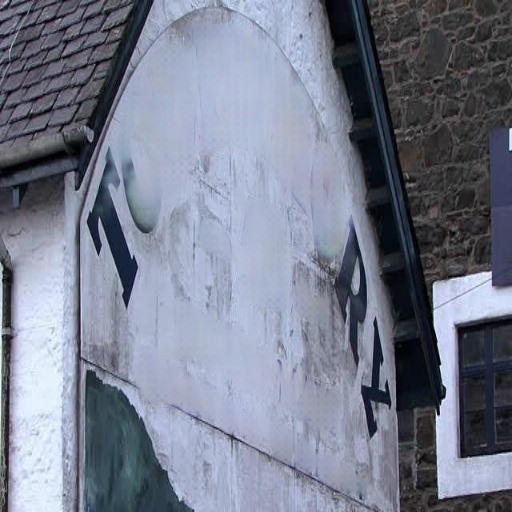}
        \end{minipage}%
    }%
    \subfigure{
        \begin{minipage}[c]{0.1\linewidth}
            \centering
            \includegraphics[width=\linewidth]{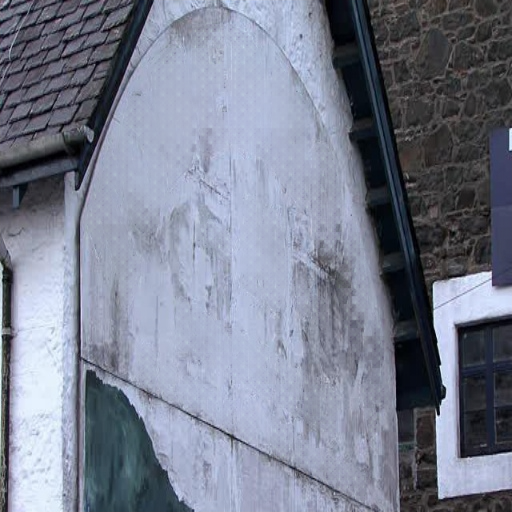}
        \end{minipage}%
    }
    \setcounter{subfigure}{0}
    \subfigure[Input]{
        \begin{minipage}[c]{0.1\linewidth}
            \centering
            \includegraphics[width=\linewidth]{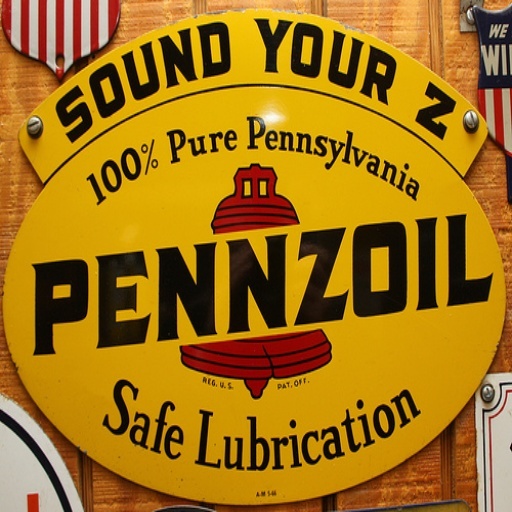}
        \end{minipage}%
    }%
    \subfigure[GT]{
        \begin{minipage}[c]{0.1\linewidth}
            \centering
            \includegraphics[width=\linewidth]{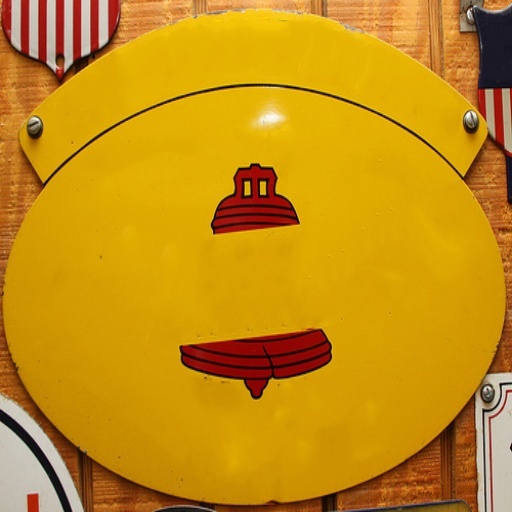}
        \end{minipage}%
    }%
    \subfigure[MTRNet++]{
        \begin{minipage}[c]{0.1\linewidth}
            \centering
            \includegraphics[width=\linewidth]{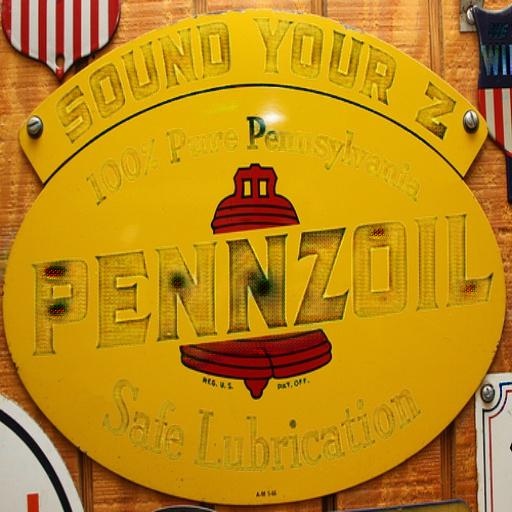}
        \end{minipage}%
    }%
    \subfigure[EraseNet]{
        \begin{minipage}[c]{0.1\linewidth}
            \centering
            \includegraphics[width=\linewidth]{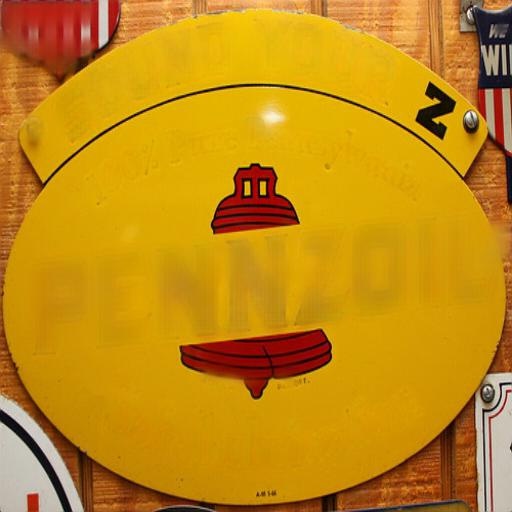}
        \end{minipage}%
    }%
    \subfigure[SSTE]{
        \begin{minipage}[c]{0.1\linewidth}
            \centering
            \includegraphics[width=\linewidth]{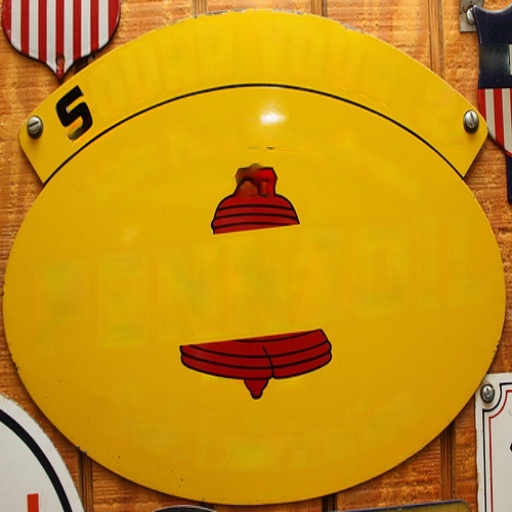}
        \end{minipage}%
    }%
    \subfigure[GaRNet]{
        \begin{minipage}[c]{0.1\linewidth}
            \centering
            \includegraphics[width=\linewidth]{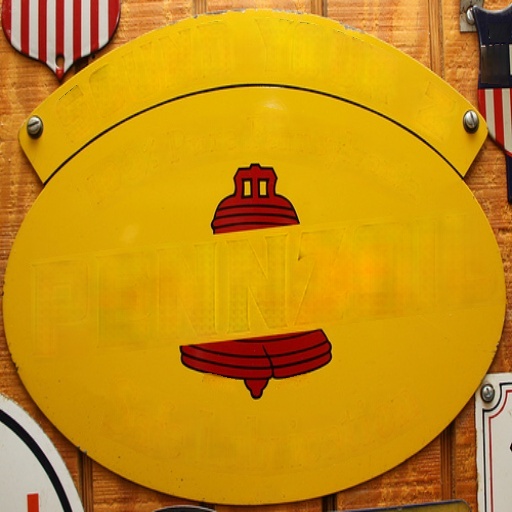}
        \end{minipage}%
    }%
    \subfigure[CTRNet]{
        \begin{minipage}[c]{0.1\linewidth}
            \centering
            \includegraphics[width=\linewidth]{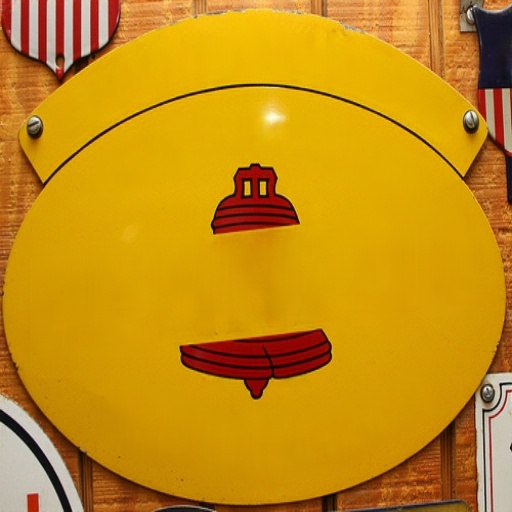}
        \end{minipage}%
    }%
    \subfigure[PERT]{
        \begin{minipage}[c]{0.1\linewidth}
            \centering
            \includegraphics[width=\linewidth]{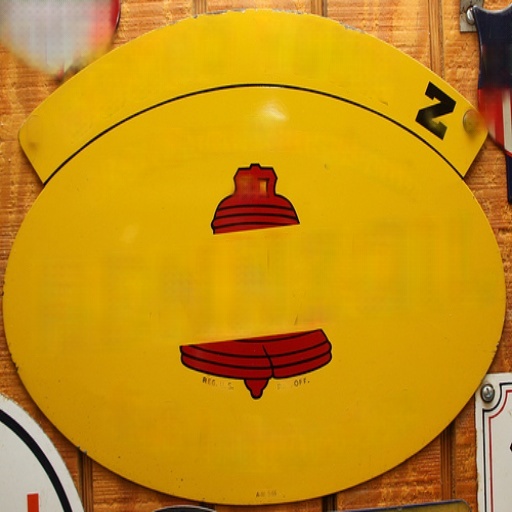}
        \end{minipage}%
    }%
    \subfigure[ViTEraser]{
        \begin{minipage}[c]{0.1\linewidth}
            \centering
            \includegraphics[width=\linewidth]{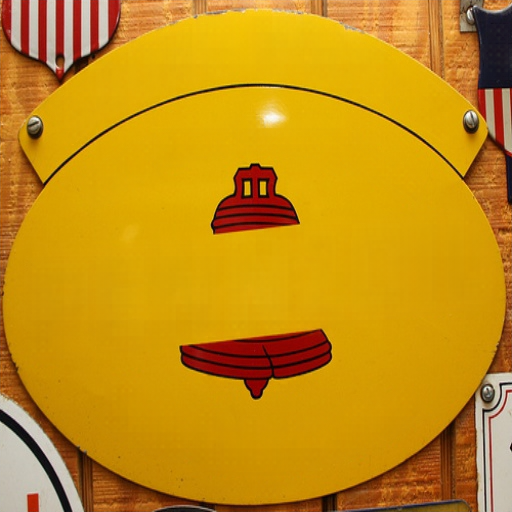}
        \end{minipage}%
    }
    \caption{
    Qualitative comparison of existing methods and ViTEraser-Small (w/ SegMIM) on SCUT-EnsText.
    }
    \label{fig:exp_vis_scutens}
\end{figure*}

\subsection{Experiments on Pretraining}
In this section, we comprehensively explore pretraining schemes for STR based on ViTEraser.
The pretraining strategies for comparison include:
\textbf{(1)} 
When using ImageNet-1k, the encoder can be pretrained with the classification or SimMIM tasks. 
Moreover, because the first four stages of the decoder are empirically set to be symmetric to the encoder, their parameters can also be initialized by the encoder's pretrained weights symmetrically.
\textbf{(2)} 
When using scene text detection datasets, the encoder or encoder-decoder can be pretrained with text box segmentation or SimMIM tasks.

The experiment results in Tab.~\ref{tab:exp_pretrain} demonstrate that SegMIM achieves the best performance.
Moreover, Fig.~\ref{fig:exp_vis_segmim} illustrates the visualizations of SegMIM.
It can be seen that the pretrained model can accurately determine text locations and realistically reconstruct masked patches.

\subsection{Experiments on Scalability}
We investigate the scalability of ViTEraser in Tab.~\ref{tab:exp_scale}.
It can be seen that as the scale goes up, the performance tends to increase in general. 
However, for ViTEraser-Swinv2 and ViTEraser-PVT, the performance of the largest scale is inferior to a smaller one. 
This may be due to the overfitting caused by the dramatically increased parameters and limited training samples.
However, the potential of large models can be stimulated when pretrained with SegMIM (Tab.~\ref{tab:exp_scut_enstext}).

\subsection{Comparison with State of the Arts}

\subsubsection{SCUT-EnsText}
\label{sec:exp_comp_sota_enstext}
In Tab.~\ref{tab:exp_scut_enstext}, we compare ViTEraser with existing approaches on SCUT-EnsText.
For a fair comparison, instead of using GT text box masks, MTRNet++ \cite{tursun2020mtrnet++} uses empty coarse masks and GaRNet \cite{lee2022surprisingly} uses the text box masks produced by pretrained CRAFT \cite{baek2019character}.
All inference speeds are tested using an RTX3090 GPU with a batch size of 1, considering the time consumption of the model forward and post-processing.
As for the model size, we calculate the number of minimum required parameters during inference.
Besides, the parameters and time cost of external text detectors are considered for SSTE \cite{tang2021stroke}, GaRNet, and CTRNet \cite{liu2022don}.

The quantitative results in Tab.~\ref{tab:exp_scut_enstext} demonstrate the state-of-the-art performance of ViTEraser on real-world STR.
For image-eval metrics, a substantial improvement can be observed over previous methods, \textit{e.g.}, boosting PSNR from 35.72 dB to 37.11 dB. 
For detection-eval metrics, the recall and f-measure reach a milestone of lower than 1\%, indicating nearly all the texts have been effectively erased. 
Especially for ViTEraser-Base with SegMIM, remarkably low recall (0.389\%) and f-measure (0.768\%) have been achieved.
Moreover, SegMIM significantly boosts all three scales of ViTEraser, improving PSNRs of ViTEraser-Tiny, Small, and Base by 0.48, 0.53, and 0.79 dB, respectively.

\begin{table*}
    \centering
    \normalsize
    \resizebox{1.55\columnwidth}{!}{
        \begin{tabular}{llcccccc}
        \toprule 
        Method & Venue & PSNR$\uparrow$ & MSSIM$\uparrow$ & MSE$\downarrow$ & AGE$\downarrow$ & pEPs$\downarrow$ & pCEPs$\downarrow$ \\
        \midrule 
        Pix2pix \cite{isola2017image} & CVPR'17 & 26.76 & 91.08 & 0.27 & 5.47 & 0.0473 & 0.0244 \\
        STE \cite{nakamura2017scene} & ICDAR'17 & 25.40 & 90.12 & 0.65 & 9.49 & 0.0553 & 0.0347 \\
        EnsNet \cite{zhang2019ensnet} &  AAAI'19 & 37.36 & 96.44 & 0.21 & 1.73 & 0.0069 & 0.0020 \\
        MTRNet++ \cite{tursun2020mtrnet++} & CVIU'20 & 34.55 & 98.45 & 0.04 & - & - & - \\
        EraseNet \cite{liu2020erasenet} & TIP'20 & 38.32 & 97.67 & 0.02 & 1.60 & 0.0048 & 0.0004 \\
        \citet{zdenek2020erasing} & WACV'20 & 37.46 & 93.64 & - & - & - & - \\
        \citet{conrad2021two} & ICIP'21 & 32.97 & 94.90 & - & - & - & - \\
        SSTE \cite{tang2021stroke} & TIP'21 & 38.60 & 97.55 & 0.02 & - & - & - \\
        PSSTRNet \cite{lyu2022psstrnet} & ICME'22 & 39.25 & 98.15 & 0.02 & 1.20 & 0.0043 & 0.0008 \\
        CTRNet \cite{liu2022don} & ECCV'22 & 41.28 & 98.52 & 0.02 & 1.33 & 0.0030 & 0.0007 \\
        MBE \cite{hou2022multi} & ACCV'22 & \textbf{43.85} & \textbf{98.64} & - & \textbf{0.94} & \textbf{0.0013} & 0.00004 \\
        PEN \cite{du2022progressive} & CVIU'23 & 39.26 & 98.03 & 0.02 & 1.29 & 0.0038 & 0.0004 \\
        PEN* \cite{du2022progressive} & CVIU'23 & 38.87 & 97.83 & 0.03 & 1.38 & 0.0041 & 0.0004 \\
        PERT \cite{wang2023pert} & TIP'23 & 39.40 & 97.87 & 0.02 & 1.41 & 0.0046 & 0.0007 \\
        SEAN \cite{du2023modeling} & WACV'23 & 38.63 & 98.27 & 0.03 & 1.39 & 0.0043 & 0.0004 \\
        FETNet \cite{lyu2023fetnet} & PR'23 & 39.14 & 97.97 & 0.02 & 1.26 & 0.0046 & 0.0008 \\
        \midrule 
        ViTEraser-Tiny & - & 42.24 & 98.42 & 0.0112 & 1.23 & 0.0021 & 0.000020 \\ 
        ViTEraser-Tiny + SegMIM & - & 42.40 & 98.44 & 0.0106 & 1.17 & 0.0018 & 0.000015 \\ 
        ViTEraser-Small & - & 42.45 & 98.43 & 0.0109 & 1.19 & 0.0020 & 0.000019 \\
        ViTEraser-Small + SegMIM & - & 42.66 & 98.49 & \underline{0.0099} & 1.13 & 0.0016 & \underline{0.000012} \\
        ViTEraser-Base & - & 42.53 & 98.45 & 0.0102 & 1.19 & 0.0018 & 0.000016\\
        ViTEraser-Base + SegMIM & - & \underline{42.97} & \underline{98.55} & \textbf{0.0092} & \underline{1.11} & \underline{0.0015} & \textbf{0.000011} \\
        \bottomrule 
    \end{tabular}}
    \caption{Comparison with state of the arts on SCUT-Syn.}
    \label{tab:exp_scut-syn}
\end{table*}

\begin{figure}[t]
    \centering
    \subfigtopskip=0pt 
    \subfigbottomskip=2pt 
    \subfigcapskip=1pt 
    \renewcommand{\subcapsize}{\scriptsize}
    \subfigure{
        \begin{minipage}[c]{0.188\linewidth}
            \centering
            \includegraphics[width=\linewidth]{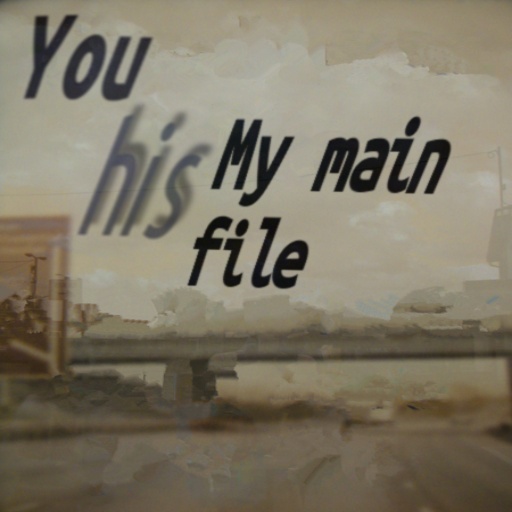}
        \end{minipage}%
    }%
    \subfigure{
        \begin{minipage}[c]{0.188\linewidth}
            \centering
            \includegraphics[width=\linewidth]{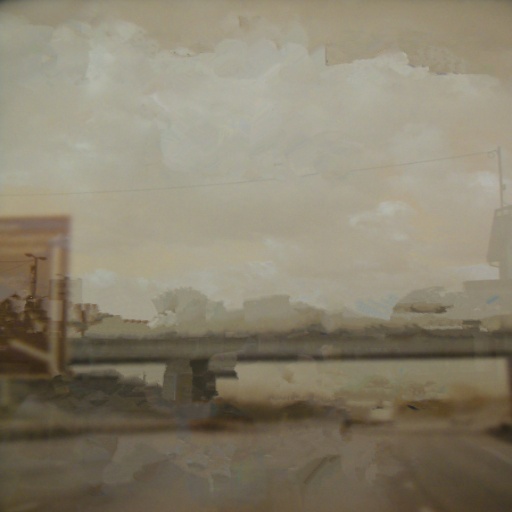}
        \end{minipage}%
    }%
    \subfigure{
        \begin{minipage}[c]{0.188\linewidth}
            \centering
            \includegraphics[width=\linewidth]{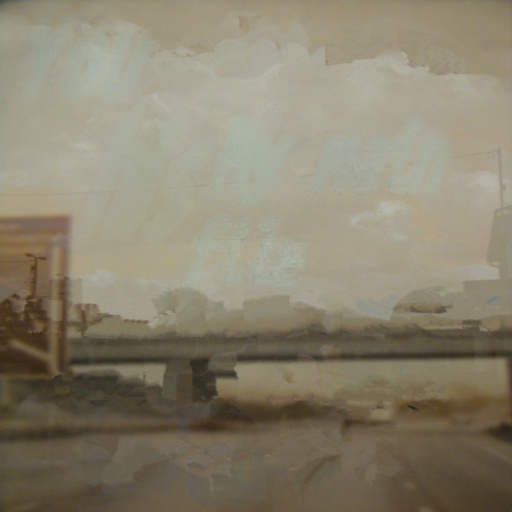}
        \end{minipage}%
    }%
    \subfigure{
        \begin{minipage}[c]{0.188\linewidth}
            \centering
            \includegraphics[width=\linewidth]{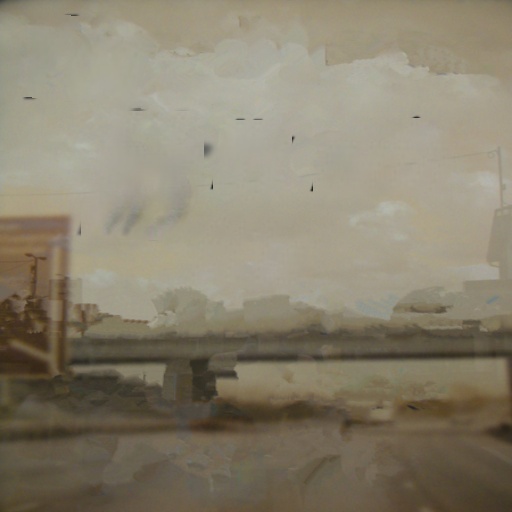}
        \end{minipage}%
    }%
    \subfigure{
        \begin{minipage}[c]{0.188\linewidth}
            \centering
            \includegraphics[width=\linewidth]{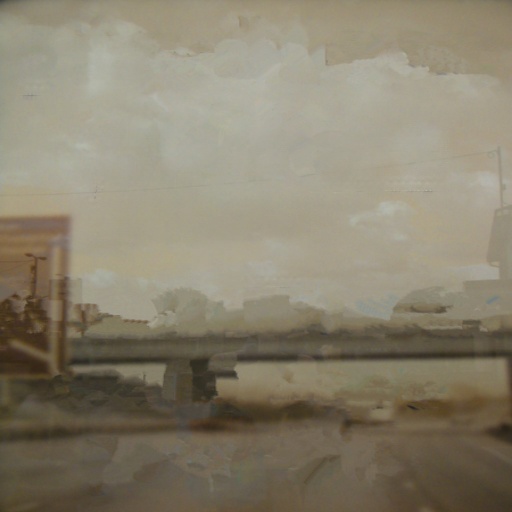}
        \end{minipage}%
    }
    \setcounter{subfigure}{0}
    \subfigure[Input]{
        \begin{minipage}[c]{0.188\linewidth}
            \centering
            \includegraphics[width=\linewidth]{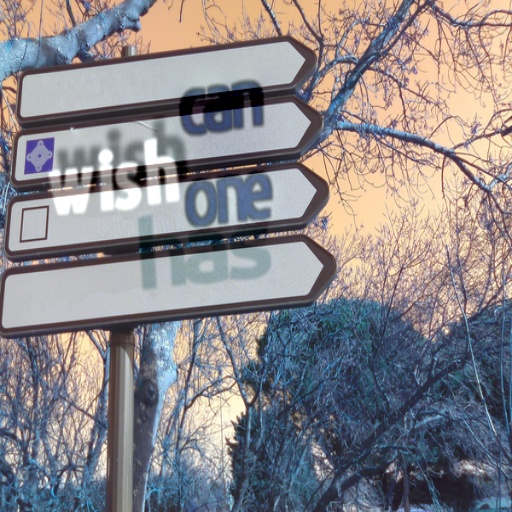}
        \end{minipage}%
    }%
    \subfigure[GT]{
        \begin{minipage}[c]{0.188\linewidth}
            \centering
            \includegraphics[width=\linewidth]{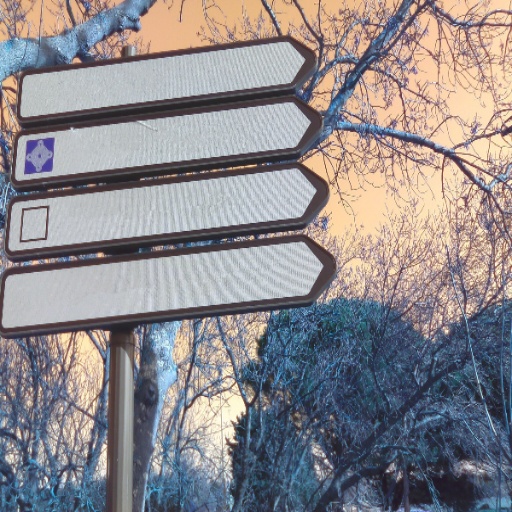}
        \end{minipage}%
    }%
    \subfigure[EraseNet]{
        \begin{minipage}[c]{0.188\linewidth}
            \centering
            \includegraphics[width=\linewidth]{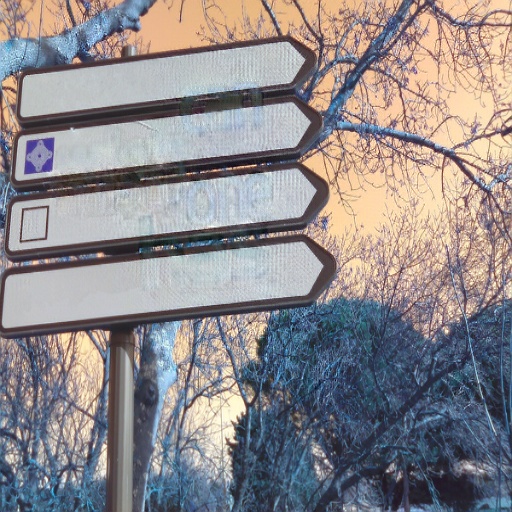}
        \end{minipage}%
    }%
    \subfigure[SSTE]{
        \begin{minipage}[c]{0.188\linewidth}
            \centering
            \includegraphics[width=\linewidth]{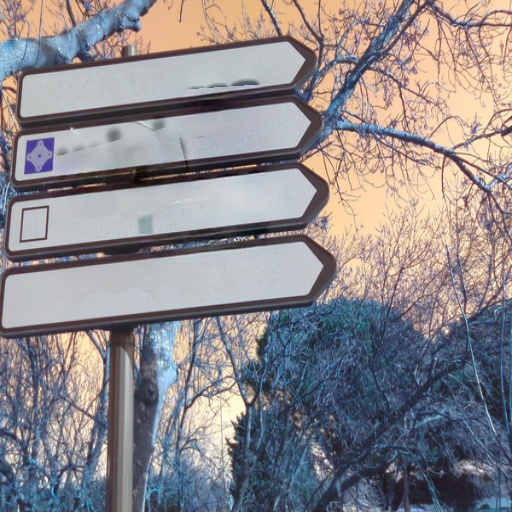}
        \end{minipage}%
    }%
    \subfigure[ViTEraser]{
        \begin{minipage}[c]{0.188\linewidth}
            \centering
            \includegraphics[width=\linewidth]{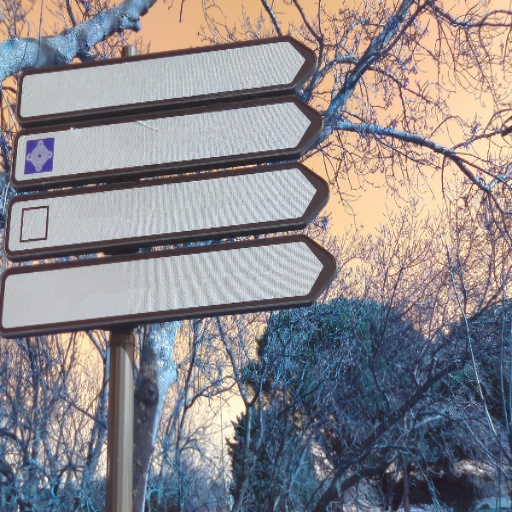}
        \end{minipage}%
    }
    \caption{Qualitative comparison of previous methods and ViTEraser-Base (w/ SegMIM) on SCUT-Syn.}
    \label{fig:exp_vis_scutsyn}
\end{figure}

The visualizations on SCUT-EnsText are shown in Fig.~\ref{fig:exp_vis_scutens}, qualitatively demonstrating the effectiveness of ViTEraser.

\subsubsection{SCUT-Syn}
The quantitative and qualitative comparisons on synthetic SCUT-Syn are presented in Tab.~\ref{tab:exp_scut-syn} and Fig.~\ref{fig:exp_vis_scutsyn}, respectively.
It can be observed that ViTEraser outperforms existing methods except for the MBE \cite{hou2022multi} that ensembles multiple STR networks.

\begin{table*}[t]
    \centering 
    \normalsize
    \resizebox{1.4\columnwidth}{!}{
    \begin{tabular}{lccccccc}
        \toprule
        \multirow{2}*{Method} & \multicolumn{3}{c}{Tampered Text} & \multicolumn{3}{c}{Real Text} & \multirow{2}*{mF}  \\
        \cmidrule(r){2-4} \cmidrule(r){5-7}
        & R & P & F & R & P & F & \\
        \midrule
        S3R \cite{wang2022detecting} + EAST & 69.97 & 70.23 & 69.94 & 27.32 & 50.46 & 35.45 & 52.70 \\
        ViTEraser-Tiny + EAST & 77.87 & 79.66 & 78.76 & 32.45 & 65.23 & 43.34 & 61.05 \\
        \midrule
        S3R \cite{wang2022detecting} + PSENet & 79.43 & 79.92 & 79.67 & 41.89 & 61.56 & 49.85 & 64.76 \\
        ViTEraser-Tiny + PSENet & 82.38 & 83.23 & 82.80 & 39.70 & 64.96 & 49.28 & 66.04 \\
        \midrule 
        S3R \cite{wang2022detecting} + ContourNet & \underline{91.45} & \textbf{86.68} & \underline{88.99} & \underline{54.80} & \textbf{77.88} & \underline{64.33} & \underline{76.66} \\
        ViTEraser-Tiny + ContourNet & \textbf{92.62} & \underline{85.77} & \textbf{89.06} & \textbf{56.84} & \underline{75.82} & \textbf{64.97} & \textbf{77.02} \\
        \bottomrule
    \end{tabular}}
    \caption{Comparison with existing methods on Tampered-IC13. (mF: Average f-measure of real and tampered texts) }
    \label{tab:exp_tamperd_ic13}
\end{table*}

\begin{figure}
    \centering
    \subfigtopskip=0pt 
    \subfigbottomskip=2pt 
    \subfigcapskip=1pt 
    \renewcommand{\subcapsize}{\scriptsize}
    \subfigure{
        \begin{minipage}[c]{0.188\linewidth}
            \centering
            \includegraphics[width=\linewidth,height=\linewidth]{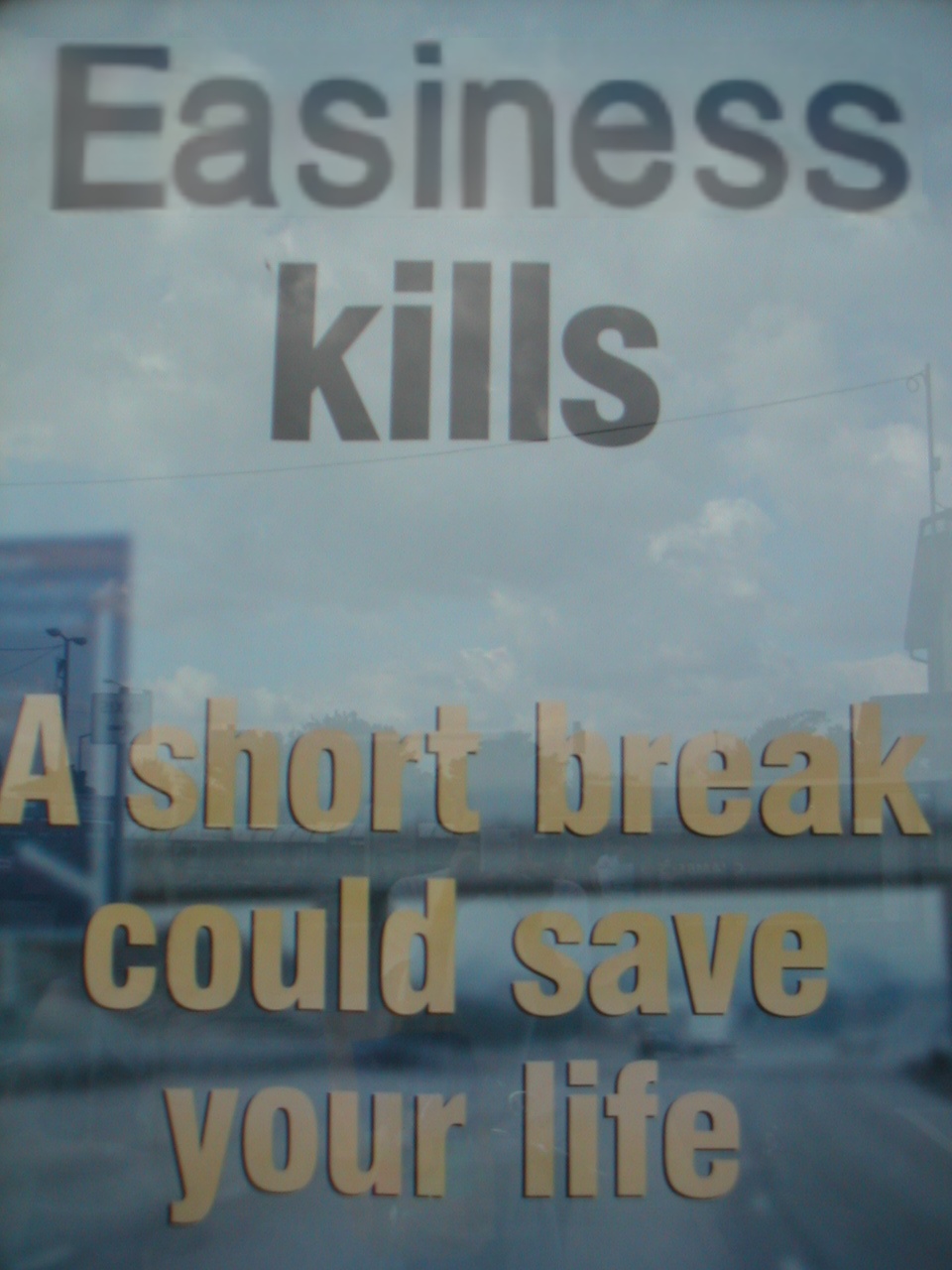}
        \end{minipage}%
    }%
    \subfigure{
        \begin{minipage}[c]{0.188\linewidth}
            \centering
            \includegraphics[width=\linewidth,height=\linewidth]{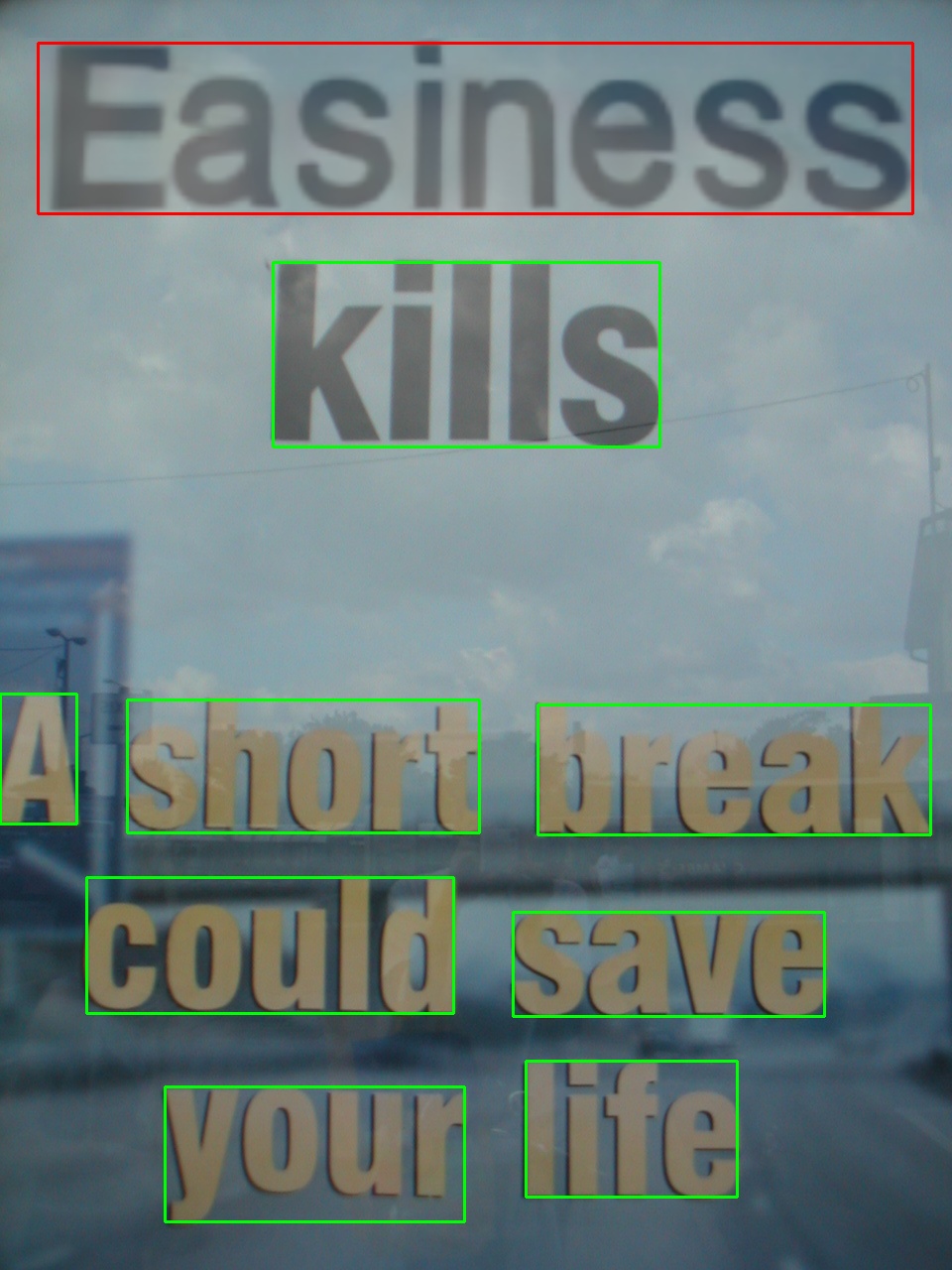}
        \end{minipage}%
    }%
    \subfigure{
        \begin{minipage}[c]{0.188\linewidth}
            \centering
            \includegraphics[width=\linewidth]{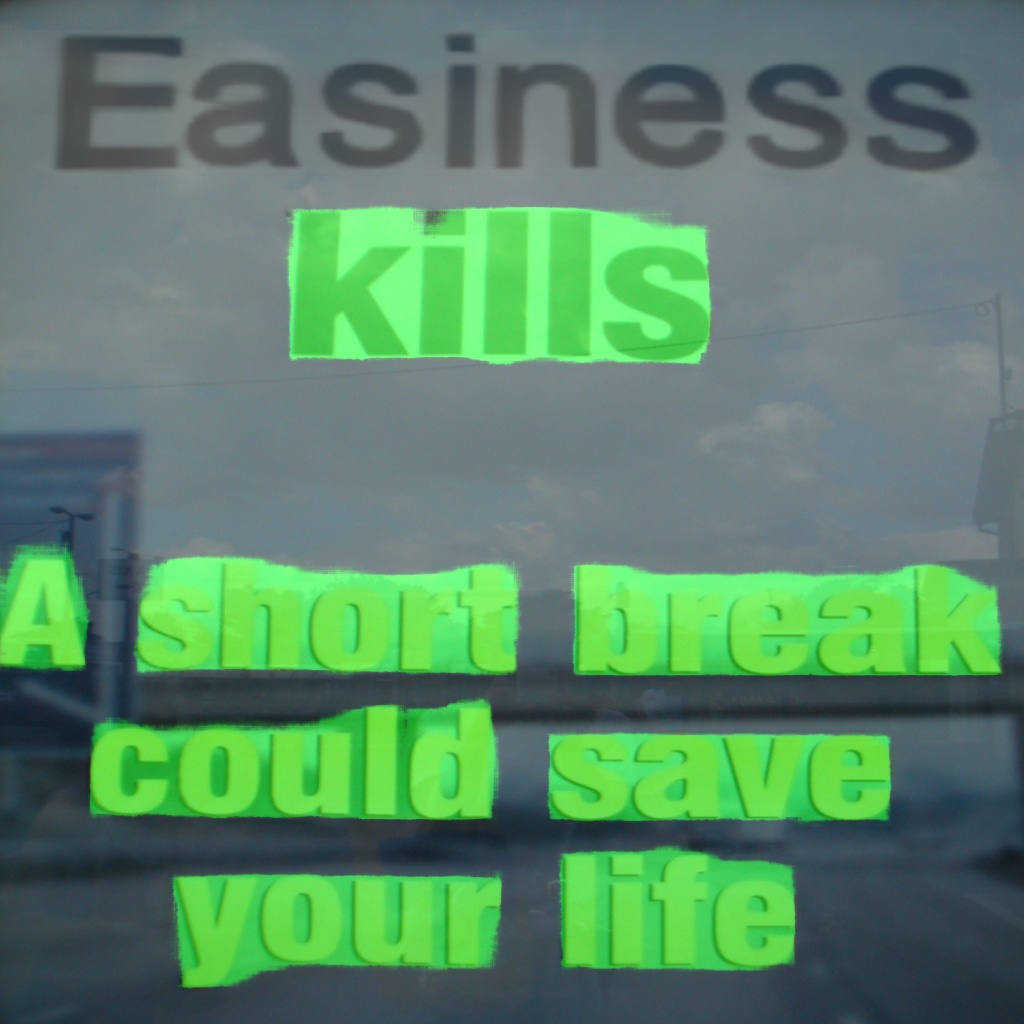}
        \end{minipage}%
    }%
    \subfigure{
        \begin{minipage}[c]{0.188\linewidth}
            \centering
            \includegraphics[width=\linewidth]{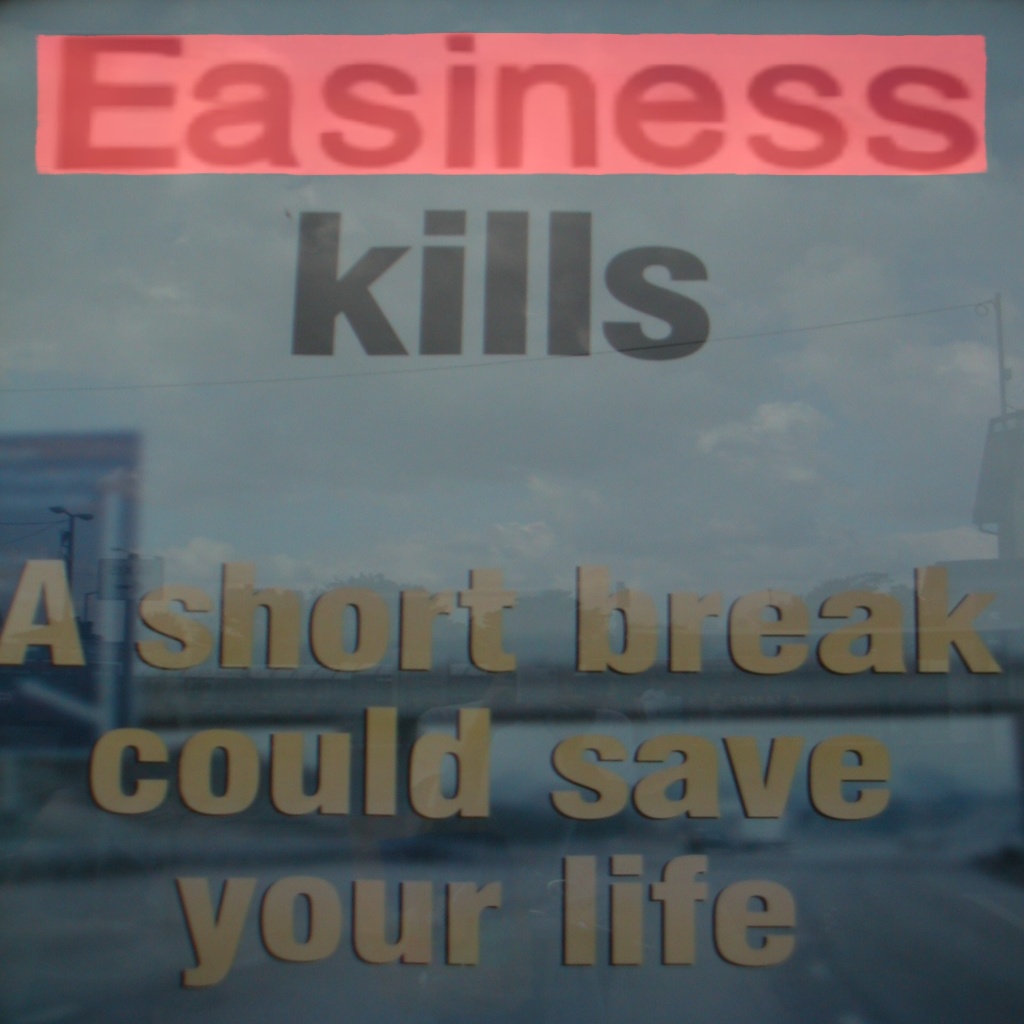}
        \end{minipage}%
    }%
    \subfigure{
        \begin{minipage}[c]{0.188\linewidth}
            \centering
            \includegraphics[width=\linewidth]{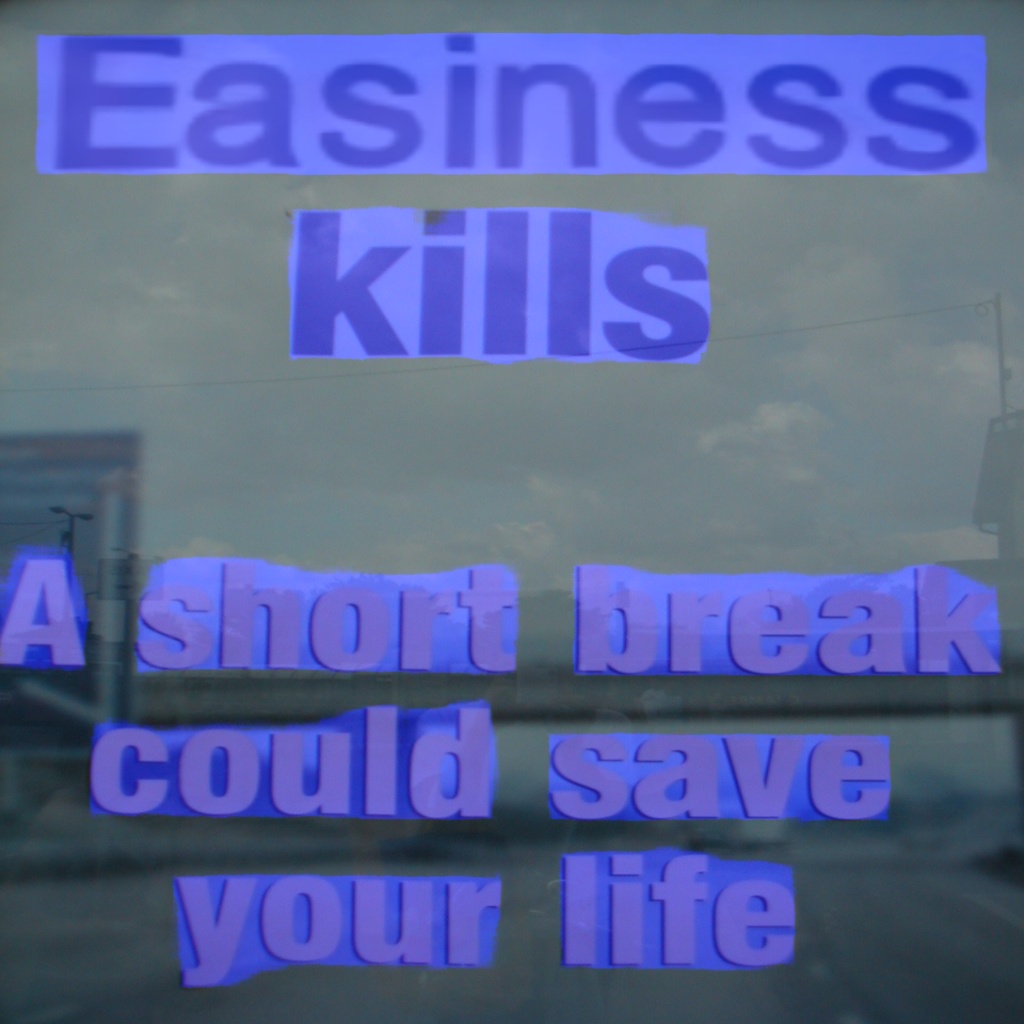}
        \end{minipage}%
    }
    \setcounter{subfigure}{0}
    \subfigure[Input]{
        \begin{minipage}[c]{0.188\linewidth}
            \centering
            \includegraphics[width=\linewidth,height=\linewidth]{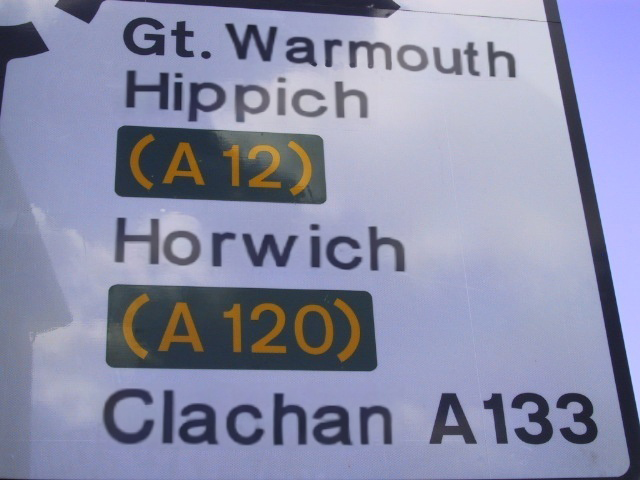}
        \end{minipage}%
    }%
    \subfigure[GT]{
        \begin{minipage}[c]{0.188\linewidth}
            \centering
            \includegraphics[width=\linewidth,height=\linewidth]{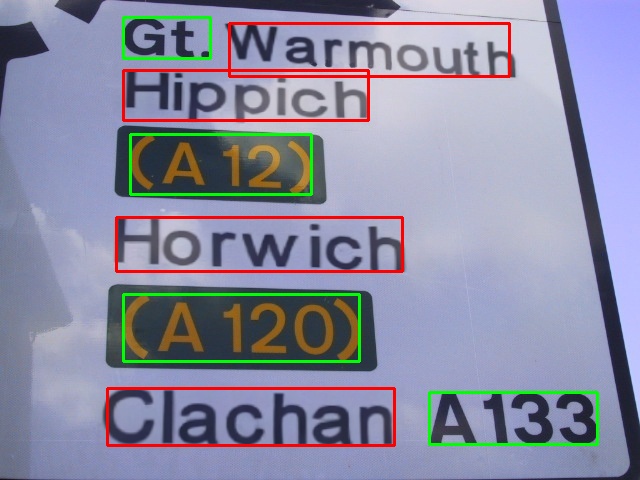}
        \end{minipage}%
    }%
    \subfigure[Real]{
        \begin{minipage}[c]{0.188\linewidth}
            \centering
            \includegraphics[width=\linewidth]{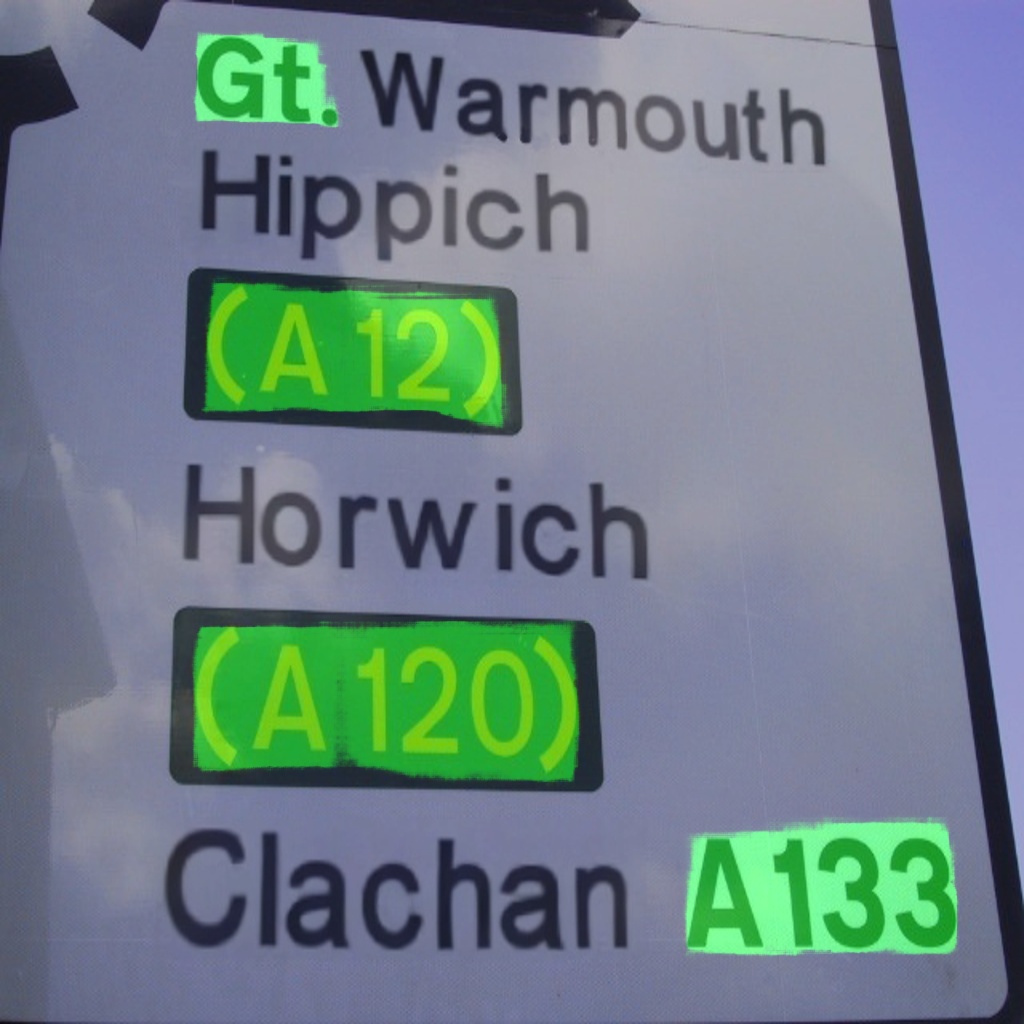}
        \end{minipage}%
    }%
    \subfigure[Tampered]{
        \begin{minipage}[c]{0.188\linewidth}
            \centering
            \includegraphics[width=\linewidth]{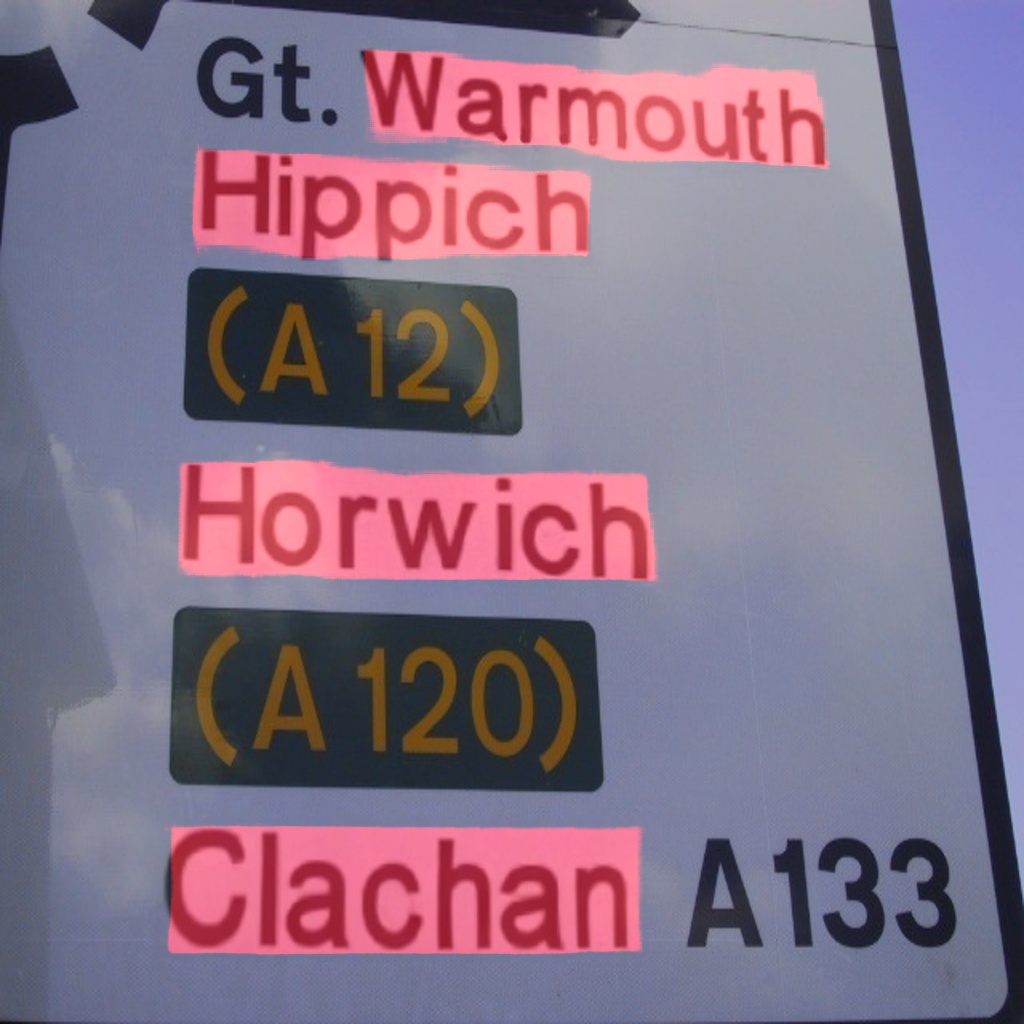}
        \end{minipage}%
    }%
    \subfigure[Text]{
        \begin{minipage}[c]{0.188\linewidth}
            \centering
            \includegraphics[width=\linewidth]{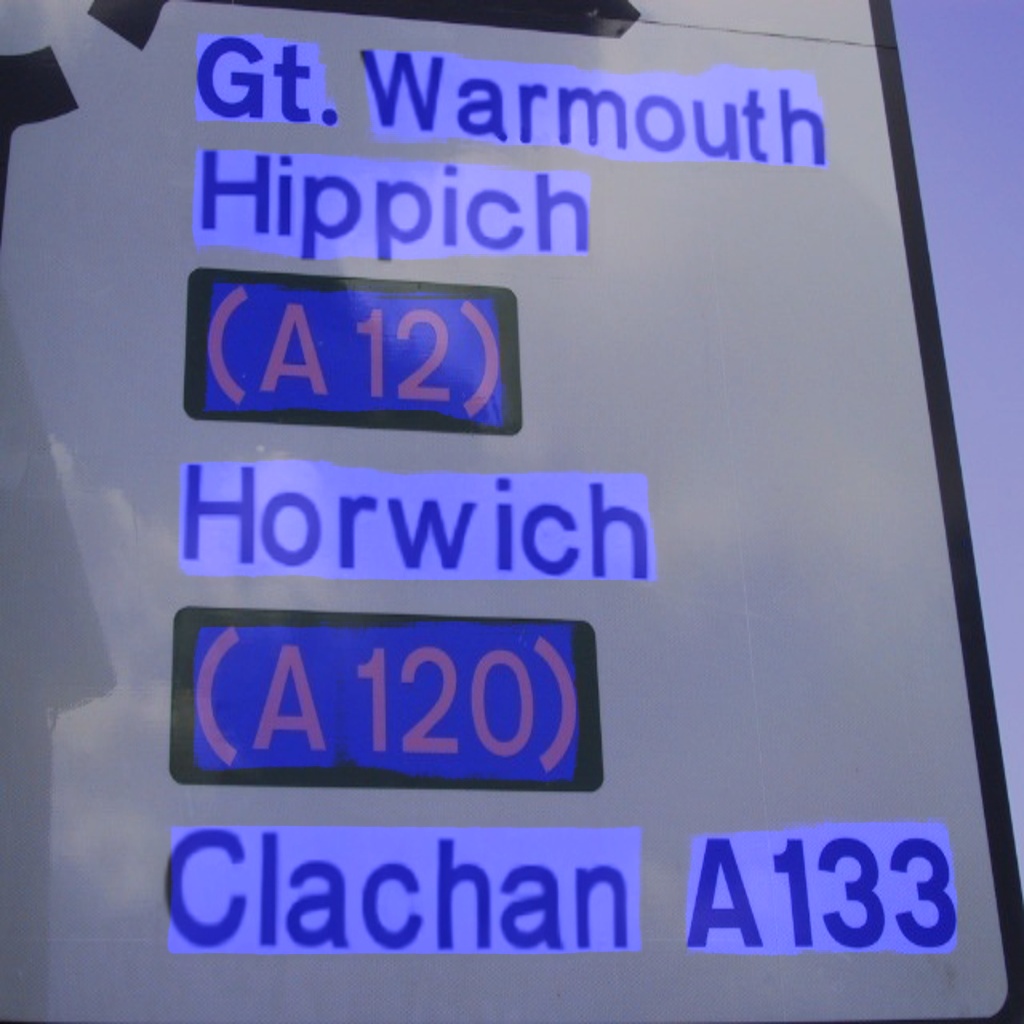}
        \end{minipage}%
    }
    \caption{Visualization results on Tampered-IC13. }
    \label{fig:exp_vis_tampered_ic13}
\end{figure}

\subsection{Extension to Tampered Scene Text Detection}
To verify the generalization ability of ViTEraser, we extend it to the tampered scene text detection (TSTD) task \cite{wang2022detecting} that aims to localize both tampered and real texts from natural scenes.
Using Tampered-IC13 dataset \cite{wang2022detecting}, we train a ViTEraser-Tiny (w/o SegMIM) whose three-channel outputs correspond to the box-level segmentation of real texts, tampered texts, and both of them, respectively.
The dice losses on these three segmentation maps are utilized to optimize the network.
Furthermore, to calculate the evaluation metrics including recall (R), precision (P), and f-measure (F) of tampered and real texts, we incorporate an EAST \cite{zhou2017east}, PSENet \cite{wang2019shape}, or ContourNet \cite{wang2020contournet} trained with Tampered-IC13 to produce text bounding boxes.
Specifically, a bounding box will be regarded as tampered if more than 50\% pixels within it are classified as tampered by ViTEraser. 
Similarly, the bounding boxes of real texts can also be determined.
The quantitative performance is presented in Tab.~\ref{tab:exp_tamperd_ic13} and the visualizations are shown in Fig.~\ref{fig:exp_vis_tampered_ic13}.
It can be seen that ViTEraser can achieve state-of-the-art performance on Tampered-IC13, showing strong generalization potential.

\section{Conclusion}

In this paper, we propose a novel simple-yet-effective one-stage ViT-based approach for STR, termed ViTEraser. 
ViTEraser employs a concise encoder-decoder paradigm, eliminating the need for text localizing modules, external text detectors, and progressive refinements.
Moreover, ViTEraser pioneers in thoroughly utilizing ViTs in place of CNNs in both the encoder and decoder, significantly enhancing the long-range modeling ability.
Furthermore, we propose a novel pretraining scheme, called SegMIM, 
which focuses the encoder and decoder on the text box segmentation and MIM tasks, respectively.
Without bells and whistles, the proposed method substantially outperforms previous STR approaches.
ViTEraser also exhibits outstanding performance in tampered scene text detection, exhibiting strong generalization potential.
Additionally, we comprehensively explore the architecture, pretraining, and scalability of ViT-based encoder-decoder for STR. 
We believe this study can inspire more research on ViT-based STR and contribute to the development of the unified model for pixel-level OCR tasks.

\section{Acknowledgements}
This research is supported in part by NSFC (Grant No.: 61936003), National Key Research and Development Program of China (2022YFC3301703), and INTSIG-SCUT Joint Lab Fundation (CG-0274-200703).

\bibliography{aaai24}

\vspace{20pt}
\appendix
\noindent\textbf{\LARGE Appendix}
\section{Training Details of ViTEraser}

During training, the ViTEraser is end-to-end optimized using a weighted sum of multi-scale reconstruction loss, perceptual loss, style loss, segmentation loss, and adversarial loss.

Before introducing training details, we first denote one training sample as $\{I_{in}, I_{gt}, M_{gt}\}$, where $I_{in} \in \mathbb{R}^{H \times W \times 3}$ is the input image with texts to remove, $I_{gt} \in \mathbb{R}^{H \times W \times 3}$ is the paired ground-truth (GT) image with all texts erased, and $M_{gt} \in \mathbb{R}^{H \times W \times 1}$ is the GT binary text box mask (0 for backgrounds and 1 for text regions).

Moreover, as described in the \textit{ViTEraser} section and illustrated in \textit{Figs.}~\ref{fig:viteraser} and \ref{fig:aux_output} of the \textit{main paper}, given the input image $I_{in} \in \mathbb{R}^{H \times W \times 3}$, ViTEraser produces mutli-scale text erasing results $\mathbb{I}_{out}$ and a text box segmentation map $M_{out} \in \mathbb{R}^{H \times W \times 1}$ during training.
Concretely, the $\mathbb{I}_{out}$ is defined as 
\begin{equation}
    \small
    \mathbb{I}_{out} = \{I_{out} \in \mathbb{R}^{H \times W \times 3}, I_{out}^{\frac{1}{2}} \in \mathbb{R}^{\frac{H}{2} \times \frac{W}{2} \times 3}, I_{out}^{\frac{1}{4}} \in \mathbb{R}^{\frac{H}{4} \times \frac{W}{4} \times 3}\}.
\end{equation}
Then the details of the losses adopted for training ViTEraser are described in the following sections.

\subsubsection{Multi-Scale Reconstruction Loss}
The \textit{L1} loss is employed to measure the discrepancy between the predicted and GT text-erased images. 
To weigh the reconstruction of text and non-text regions, the GT text box masks are utilized to endow the loss with text-aware ability.
Moreover, the reconstruction supervision is applied to the predictions at different levels of the decoder, so as to improve the multi-scale visual perception of the model.

Firstly, corresponding to the multi-scale text erasing results $\mathbb{I}_{out}$, the multi-scale GT text-erased images $\mathbb{I}_{gt}$ and GT text box mask $\mathbb{M}_{gt}$ is defined as 
\begin{equation}
    \small
    \mathbb{I}_{gt} = \{I_{gt} \in \mathbb{R}^{H \times W \times 3}, I_{gt}^{\frac{1}{2}} \in \mathbb{R}^{\frac{H}{2} \times \frac{W}{2} \times 3}, I_{gt}^{\frac{1}{4}} \in \mathbb{R}^{\frac{H}{4} \times \frac{W}{4} \times 3}\},
\end{equation}
\begin{equation}
    \small
    \mathbb{M}_{gt} = \{M_{gt} \in \mathbb{R}^{H \times W \times 1}, M_{gt}^{\frac{1}{2}} \in \mathbb{R}^{\frac{H}{2} \times \frac{W}{2} \times 1}, M_{gt}^{\frac{1}{4}} \in \mathbb{R}^{\frac{H}{4} \times \frac{W}{4} \times 1}\},
\end{equation}
where we resize the $I_{gt}$ and $M_{gt}$ to obtain the corresponding $\frac{1}{2}$ and $\frac{1}{4}$ versions.
Then, the multi-scale reconstruction loss $L_{msr}$ is formulated as
\begin{align}
    \mathcal{L}_{msr}  = & \sum\nolimits_{i=1}^{3} \lambda_{i} || (\mathbb{I}_{out(i)} - \mathbb{I}_{gt(i)}) \odot \mathbb{M}_{gt(i)} ||_{1} \notag \\
                         & + \beta_{i} || (\mathbb{I}_{out(i)} - \mathbb{I}_{gt(i)}) \odot (1 - \mathbb{M}_{gt(i)}) ||_{1},
\end{align}
where the $\odot$ calculates the element-wise product.
Besides, $\lambda$ and $\beta$ are empirically set to $\{10, 6, 5\}$ and $\{2, 1, 0.8\}$, respectively.

\subsubsection{Perceptual Loss}
The perceptual loss \cite{johnson2016perceptual} penalizes the differences between images from a human-like perceptual perspective.
A VGG-16 \cite{Simonyan15} network $\Phi$ pretrained on ImageNet \cite{deng2009imagenet} is employed and $\Phi_{i}(x)$ represent the feature map produced by the i\textit{-th} pooling layer with an input $x$.
Mathematically, the perceptual loss $\mathcal{L}_{per}$ is calculated as 
\begin{align}
    I_{out}^{*} = & I_{out} \odot M_{gt} + I_{in} \odot (1 - M_{gt}), \label{equ:I_out^*} \\
    \mathcal{L}_{per} = & \sum\nolimits_{i=1}^3 || \Phi_{i}(I_{out}) - \Phi_{i}(I_{gt}) ||_1 \notag \\
                        & + || \Phi_{i}(I_{out}^{*}) - \Phi_{i}(I_{gt}) ||_1 \ .
\end{align}

\subsubsection{Style Loss}
The style loss \cite{gatys2016image} constrains the style difference between images, which adopts a Gram matrix for high-level feature correlations. 
Using the same VGG-16-based $\Phi_{i}(\cdot)$ as the perceptual loss, the style loss $\mathcal{L}_{sty}$ is defined as 
\begin{align}
    \mathcal{L}_{sty} = & \sum\nolimits_{i=1}^{3} || Gram(\Phi_{i}(I_{out})) - Gram(\Phi_{i}(I_{gt})) ||_1 \notag \\
                        & + || Gram(\Phi_{i}(I_{out}^{*})) - Gram(\Phi_{i}(I_{gt})) ||_1, 
\end{align}
where $I_{out}^{*}$ is computed as Eq. (\ref{equ:I_out^*}) and $Gram(\cdot)$ function calculates the Gram matrix of the given activation map.

\subsubsection{Segmentation Loss}
The ViTEraser end-to-end produces text-erased images from the scene text image inputs, without the guidance of any forms of text locations. 
However, the perception of text locations is critical to the text removal.
Therefore, a segmentation loss $\mathcal{L}_{seg}$ is employed during training, with which the model learns to implicitly perceive text locations during inference. 
Specifically, $\mathcal{L}_{seg}$ is the dice loss between the predicted ($M_{out}$) and GT ($M_{gt}$) text box masks:
\begin{equation}
    \mathcal{L}_{seg} = 1 - \frac{2\sum_{i,j}M_{out(i,j)} \times M_{gt(i,j)}}{\sum_{i,j}(M_{out(i,j)})^2 + \sum_{i,j}(M_{gt(i,j)})^2}.
\end{equation}

\subsubsection{Adversarial Loss}
Adversarial training has been demonstrated effective for generating visually plausible content \cite{zhang2019ensnet,liu2020erasenet,liu2022don,du2022progressive,lyu2023fetnet}.
Inspired by EraseNet \cite{liu2020erasenet}, a global-local discriminator $D$ is devised in our model training. 
The network $D$ takes a real/fake text-erased image and a binary mask as input and generates a value within $(-1, 1)$ (1 for real, -1 for fake).
To achieve this, the final feature map of $D$ is first activated with a sigmoid function, then rescaled to the range of $(-1, 1)$, and finally averaged.
Given the discriminator $D$, the adversarial losses can be defined as
\begin{align}
    \mathcal{L}_{adv}^{D} = & max(0, 1 - D(I_{in}, M_{gt})) \notag \\
                            & + max(0, 1 + D(I_{out}, M_{gt})), \\    
    \mathcal{L}_{adv}^{G} = & - D(I_{out}, M_{gt}),
\end{align}
where $\mathcal{L}_{adv}^{D}$ and $\mathcal{L}_{adv}^{G}$ are used to optimize the discriminator $D$ and the generator (\textit{i.e.}, ViTEraser), respectively.

\begin{table*}[t]
    \centering 
    \resizebox{2\columnwidth}{!}{
    \begin{tabular}{c|c|c|c|c|c|c|c}
    \hline
     & Stage & Output Size & Layer Name & \multicolumn{1}{c|}{ViTEraser-PVT-Tiny} & \multicolumn{1}{c|}{ViTEraser-PVT-Small} & \multicolumn{1}{c|}{ViTEraser-PVT-Medium} & \multicolumn{1}{c}{ViTEraser-PVT-Large} \\
    \hline
    \hline
    \multirow{8}{*}[-4.5em]{\rotatebox{90}{Encoder}} & \multirow{2}{*}[-1em]{Stage 1} & \multirow{2}{*}[-1em]{\LARGE $\frac{H}{4} \times \frac{W}{4}$} & Patch Embedding & \multicolumn{4}{c}{$D_1^{enc} = 4$, $C_1^{enc} = 64$} \\
    \cline{4-8}
     & & & PVT block & $\left [ \begin{matrix} R_1^{enc} = 8 \\ H_1^{enc} = 1 \\ E_1^{enc} = 8 \\ \end{matrix} \right ] \times 2$ & $\left [ \begin{matrix} R_1^{enc} = 8 \\ H_1^{enc} = 1 \\ E_1^{enc} = 8 \\ \end{matrix} \right ] \times 3$ & $\left [ \begin{matrix} R_1^{enc} = 8 \\ H_1^{enc} = 1 \\ E_1^{enc} = 8 \\ \end{matrix} \right ] \times 3$ & $\left [ \begin{matrix} R_1^{enc} = 8 \\ H_1^{enc} = 1 \\ E_1^{enc} = 8 \\ \end{matrix} \right ] \times 3$ \\
    \cline{2-8}

     & \multirow{2}{*}[-1em]{Stage 2} & \multirow{2}{*}[-1em]{\LARGE $\frac{H}{8} \times \frac{W}{8}$} & Patch Embedding & \multicolumn{4}{c}{$D_2^{enc} = 2$, $C_2^{enc} = 128$} \\
    \cline{4-8}
     & & & PVT block & $\left [ \begin{matrix} R_2^{enc} = 4 \\ H_2^{enc} = 2 \\ E_2^{enc} = 8 \\ \end{matrix} \right ] \times 2$ & $\left [ \begin{matrix} R_2^{enc} = 4 \\ H_2^{enc} = 2 \\ E_2^{enc} = 8 \\ \end{matrix} \right ] \times 3$ & $\left [ \begin{matrix} R_2^{enc} = 4 \\ H_2^{enc} = 2 \\ E_2^{enc} = 8 \\ \end{matrix} \right ] \times 3$ & $\left [ \begin{matrix} R_2^{enc} = 4 \\ H_2^{enc} = 2 \\ E_2^{enc} = 8 \\ \end{matrix} \right ] \times 8$ \\
    \cline{2-8}

     & \multirow{2}{*}[-1em]{Stage 3} & \multirow{2}{*}[-1em]{\LARGE $\frac{H}{16} \times \frac{W}{16}$} & Patch Embedding & \multicolumn{4}{c}{$D_3^{enc} = 2$, $C_3^{enc} = 320$} \\
    \cline{4-8}
     & & & PVT block & $\left [ \begin{matrix} R_3^{enc} = 2 \\ H_3^{enc} = 5 \\ E_3^{enc} = 4 \\ \end{matrix} \right ] \times 2$ & $\left [ \begin{matrix} R_3^{enc} = 2 \\ H_3^{enc} = 5 \\ E_3^{enc} = 4 \\ \end{matrix} \right ] \times 6$ & $\left [ \begin{matrix} R_3^{enc} = 2 \\ H_3^{enc} = 5 \\ E_3^{enc} = 4 \\ \end{matrix} \right ] \times 18$ & $\left [ \begin{matrix} R_3^{enc} = 2 \\ H_3^{enc} = 5 \\ E_3^{enc} = 4 \\ \end{matrix} \right ] \times 27$ \\
    \cline{2-8}

     & \multirow{2}{*}[-1em]{Stage 4} & \multirow{2}{*}[-1em]{\LARGE $\frac{H}{32} \times \frac{W}{32}$} & Patch Embedding & \multicolumn{4}{c}{$D_4^{enc} = 2$, $C_4^{enc} = 512$} \\
    \cline{4-8}
     & & & PVT block & $\left [ \begin{matrix} R_4^{enc} = 1 \\ H_4^{enc} = 8 \\ E_4^{enc} = 4 \\ \end{matrix} \right ] \times 2$ & $\left [ \begin{matrix} R_4^{enc} = 1 \\ H_4^{enc} = 8 \\ E_4^{enc} = 4 \\ \end{matrix} \right ] \times 3$ & $\left [ \begin{matrix} R_4^{enc} = 1 \\ H_4^{enc} = 8 \\ E_4^{enc} = 4 \\ \end{matrix} \right ] \times 3$ & $\left [ \begin{matrix} R_4^{enc} = 1 \\ H_4^{enc} = 8 \\ E_4^{enc} = 4 \\ \end{matrix} \right ] \times 3$ \\
    \hline
    \hline

    \multirow{10}{*}[-5.5em]{\rotatebox{90}{Decoder}} & \multirow{2}{*}{Stage 1} & \multirow{2}{*}{\LARGE $\frac{H}{16} \times \frac{W}{16}$} & PVT block & $\left [ \begin{matrix} R_1^{dec} = 1 \\ H_1^{dec} = 8 \\ E_1^{dec} = 4 \\ \end{matrix} \right ] \times 2$ & $\left [ \begin{matrix} R_1^{dec} = 1 \\ H_1^{dec} = 8 \\ E_1^{dec} = 4 \\ \end{matrix} \right ] \times 3$ & $\left [ \begin{matrix} R_1^{dec} = 1 \\ H_1^{dec} = 8 \\ E_1^{dec} = 4 \\ \end{matrix} \right ] \times 3$ & $\left [ \begin{matrix} R_1^{dec} = 1 \\ H_1^{dec} = 8 \\ E_1^{dec} = 4 \\ \end{matrix} \right ] \times 3$ \\
    \cline{4-8}
     & & & Patch Splitting & \multicolumn{4}{c}{$U_1^{dec} = 2$, $C_1^{dec} = 320$} \\
    \cline{2-8}

    & \multirow{2}{*}{Stage 2} & \multirow{2}{*}{\LARGE $\frac{H}{8} \times \frac{W}{8}$} & PVT block & $\left [ \begin{matrix} R_2^{dec} = 2 \\ H_2^{dec} = 5 \\ E_2^{dec} = 4 \\ \end{matrix} \right ] \times 2$ & $\left [ \begin{matrix} R_2^{dec} = 2 \\ H_2^{dec} = 5 \\ E_2^{dec} = 4 \\ \end{matrix} \right ] \times 6$ & $\left [ \begin{matrix} R_2^{dec} = 2 \\ H_2^{dec} = 5 \\ E_2^{dec} = 4 \\ \end{matrix} \right ] \times 18$ & $\left [ \begin{matrix} R_2^{dec} = 2 \\ H_2^{dec} = 5 \\ E_2^{dec} = 4 \\ \end{matrix} \right ] \times 27$ \\
    \cline{4-8}
     & & & Patch Splitting & \multicolumn{4}{c}{$U_2^{dec} = 2$, $C_2^{dec} = 128$} \\
    \cline{2-8}

    & \multirow{2}{*}{Stage 3} & \multirow{2}{*}{\LARGE $\frac{H}{4} \times \frac{W}{4}$} & PVT block & $\left [ \begin{matrix} R_3^{dec} = 4 \\ H_3^{dec} = 2 \\ E_3^{dec} = 8 \\ \end{matrix} \right ] \times 2$ & $\left [ \begin{matrix} R_3^{dec} = 4 \\ H_3^{dec} = 2 \\ E_3^{dec} = 8 \\ \end{matrix} \right ] \times 3$ & $\left [ \begin{matrix} R_3^{dec} = 4 \\ H_3^{dec} = 2 \\ E_3^{dec} = 8 \\ \end{matrix} \right ] \times 3$ & $\left [ \begin{matrix} R_3^{dec} = 4 \\ H_3^{dec} = 2 \\ E_3^{dec} = 8 \\ \end{matrix} \right ] \times 8$ \\
    \cline{4-8}
     & & & Patch Splitting & \multicolumn{4}{c}{$U_3^{dec} = 2$, $C_3^{dec} = 64$} \\
    \cline{2-8}

    & \multirow{2}{*}{Stage 4} & \multirow{2}{*}{\LARGE $\frac{H}{2} \times \frac{W}{2}$} & PVT block & $\left [ \begin{matrix} R_4^{dec} = 8 \\ H_4^{dec} = 1 \\ E_4^{dec} = 8 \\ \end{matrix} \right ] \times 2$ & $\left [ \begin{matrix} R_4^{dec} = 8 \\ H_4^{dec} = 1 \\ E_4^{dec} = 8 \\ \end{matrix} \right ] \times 3$ & $\left [ \begin{matrix} R_4^{dec} = 8 \\ H_4^{dec} = 1 \\ E_4^{dec} = 8 \\ \end{matrix} \right ] \times 3$ & $\left [ \begin{matrix} R_4^{dec} = 8 \\ H_4^{dec} = 1 \\ E_4^{dec} = 8 \\ \end{matrix} \right ] \times 3$ \\
    \cline{4-8}
     & & & Patch Splitting & \multicolumn{4}{c}{$U_4^{dec} = 2$, $C_4^{dec} = 32$} \\
    \cline{2-8}

    & \multirow{2}{*}{Stage 5} & \multirow{2}{*}{\large $H \times W$} & PVT block & $\left [ \begin{matrix} R_5^{dec} = 16 \\ H_5^{dec} = 1 \\ E_5^{dec} = 8 \\ \end{matrix} \right ] \times 2$ & $\left [ \begin{matrix} R_5^{dec} = 16 \\ H_5^{dec} = 1 \\ E_5^{dec} = 8 \\ \end{matrix} \right ] \times 2$ & $\left [ \begin{matrix} R_5^{dec} = 16 \\ H_5^{dec} = 1 \\ E_5^{dec} = 8 \\ \end{matrix} \right ] \times 2$ & $\left [ \begin{matrix} R_5^{dec} = 16 \\ H_5^{dec} = 1 \\ E_5^{dec} = 8 \\ \end{matrix} \right ] \times 2$ \\
    \cline{4-8}
     & & & Patch Splitting & \multicolumn{4}{c}{$U_5^{dec} = 2$, $C_5^{dec} = 32$} \\
    \hline

    \end{tabular}}
    \caption{Detailed network architectures of ViTEraser-PVT-Tiny, Small, Medium, and Large.}
    \label{tab:arch_pvt}
\end{table*}

\begin{table*}[t]
    \centering 
    \resizebox{1.96\columnwidth}{!}{
    \begin{tabular}{c|c|c|c|c|c|c}
    \hline
    & Stage & Output Size & Layer Name & ViTEraser-Swin-Tiny & ViTEraser-Swin-Small & ViTEraser-Swin-Base  \\
    \hline 
    \hline
    \multirow{8}{*}[-2.5em]{\rotatebox{90}{Encoder}} & \multirow{2}{*}[-0.75em]{Stage 1} & \multirow{2}{*}[-0.75em]{\Large $\frac{H}{4} \times \frac{W}{4}$} & Patch Embedding & $D_1^{enc} = 4$, $C_1^{enc} = 96$ & $D_1^{enc} = 4$, $C_1^{enc} = 96$ & $D_1^{enc} = 4$, $C_1^{enc} = 128$\\
    \cline{4-7}
    & & & Swin block & $\left [ \begin{matrix} W_1^{enc} = 7 \\ H_1^{enc} = 3 \end{matrix} \right ] \times 2$ & $\left [ \begin{matrix} W_1^{enc} = 7 \\ H_1^{enc} = 3 \end{matrix} \right ] \times 2$ & $\left [ \begin{matrix} W_1^{enc} = 7 \\ H_1^{enc} = 4 \end{matrix} \right ] \times 2$ \\
    \cline{2-7}

    & \multirow{2}{*}[-0.75em]{Stage 2} & \multirow{2}{*}[-0.75em]{\Large $\frac{H}{8} \times \frac{W}{8}$} & Patch Embedding & $D_2^{enc} = 2$, $C_2^{enc} = 192$ & $D_2^{enc} = 2$, $C_2^{enc} = 192$ & $D_2^{enc} = 2$, $C_2^{enc} = 256$\\
    \cline{4-7}
    & & & Swin block & $\left [ \begin{matrix} W_2^{enc} = 7 \\ H_2^{enc} = 6 \end{matrix} \right ] \times 2$ & $\left [ \begin{matrix} W_2^{enc} = 7 \\ H_2^{enc} = 6 \end{matrix} \right ] \times 2$ & $\left [ \begin{matrix} W_2^{enc} = 7 \\ H_2^{enc} = 8 \end{matrix} \right ] \times 2$ \\
    \cline{2-7}

    & \multirow{2}{*}[-0.75em]{Stage 3} & \multirow{2}{*}[-0.75em]{\Large $\frac{H}{16} \times \frac{W}{16}$} & Patch Embedding & $D_3^{enc} = 2$, $C_3^{enc} = 384$ & $D_3^{enc} = 2$, $C_3^{enc} = 384$ & $D_3^{enc} = 2$, $C_3^{enc} = 512$\\
    \cline{4-7}
    & & & Swin block & $\left [ \begin{matrix} W_3^{enc} = 7 \\ H_3^{enc} = 12 \end{matrix} \right ] \times 6$ & $\left [ \begin{matrix} W_3^{enc} = 7 \\ H_3^{enc} = 12 \end{matrix} \right ] \times 18$ & $\left [ \begin{matrix} W_3^{enc} = 7 \\ H_3^{enc} = 16 \end{matrix} \right ] \times 18$ \\
    \cline{2-7}

    & \multirow{2}{*}[-0.75em]{Stage 4} & \multirow{2}{*}[-0.75em]{\Large $\frac{H}{32} \times \frac{W}{32}$} & Patch Embedding & $D_4^{enc} = 2$, $C_4^{enc} = 768$ & $D_4^{enc} = 2$, $C_4^{enc} = 768$ & $D_4^{enc} = 2$, $C_4^{enc} = 1024$\\
    \cline{4-7}
    & & & Swin block & $\left [ \begin{matrix} W_4^{enc} = 7 \\ H_4^{enc} = 24 \end{matrix} \right ] \times 2$ & $\left [ \begin{matrix} W_4^{enc} = 7 \\ H_4^{enc} = 24 \end{matrix} \right ] \times 2$ & $\left [ \begin{matrix} W_4^{enc} = 7 \\ H_4^{enc} = 32 \end{matrix} \right ] \times 2$ \\
    \hline
    \hline

    \multirow{8}{*}[-4em]{\rotatebox{90}{Decoder}} & \multirow{2}{*}{Stage 1} & \multirow{2}{*}{\Large $\frac{H}{16} \times \frac{W}{16}$} & Swin block & $\left [ \begin{matrix} W_1^{dec} = 7 \\ H_1^{dec} = 24 \end{matrix} \right ] \times 2$ & $\left [ \begin{matrix} W_1^{dec} = 7 \\ H_1^{dec} = 24 \end{matrix} \right ] \times 2$ & $\left [ \begin{matrix} W_1^{dec} = 7 \\ H_1^{dec} = 32 \end{matrix} \right ] \times 2$ \\
    \cline{4-7}
    & & & Patch Splitting & $U_1^{dec} = 2$, $C_1^{dec} = 384$ & $U_1^{dec} = 2$, $C_1^{dec} = 384$ & $U_1^{dec} = 2$, $C_1^{dec} = 512$\\
    \cline{2-7}

    & \multirow{2}{*}{Stage 2} & \multirow{2}{*}{\Large $\frac{H}{8} \times \frac{W}{8}$} & Swin block & $\left [ \begin{matrix} W_2^{dec} = 7 \\ H_2^{dec} = 12 \end{matrix} \right ] \times 6$ & $\left [ \begin{matrix} W_2^{dec} = 7 \\ H_2^{dec} = 12 \end{matrix} \right ] \times 18$ & $\left [ \begin{matrix} W_2^{dec} = 7 \\ H_2^{dec} = 16 \end{matrix} \right ] \times 18$ \\
    \cline{4-7}
    & & & Patch Splitting & $U_2^{dec} = 2$, $C_2^{dec} = 192$ & $U_2^{dec} = 2$, $C_2^{dec} = 192$ & $U_2^{dec} = 2$, $C_2^{dec} = 256$\\
    \cline{2-7}

    & \multirow{2}{*}{Stage 3} & \multirow{2}{*}{\Large $\frac{H}{4} \times \frac{W}{4}$} & Swin block & $\left [ \begin{matrix} W_3^{dec} = 7 \\ H_3^{dec} = 6 \end{matrix} \right ] \times 2$ & $\left [ \begin{matrix} W_3^{dec} = 7 \\ H_3^{dec} = 6 \end{matrix} \right ] \times 2$ & $\left [ \begin{matrix} W_3^{dec} = 7 \\ H_3^{dec} = 8 \end{matrix} \right ] \times 2$ \\
    \cline{4-7}
    & & & Patch Splitting & $U_3^{dec} = 2$, $C_3^{dec} = 96$ & $U_3^{dec} = 2$, $C_3^{dec} = 96$ & $U_3^{dec} = 2$, $C_3^{dec} = 128$\\
    \cline{2-7}

    & \multirow{2}{*}{Stage 4} & \multirow{2}{*}{\Large $\frac{H}{2} \times \frac{W}{2}$} & Swin block & $\left [ \begin{matrix} W_4^{dec} = 7 \\ H_4^{dec} = 3 \end{matrix} \right ] \times 2$ & $\left [ \begin{matrix} W_4^{dec} = 7 \\ H_4^{dec} = 3 \end{matrix} \right ] \times 2$ & $\left [ \begin{matrix} W_4^{dec} = 7 \\ H_4^{dec} = 4 \end{matrix} \right ] \times 2$ \\
    \cline{4-7}
    & & & Patch Splitting & $U_4^{dec} = 2$, $C_4^{dec} = 48$ & $U_4^{dec} = 2$, $C_4^{dec} = 48$ & $U_4^{dec} = 2$, $C_4^{dec} = 64$\\
    \cline{2-7}

    & \multirow{2}{*}{Stage 5} & \multirow{2}{*}{$H \times W$} & Swin block & $\left [ \begin{matrix} W_5^{dec} = 7 \\ H_5^{dec} = 2 \end{matrix} \right ] \times 2$ & $\left [ \begin{matrix} W_5^{dec} = 7 \\ H_5^{dec} = 2 \end{matrix} \right ] \times 2$ & $\left [ \begin{matrix} W_5^{dec} = 7 \\ H_5^{dec} = 2 \end{matrix} \right ] \times 2$ \\
    \cline{4-7}
    & & & Patch Splitting & $U_5^{dec} = 2$, $C_5^{dec} = 24$ & $U_5^{dec} = 2$, $C_5^{dec} = 24$ & $U_5^{dec} = 2$, $C_5^{dec} = 32$\\
    \hline

    \end{tabular}}
    \caption{Detailed network architectures of ViTEraser-Swin-Tiny, Small, and Base.}
    \label{tab:arch_swin}
\end{table*}

\subsubsection{Total Loss}
\label{sec:viteraser_total_loss}
The total loss $\mathcal{L}$ for ViTEraser is the weighted sum of the above losses, which is given by
\begin{equation}
    \mathcal{L} = \alpha_{msr} \mathcal{L}_{msr} + \alpha_{per} \mathcal{L}_{per} + \alpha_{sty} \mathcal{L}_{sty} + 
                       \alpha_{seg} \mathcal{L}_{seg} + \alpha_{adv} \mathcal{L}_{adv}^{G},
\end{equation}
where the weights $\alpha_{msr}$, $\alpha_{per}$, $\alpha_{sty}$, $\alpha_{seg}$, and $\alpha_{adv}$ are empirically set to 1, 0.01, 120, 1, and 0.1, respectively.

\section{Detailed Network Architecture}

As described in the \textit{ViTEraser} section of the \textit{main paper}, the detailed network architectures of ViTEraser involve the following hyper-parameters.
\begin{itemize}
    \item $C_i^{enc}$: the number of channels of the i\textit{-th} stage of the encoder, which is equal to the number of output channels of the patch embedding layer in this stage.
    \item $N_i^{enc}$: the number of ViT blocks in the i\textit{-th} stage of the encoder.
    \item $H_i^{enc}$: the number of heads of the ViT blocks in the i\textit{-th} stage of the encoder.
    \item $D_i^{enc}$: the downsampling ratio (\textit{i.e.}, patch size) of the patch embedding layer in the i\textit{-th} stage of the encoder.
    \item $C_i^{dec}$: the number of channels of the i\textit{-th} stage of the decoder, which is equal to the number of output channels of the patch splitting layer in this stage.
    \item $N_i^{dec}$: the number of ViT blocks in the i\textit{-th} stage of the decoder.
    \item $H_i^{dec}$: the number of heads of the ViT blocks in the i\textit{-th} stage of the decoder.
    \item $U_i^{dec}$: the upsampling ratio of the patch splitting layer in the i\textit{-th} stage of the decoder.
\end{itemize}

Furthermore, we explore three prevalent ViT blocks to implement ViTEraser, including Pyramid Vision Transformer (\textbf{PVT}) block \cite{wang2021pyramid}, Swin Transformer (\textbf{Swin}) block \cite{liu2021swin}, and Swin Transformer v2 (\textbf{Swinv2}) block \cite{liu2022swin}.
The details regarding different ViT blocks are introduced in the following sections.

\subsubsection{PVT-based ViTEraser}
Following \citet{wang2021pyramid}, the PVT-based ViTEraser involves additional hyper-parameters for network architecture as follows.
\begin{itemize}
    \item $R_i^{enc}$: the reduction ratio of the Spatial-Reduction Attention (SRA) in the PVT blocks of the i\textit{-th} stage of the encoder.
    \item $E_i^{enc}$: the expansion ratio of the feed-forward layer in the PVT blocks of the i\textit{-th} stage of the encoder.
    \item $R_i^{dec}$: the reduction ratio of the SRA in the PVT blocks of the i\textit{-th} stage of the decoder.
    \item $E_i^{dec}$: the expansion ratio of the feed-forward layer in the PVT blocks of the i\textit{-th} stage of the decoder.
\end{itemize}
Then the detailed architectures of ViTEraser-PVT-Tiny, Small, Medium, and Large are presented in Tab.~\ref{tab:arch_pvt}.

\subsubsection{Swin/Swinv2-based ViTEraser}
Following \citet{liu2021swin,liu2022swin}, the Swin/Swinv2-based ViTEraser involves additional hyper-parameters for network architectures as follows.
\begin{itemize}
    \item $W_i^{enc}$: the window size of the Swin/Swinv2 blocks in the i\textit{-th} stage of the encoder.
    \item $W_i^{dec}$: the window size of the Swin/Swinv2 blocks in the i\textit{-th} stage of the decoder.
\end{itemize}
Then the detailed network architectures of ViTEraser-Swin-Tiny, Small, and Base are provided in Tab.~\ref{tab:arch_swin}, and the detailed network architectures of ViTEraser-Swinv2-Tiny, Small, and Base are presented in Tab.~\ref{tab:arch_swinv2}.

\begin{table*}[t]
    \centering 
    \resizebox{1.96\columnwidth}{!}{
    \begin{tabular}{c|c|c|c|c|c|c}
    \hline
    & Stage & Output Size & Layer Name & ViTEraser-Swinv2-Tiny & ViTEraser-Swinv2-Small & ViTEraser-Swinv2-Base  \\
    \hline 
    \hline
    \multirow{8}{*}[-2.5em]{\rotatebox{90}{Encoder}} & \multirow{2}{*}[-0.75em]{Stage 1} & \multirow{2}{*}[-0.75em]{\Large $\frac{H}{4} \times \frac{W}{4}$} & Patch Embedding & $D_1^{enc} = 4$, $C_1^{enc} = 96$ & $D_1^{enc} = 4$, $C_1^{enc} = 96$ & $D_1^{enc} = 4$, $C_1^{enc} = 128$\\
    \cline{4-7}
    & & & Swinv2 block & $\left [ \begin{matrix} W_1^{enc} = 16 \\ H_1^{enc} = 3 \end{matrix} \right ] \times 2$ & $\left [ \begin{matrix} W_1^{enc} = 16 \\ H_1^{enc} = 3 \end{matrix} \right ] \times 2$ & $\left [ \begin{matrix} W_1^{enc} = 8 \\ H_1^{enc} = 4 \end{matrix} \right ] \times 2$ \\
    \cline{2-7}

    & \multirow{2}{*}[-0.75em]{Stage 2} & \multirow{2}{*}[-0.75em]{\Large $\frac{H}{8} \times \frac{W}{8}$} & Patch Embedding & $D_2^{enc} = 2$, $C_2^{enc} = 192$ & $D_2^{enc} = 2$, $C_2^{enc} = 192$ & $D_2^{enc} = 2$, $C_2^{enc} = 256$\\
    \cline{4-7}
    & & & Swinv2 block & $\left [ \begin{matrix} W_2^{enc} = 16 \\ H_2^{enc} = 6 \end{matrix} \right ] \times 2$ & $\left [ \begin{matrix} W_2^{enc} = 16 \\ H_2^{enc} = 6 \end{matrix} \right ] \times 2$ & $\left [ \begin{matrix} W_2^{enc} = 8 \\ H_2^{enc} = 8 \end{matrix} \right ] \times 2$ \\
    \cline{2-7}

    & \multirow{2}{*}[-0.75em]{Stage 3} & \multirow{2}{*}[-0.75em]{\Large $\frac{H}{16} \times \frac{W}{16}$} & Patch Embedding & $D_3^{enc} = 2$, $C_3^{enc} = 384$ & $D_3^{enc} = 2$, $C_3^{enc} = 384$ & $D_3^{enc} = 2$, $C_3^{enc} = 512$\\
    \cline{4-7}
    & & & Swinv2 block & $\left [ \begin{matrix} W_3^{enc} = 16 \\ H_3^{enc} = 12 \end{matrix} \right ] \times 6$ & $\left [ \begin{matrix} W_3^{enc} = 16 \\ H_3^{enc} = 12 \end{matrix} \right ] \times 18$ & $\left [ \begin{matrix} W_3^{enc} = 8 \\ H_3^{enc} = 16 \end{matrix} \right ] \times 18$ \\
    \cline{2-7}

    & \multirow{2}{*}[-0.75em]{Stage 4} & \multirow{2}{*}[-0.75em]{\Large $\frac{H}{32} \times \frac{W}{32}$} & Patch Embedding & $D_4^{enc} = 2$, $C_4^{enc} = 768$ & $D_4^{enc} = 2$, $C_4^{enc} = 768$ & $D_4^{enc} = 2$, $C_4^{enc} = 1024$\\
    \cline{4-7}
    & & & Swinv2 block & $\left [ \begin{matrix} W_4^{enc} = 16 \\ H_4^{enc} = 24 \end{matrix} \right ] \times 2$ & $\left [ \begin{matrix} W_4^{enc} = 16 \\ H_4^{enc} = 24 \end{matrix} \right ] \times 2$ & $\left [ \begin{matrix} W_4^{enc} = 8 \\ H_4^{enc} = 32 \end{matrix} \right ] \times 2$ \\
    \hline
    \hline

    \multirow{8}{*}[-4em]{\rotatebox{90}{Decoder}} & \multirow{2}{*}{Stage 1} & \multirow{2}{*}{\Large $\frac{H}{16} \times \frac{W}{16}$} & Swinv2 block & $\left [ \begin{matrix} W_1^{dec} = 16 \\ H_1^{dec} = 24 \end{matrix} \right ] \times 2$ & $\left [ \begin{matrix} W_1^{dec} = 8 \\ H_1^{dec} = 24 \end{matrix} \right ] \times 2$ & $\left [ \begin{matrix} W_1^{dec} = 8 \\ H_1^{dec} = 32 \end{matrix} \right ] \times 2$ \\
    \cline{4-7}
    & & & Patch Splitting & $U_1^{dec} = 2$, $C_1^{dec} = 384$ & $U_1^{dec} = 2$, $C_1^{dec} = 384$ & $U_1^{dec} = 2$, $C_1^{dec} = 512$\\
    \cline{2-7}

    & \multirow{2}{*}{Stage 2} & \multirow{2}{*}{\Large $\frac{H}{8} \times \frac{W}{8}$} & Swinv2 block & $\left [ \begin{matrix} W_2^{dec} = 16 \\ H_2^{dec} = 12 \end{matrix} \right ] \times 6$ & $\left [ \begin{matrix} W_2^{dec} = 8 \\ H_2^{dec} = 12 \end{matrix} \right ] \times 18$ & $\left [ \begin{matrix} W_2^{dec} = 8 \\ H_2^{dec} = 16 \end{matrix} \right ] \times 18$ \\
    \cline{4-7}
    & & & Patch Splitting & $U_2^{dec} = 2$, $C_2^{dec} = 192$ & $U_2^{dec} = 2$, $C_2^{dec} = 192$ & $U_2^{dec} = 2$, $C_2^{dec} = 256$\\
    \cline{2-7}

    & \multirow{2}{*}{Stage 3} & \multirow{2}{*}{\Large $\frac{H}{4} \times \frac{W}{4}$} & Swinv2 block & $\left [ \begin{matrix} W_3^{dec} = 16 \\ H_3^{dec} = 6 \end{matrix} \right ] \times 2$ & $\left [ \begin{matrix} W_3^{dec} = 8 \\ H_3^{dec} = 6 \end{matrix} \right ] \times 2$ & $\left [ \begin{matrix} W_3^{dec} = 8 \\ H_3^{dec} = 8 \end{matrix} \right ] \times 2$ \\
    \cline{4-7}
    & & & Patch Splitting & $U_3^{dec} = 2$, $C_3^{dec} = 96$ & $U_3^{dec} = 2$, $C_3^{dec} = 96$ & $U_3^{dec} = 2$, $C_3^{dec} = 128$\\
    \cline{2-7}

    & \multirow{2}{*}{Stage 4} & \multirow{2}{*}{\Large $\frac{H}{2} \times \frac{W}{2}$} & Swinv2 block & $\left [ \begin{matrix} W_4^{dec} = 16 \\ H_4^{dec} = 3 \end{matrix} \right ] \times 2$ & $\left [ \begin{matrix} W_4^{dec} = 8 \\ H_4^{dec} = 3 \end{matrix} \right ] \times 2$ & $\left [ \begin{matrix} W_4^{dec} = 8 \\ H_4^{dec} = 4 \end{matrix} \right ] \times 2$ \\
    \cline{4-7}
    & & & Patch Splitting & $U_4^{dec} = 2$, $C_4^{dec} = 48$ & $U_4^{dec} = 2$, $C_4^{dec} = 48$ & $U_4^{dec} = 2$, $C_4^{dec} = 64$\\
    \cline{2-7}

    & \multirow{2}{*}{Stage 5} & \multirow{2}{*}{$H \times W$} & Swinv2 block & $\left [ \begin{matrix} W_5^{dec} = 16 \\ H_5^{dec} = 2 \end{matrix} \right ] \times 2$ & $\left [ \begin{matrix} W_5^{dec} = 8 \\ H_5^{dec} = 2 \end{matrix} \right ] \times 2$ & $\left [ \begin{matrix} W_5^{dec} = 8 \\ H_5^{dec} = 2 \end{matrix} \right ] \times 2$ \\
    \cline{4-7}
    & & & Patch Splitting & $U_5^{dec} = 2$, $C_5^{dec} = 24$ & $U_5^{dec} = 2$, $C_5^{dec} = 24$ & $U_5^{dec} = 2$, $C_5^{dec} = 32$\\
    \hline

    \end{tabular}}
    \caption{Detailed network architectures of ViTEraser-Swinv2-Tiny, Small, and Base.}
    \label{tab:arch_swinv2}
\end{table*}

\subsubsection{Lateral Connection}
As described in the \textit{ViTEraser} section of the \textit{main paper}, lateral connections are built between the features $\{f_{i}^{enc}\}_{i=1}^{3}$ of the encoder and the features $\{f_{4-i}^{dec}\}_{i=1}^{3}$ of the decoder.
The architecture of the lateral connections is inspired by EraseNet \cite{liu2020erasenet}, which consists of non-linear, expanding, and shrinking transformations. 
For instance, if the feature $f_1 \in \mathbb{R}^{h \times w \times c}$ is laterally connected to the feature $f_2$ of the same shape, the $f_1$ sequentially goes through one $1 \times 1$ convolution with $c$ channels for non-linear transformation, two $3 \times 3$ convolutions with $2c$ channels for expanding, and one convolution with $c$ channels for shrinking. 
Finally, the resulting feature is element-wise added to the feature $f_2$.

\section{Details of Extension to TSTD}

In this section, we introduce additional details of the extension experiment to the tampered scene text detection (TSTD) task.
The ViTEraser-Swinv2-Tiny (without SegMIM) is trained using only the training set of Tampered-IC13 dataset \cite{wang2022detecting}.
The input image is resized to $1024 \times 1024$.
The network is trained for 600 epochs using an AdamW optimizer \cite{loshchilovdecoupled} with a batch size of 8.
The training is conducted using 4 NVIDIA A6000 GPUs with 48GB memory.
The learning rate is initialized as 0.0005 and linearly decayed to 0.00001 at the last epoch.
As for the text detectors including EAST \cite{zhou2017east}. PSENet \cite{wang2019shape}, and ContourNet \cite{wang2020contournet}, we use publicly released codes\footnote{https://github.com/SakuraRiven/EAST.}$^,$\footnote{https://github.com/whai362/PSENet.}$^,$\footnote{https://github.com/wangyuxin87/ContourNet.} to train them using only the training set of Tampered-IC13.
During inference, the input image is resized to $748 \times 748$ for EAST. 
For PSENet, the short side of the image is resized to 736 pixels.
For ContourNet, the short size of the image is resized to 1200 pixels while keeping the long side shorter than 2000 pixels.

\begin{figure}[t]
\centering
\subfigtopskip=0pt 
\subfigbottomskip=2pt 
\subfigcapskip=1pt 
\renewcommand{\subcapsize}{\scriptsize}
\subfigure{
    \begin{minipage}[c]{0.25\columnwidth}
        \centering
        \includegraphics[width=0.95\columnwidth]{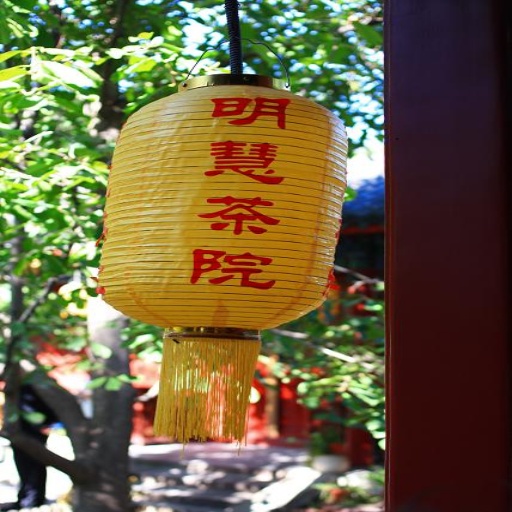}
    \end{minipage}%
}%
\subfigure{
    \begin{minipage}[c]{0.25\columnwidth}
        \centering
        \includegraphics[width=0.95\columnwidth]{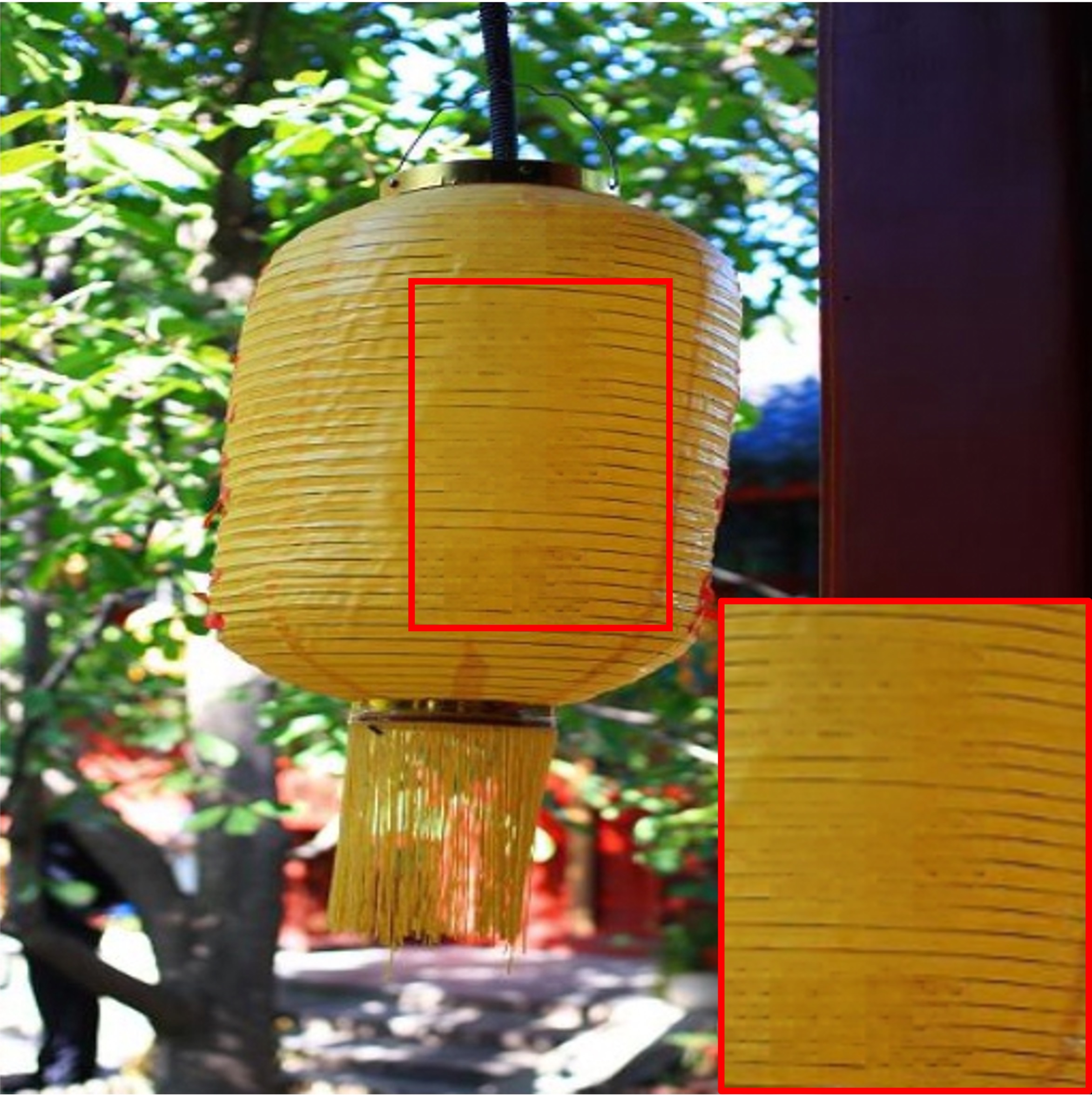}
    \end{minipage}%
}%
\subfigure{
    \begin{minipage}[c]{0.25\columnwidth}
        \centering
        \includegraphics[width=0.95\columnwidth]{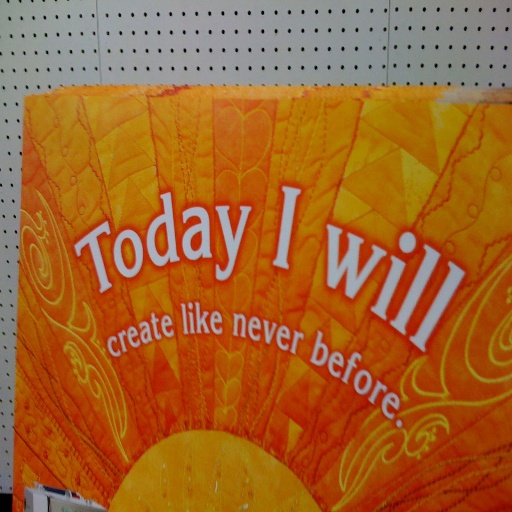}
    \end{minipage}%
}%
\subfigure{
    \begin{minipage}[c]{0.25\columnwidth}
        \centering
        \includegraphics[width=0.95\columnwidth]{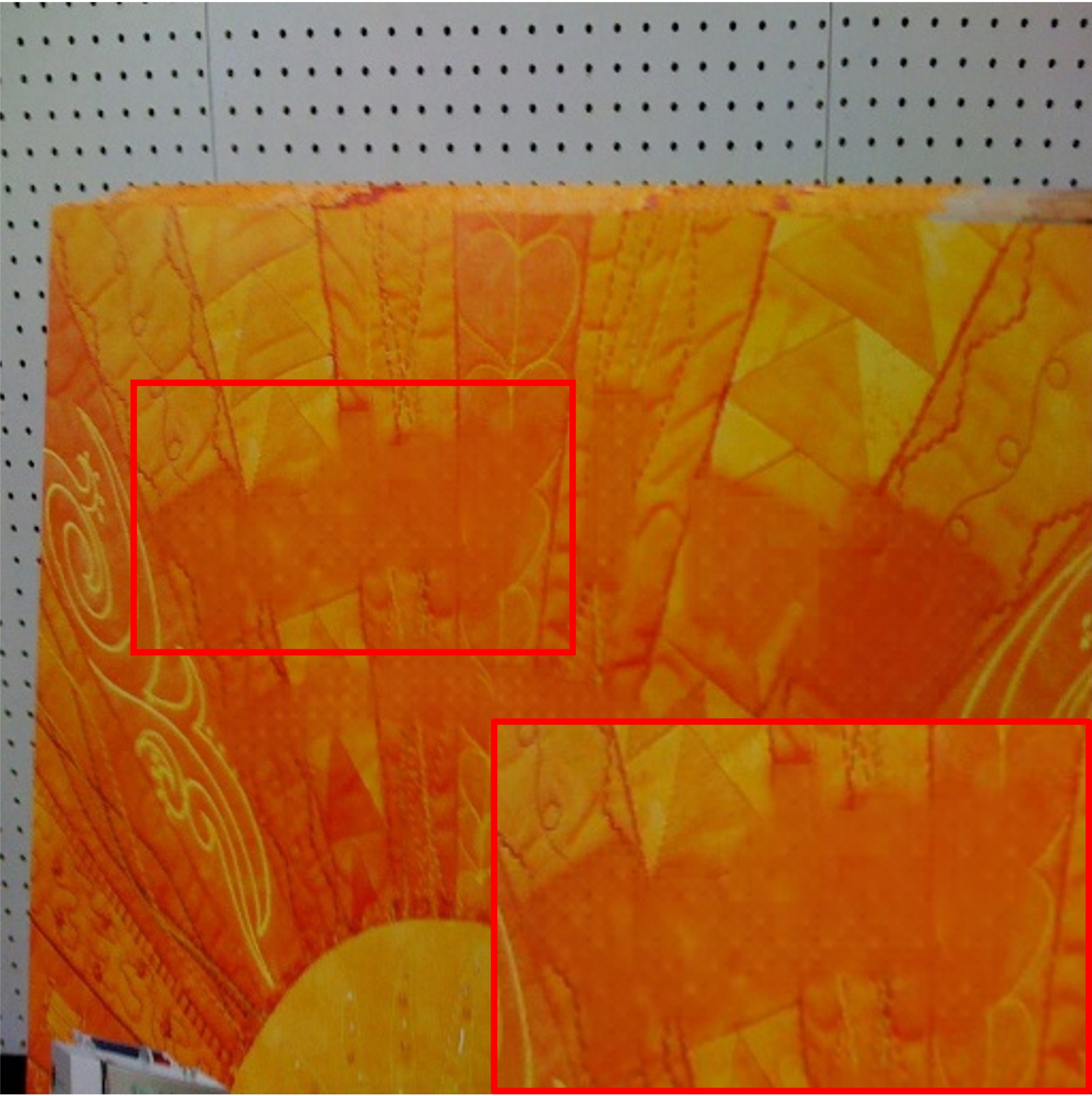}
    \end{minipage}%
}
\setcounter{subfigure}{0}
\subfigure[Input Image]{
    \begin{minipage}[c]{0.25\columnwidth}
        \centering
        \includegraphics[width=0.95\columnwidth]{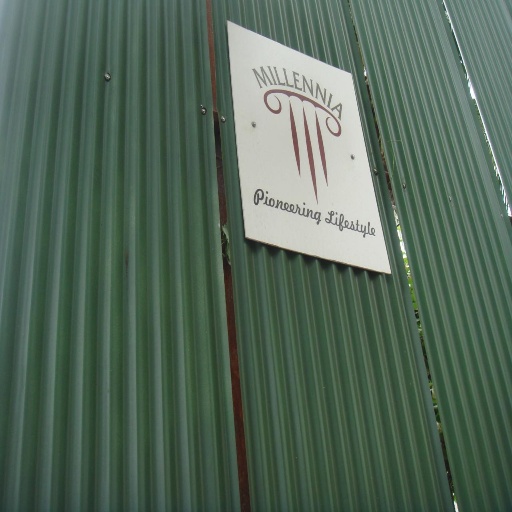}
    \end{minipage}%
}%
\subfigure[ViTEraser]{
    \begin{minipage}[c]{0.25\columnwidth}
        \centering
        \includegraphics[width=0.95\columnwidth]{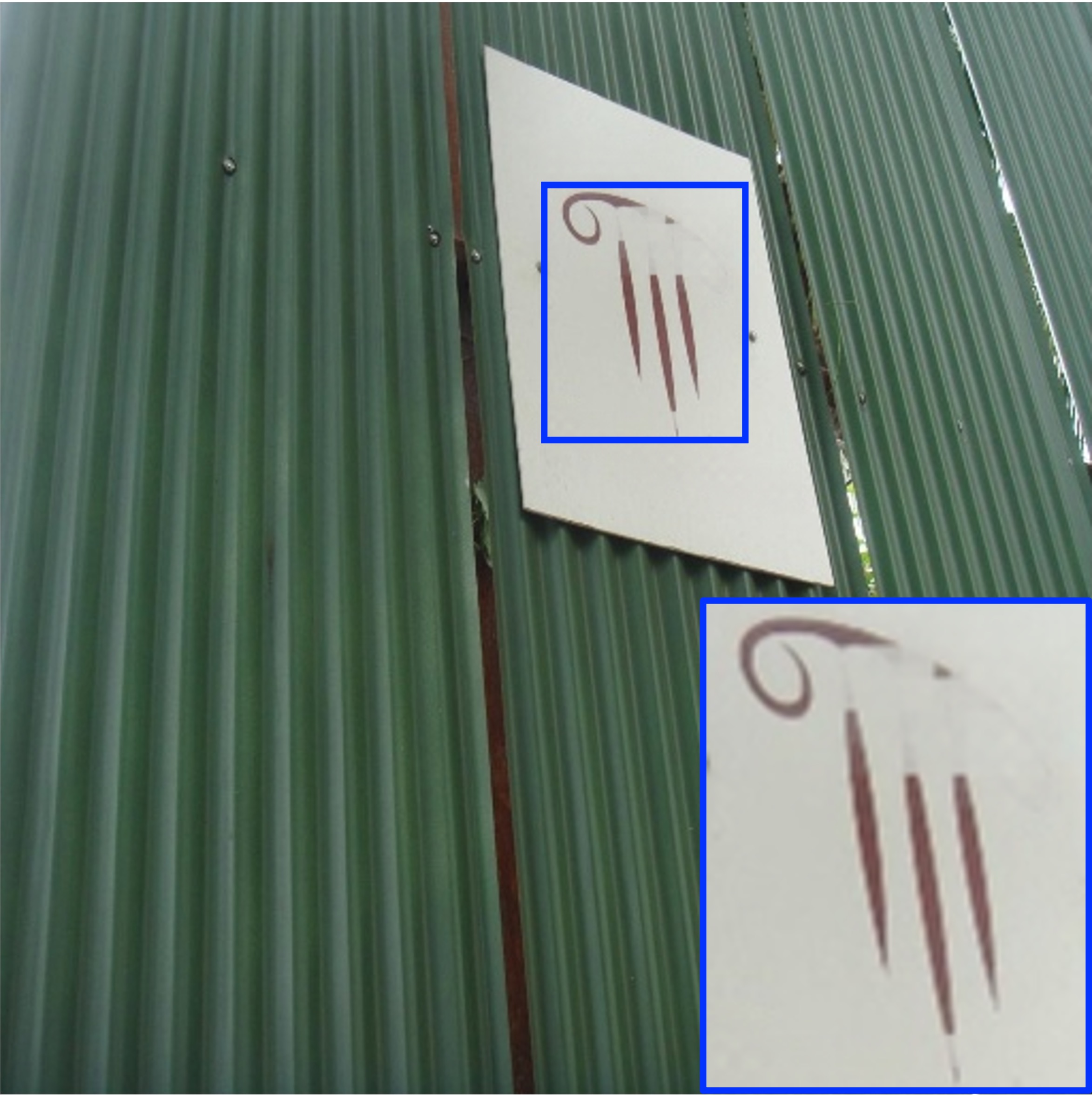}
    \end{minipage}%
}%
\subfigure[Input Image]{
    \begin{minipage}[c]{0.25\columnwidth}
        \centering
        \includegraphics[width=0.95\columnwidth]{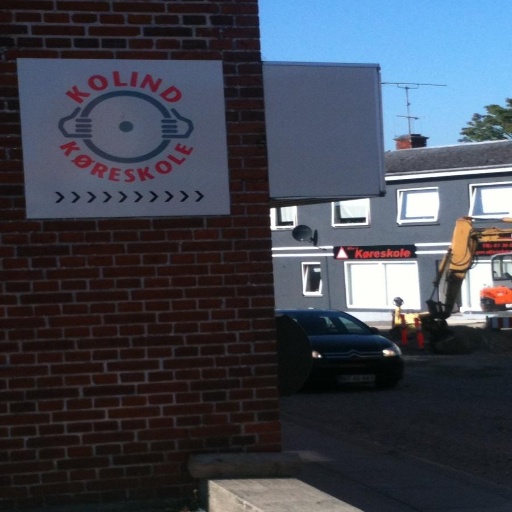}
    \end{minipage}%
}%
\subfigure[ViTEraser]{
    \begin{minipage}[c]{0.25\columnwidth}
        \centering
        \includegraphics[width=0.95\columnwidth]{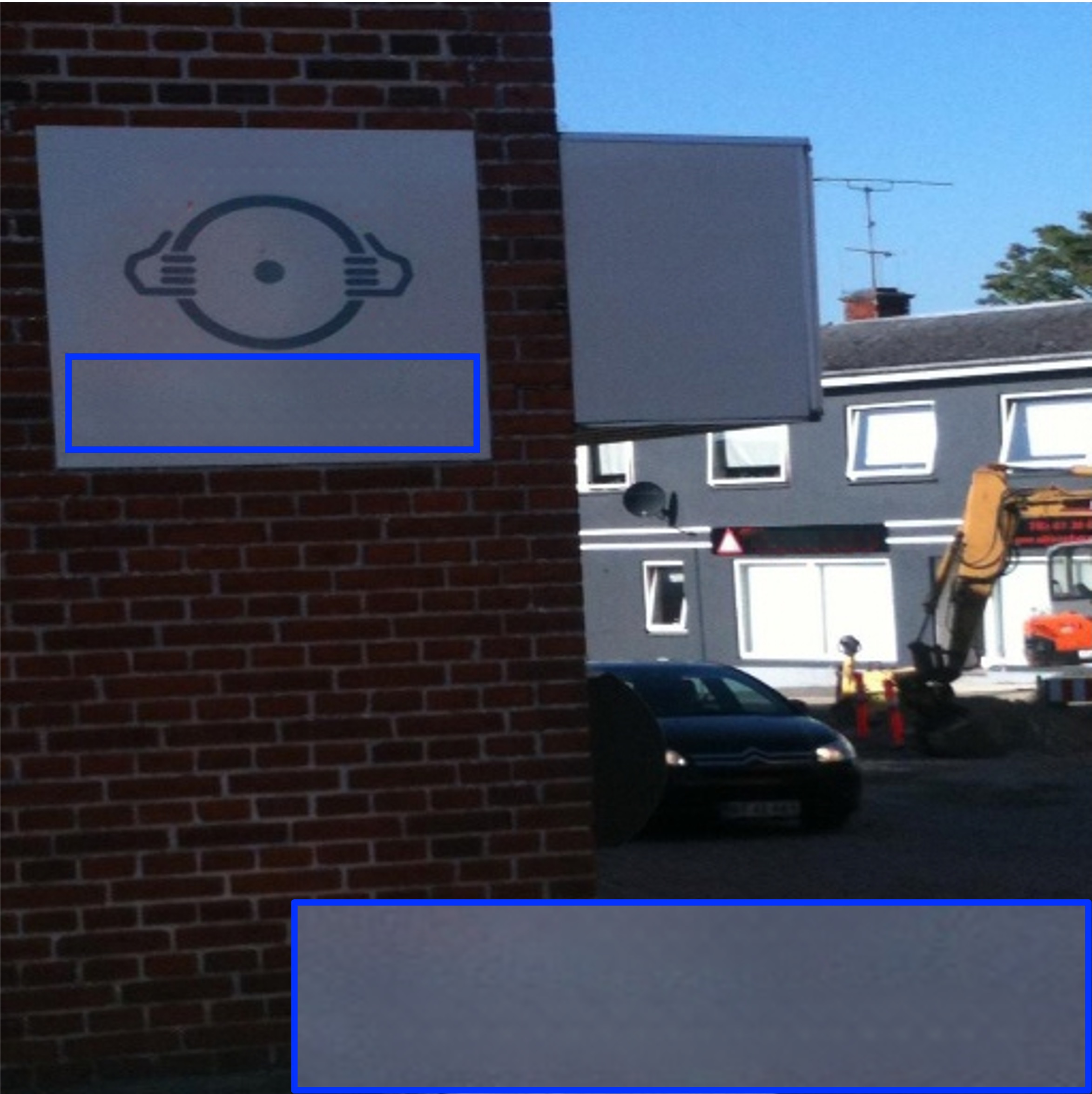}
    \end{minipage}%
}
\caption{\small Failure cases of ViTEraser.}
\label{fig:failure_cases}
\end{figure}

\section{Failure Case}
Although the proposed ViTEraser achieves state-of-the-art STR performance in both quantitative metrics and qualitative visualizations, there are still several situations that it cannot handle well.
For instance, ViTEraser struggles with reconstructing complicated background textures (red boxes) and has a chance to excessively erase text-like patterns (blue boxes), as shown in Fig.~\ref{fig:failure_cases}.

\section{Limitation}

Despite the impressive performance achieved by the ViTEraser, 
its limitations lie in the large model size and relatively slow inference speed. 
As shown in \textit{Tab.}~\ref{tab:exp_scut_enstext} of the \textit{main paper}, the ViTEraser-Swinv2-Tiny, Small, and Base contain 65M, 108M, and 192M parameters, as well as reach inference speeds of 24fps, 17fps, and 15fps, respectively.
In comparison, most previous STR models, except for Pix2Pix \cite{isola2017image} and CTRNet \cite{liu2022don}, keep their parameter sizes below 35M.
Additionally, while the inference speed of ViTEraser may be adequate in real-world applications and is faster than several approaches including SSTE \cite{tang2021stroke} and CTRNet \cite{liu2022don}, it is slower than many other methods.
To address these limitations, future research aims to incorporate existing lightweight and efficient Transformer architectures \cite{tay2022efficient} to overcome the challenges associated with the large model size and slow inference speed of ViTEraser.

\section{More Visualizations}

\subsubsection{SCUT-EnsText}

\begin{figure*}[t]
    \centering
    \subfigtopskip=0pt 
    \subfigbottomskip=2pt 
    \subfigcapskip=1pt 
    \renewcommand{\subcapsize}{\scriptsize}
    \subfigure{
        \begin{minipage}[c]{0.105\linewidth}
            \centering
            \includegraphics[width=\linewidth]{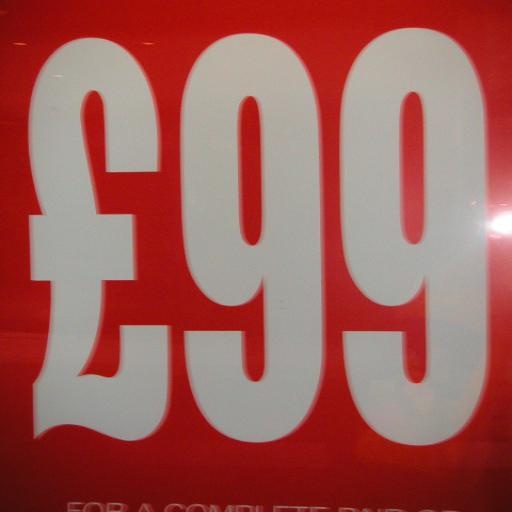}
        \end{minipage}%
    }%
    \subfigure{
        \begin{minipage}[c]{0.105\linewidth}
            \centering
            \includegraphics[width=\linewidth]{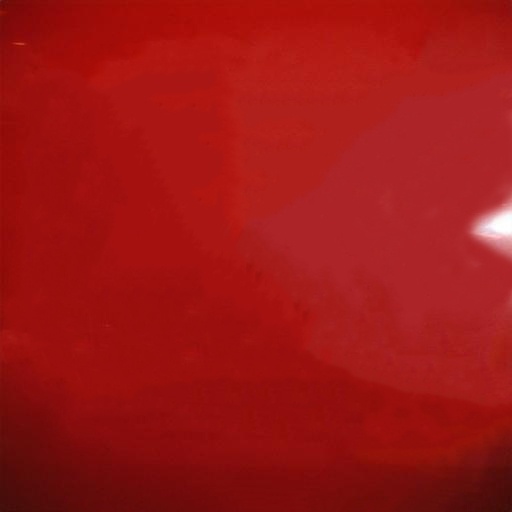}
        \end{minipage}%
    }%
    \subfigure{
        \begin{minipage}[c]{0.105\linewidth}
            \centering
            \includegraphics[width=\linewidth]{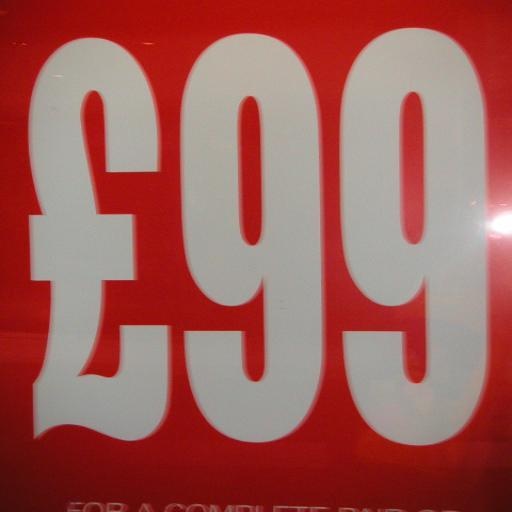}
        \end{minipage}%
    }%
    \subfigure{
        \begin{minipage}[c]{0.105\linewidth}
            \centering
            \includegraphics[width=\linewidth]{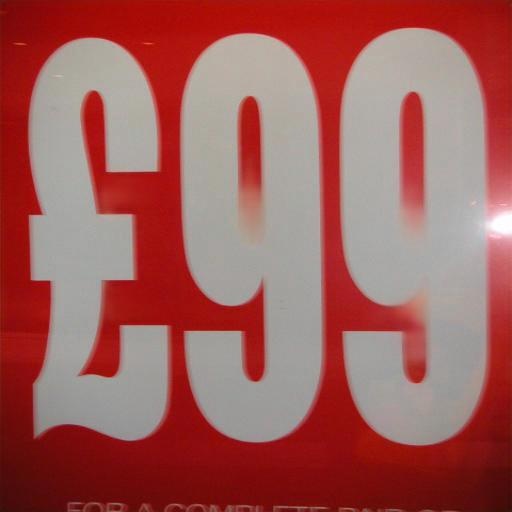}
        \end{minipage}%
    }%
    \subfigure{
        \begin{minipage}[c]{0.105\linewidth}
            \centering
            \includegraphics[width=\linewidth]{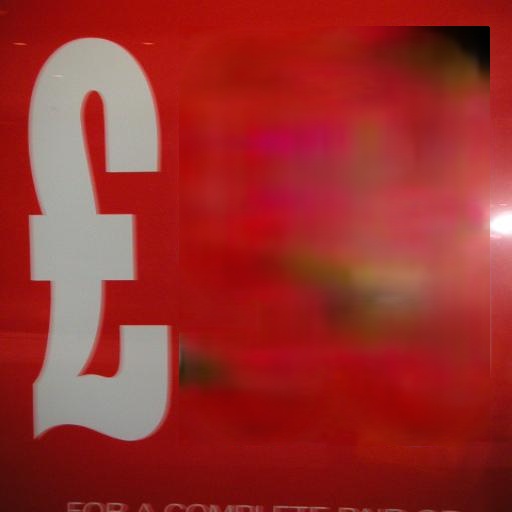}
        \end{minipage}%
    }%
    \subfigure{
        \begin{minipage}[c]{0.105\linewidth}
            \centering
            \includegraphics[width=\linewidth]{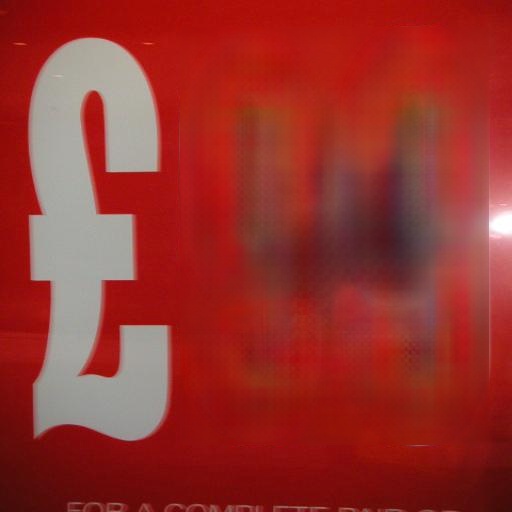}
        \end{minipage}%
    }%
    \subfigure{
        \begin{minipage}[c]{0.105\linewidth}
            \centering
            \includegraphics[width=\linewidth]{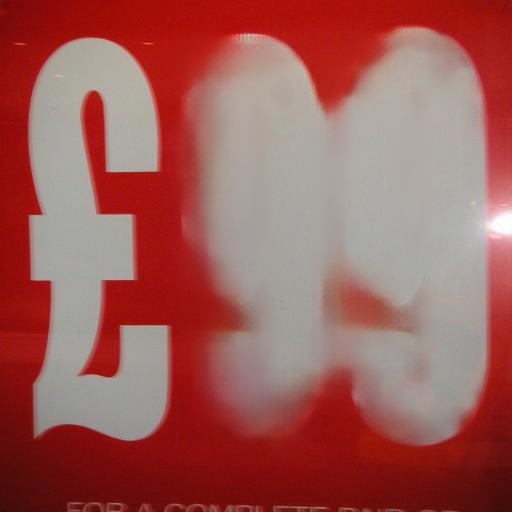}
        \end{minipage}%
    }%
    \subfigure{
        \begin{minipage}[c]{0.105\linewidth}
            \centering
            \includegraphics[width=\linewidth]{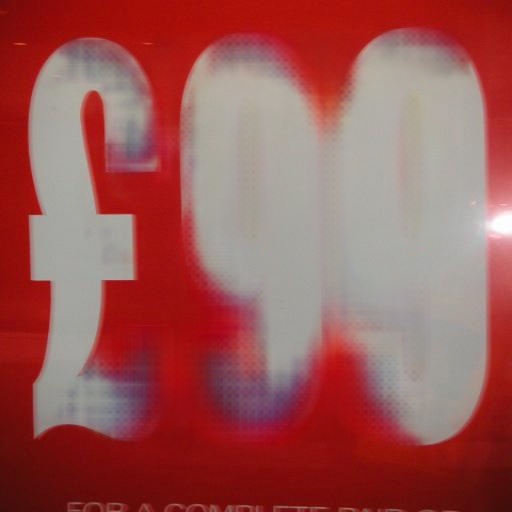}
        \end{minipage}%
    }%
    \subfigure{
        \begin{minipage}[c]{0.105\linewidth}
            \centering
            \includegraphics[width=\linewidth]{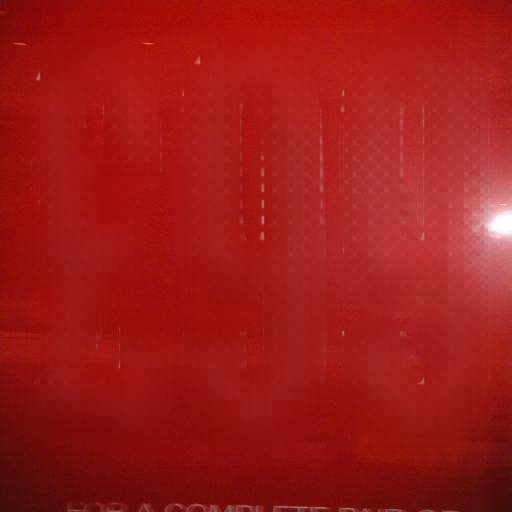}
        \end{minipage}%
    }
    \subfigure{
        \begin{minipage}[c]{0.105\linewidth}
            \centering
            \includegraphics[width=\linewidth]{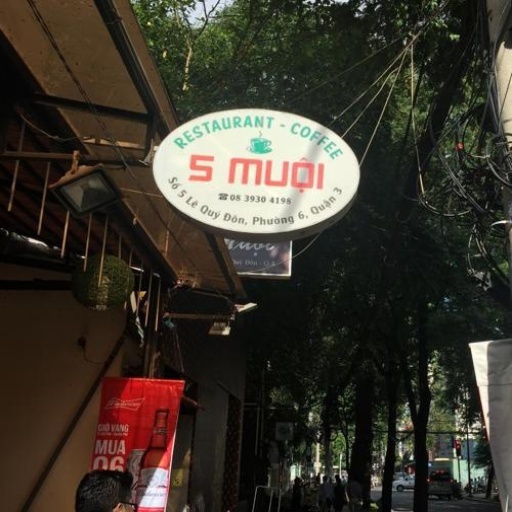}
        \end{minipage}%
    }%
    \subfigure{
        \begin{minipage}[c]{0.105\linewidth}
            \centering
            \includegraphics[width=\linewidth]{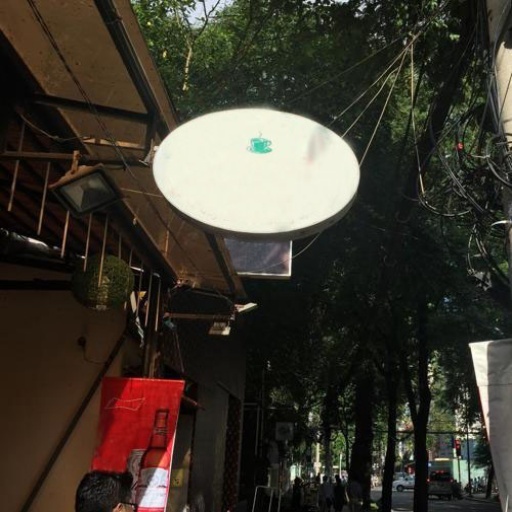}
        \end{minipage}%
    }%
    \subfigure{
        \begin{minipage}[c]{0.105\linewidth}
            \centering
            \includegraphics[width=\linewidth]{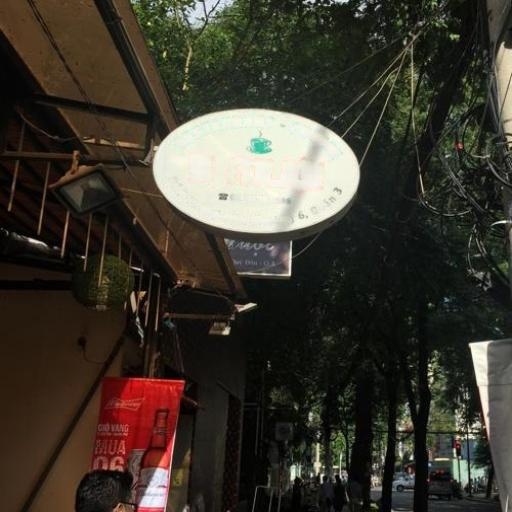}
        \end{minipage}%
    }%
    \subfigure{
        \begin{minipage}[c]{0.105\linewidth}
            \centering
            \includegraphics[width=\linewidth]{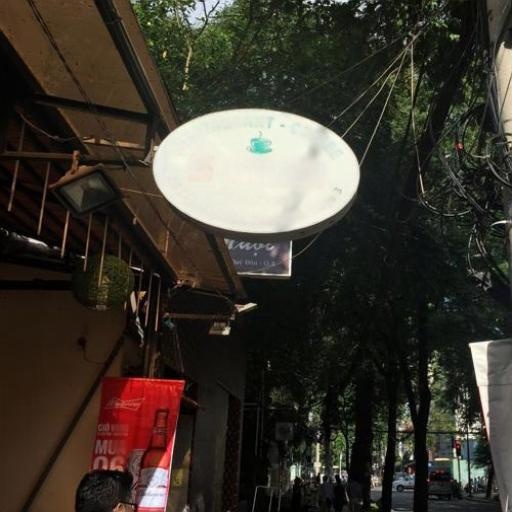}
        \end{minipage}%
    }%
    \subfigure{
        \begin{minipage}[c]{0.105\linewidth}
            \centering
            \includegraphics[width=\linewidth]{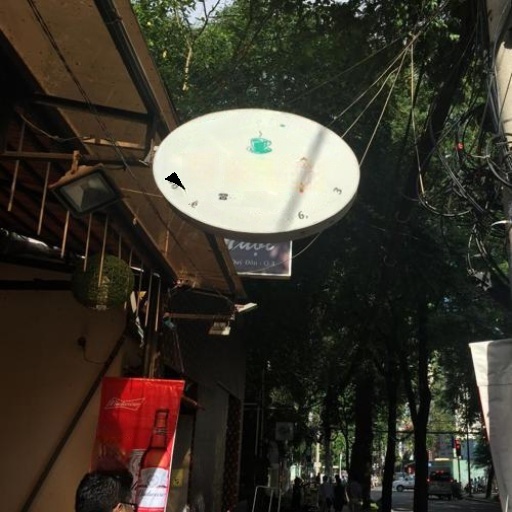}
        \end{minipage}%
    }%
    \subfigure{
        \begin{minipage}[c]{0.105\linewidth}
            \centering
            \includegraphics[width=\linewidth]{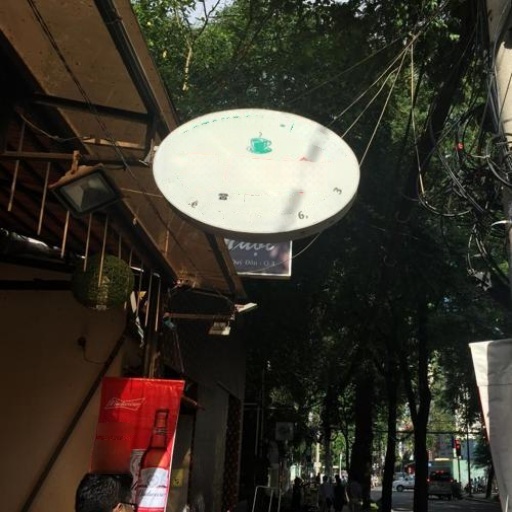}
        \end{minipage}%
    }%
    \subfigure{
        \begin{minipage}[c]{0.105\linewidth}
            \centering
            \includegraphics[width=\linewidth]{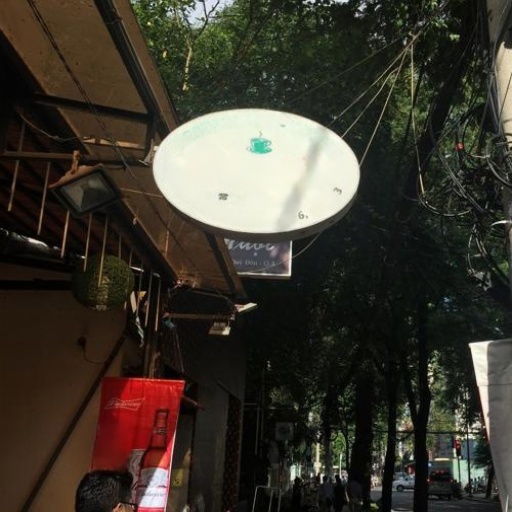}
        \end{minipage}%
    }%
    \subfigure{
        \begin{minipage}[c]{0.105\linewidth}
            \centering
            \includegraphics[width=\linewidth]{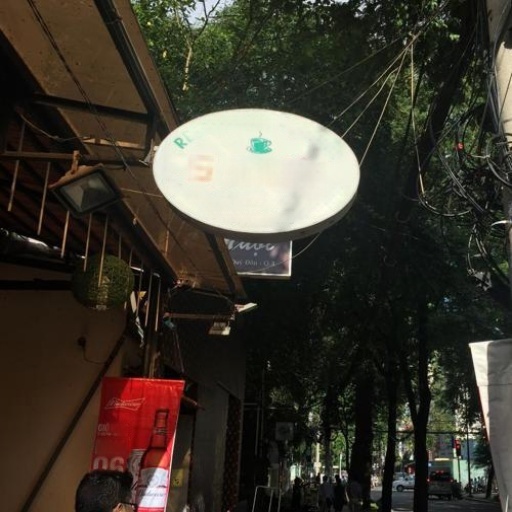}
        \end{minipage}%
    }%
    \subfigure{
        \begin{minipage}[c]{0.105\linewidth}
            \centering
            \includegraphics[width=\linewidth]{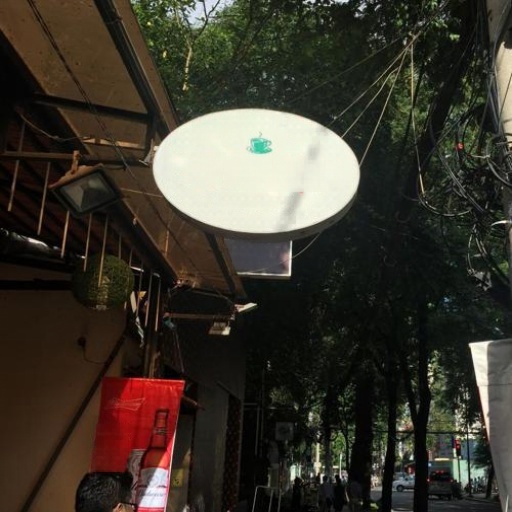}
        \end{minipage}%
    }
    \subfigure{
        \begin{minipage}[c]{0.105\linewidth}
            \centering
            \includegraphics[width=\linewidth]{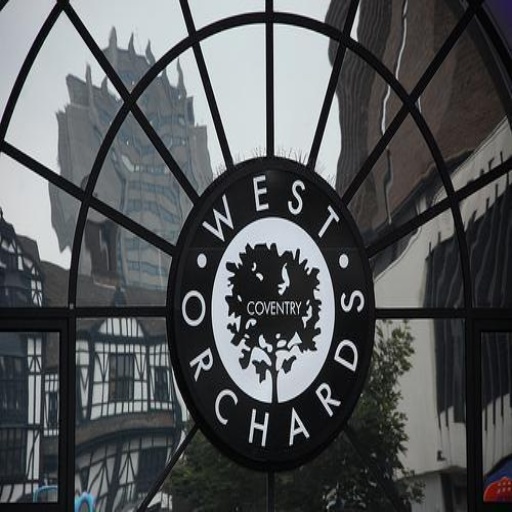}
        \end{minipage}%
    }%
    \subfigure{
        \begin{minipage}[c]{0.105\linewidth}
            \centering
            \includegraphics[width=\linewidth]{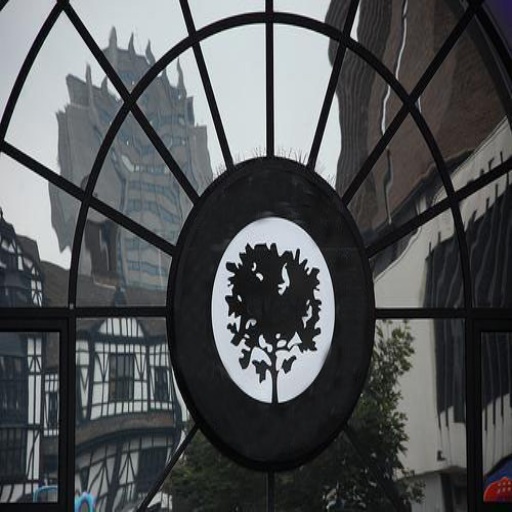}
        \end{minipage}%
    }%
    \subfigure{
        \begin{minipage}[c]{0.105\linewidth}
            \centering
            \includegraphics[width=\linewidth]{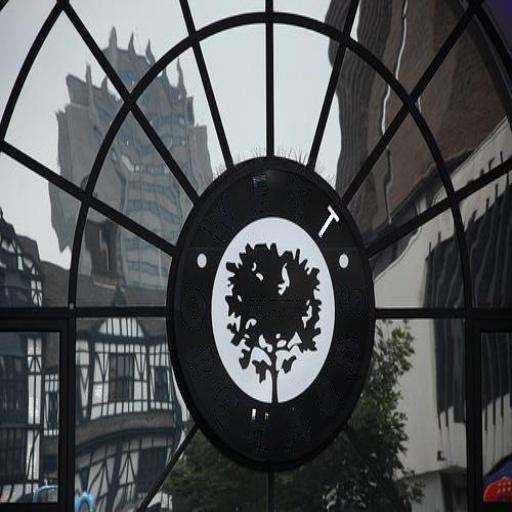}
        \end{minipage}%
    }%
    \subfigure{
        \begin{minipage}[c]{0.105\linewidth}
            \centering
            \includegraphics[width=\linewidth]{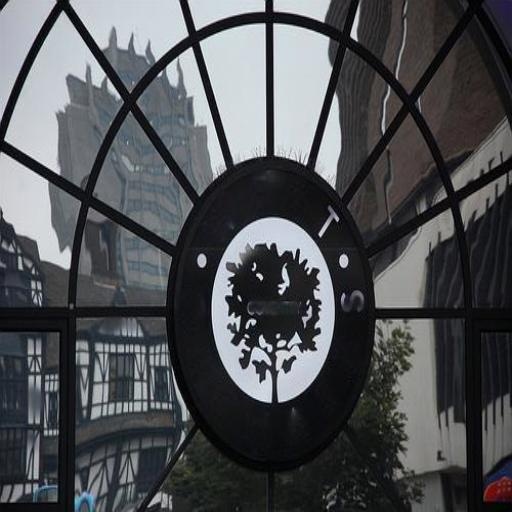}
        \end{minipage}%
    }%
    \subfigure{
        \begin{minipage}[c]{0.105\linewidth}
            \centering
            \includegraphics[width=\linewidth]{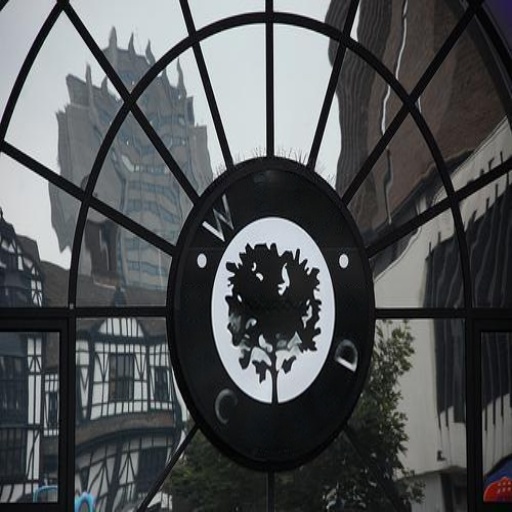}
        \end{minipage}%
    }%
    \subfigure{
        \begin{minipage}[c]{0.105\linewidth}
            \centering
            \includegraphics[width=\linewidth]{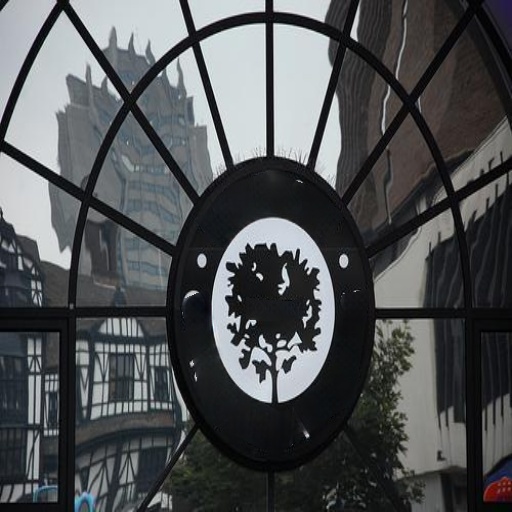}
        \end{minipage}%
    }%
    \subfigure{
        \begin{minipage}[c]{0.105\linewidth}
            \centering
            \includegraphics[width=\linewidth]{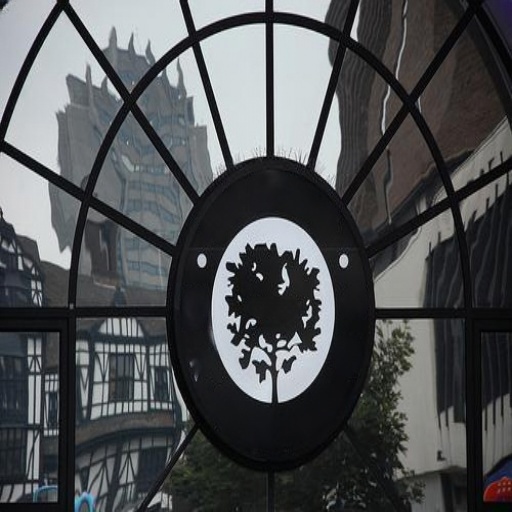}
        \end{minipage}%
    }%
    \subfigure{
        \begin{minipage}[c]{0.105\linewidth}
            \centering
            \includegraphics[width=\linewidth]{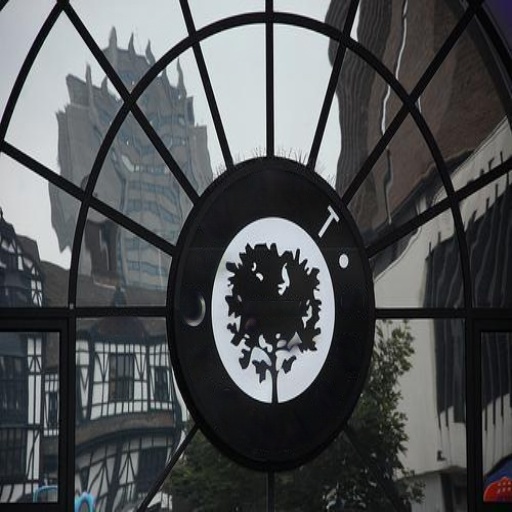}
        \end{minipage}%
    }%
    \subfigure{
        \begin{minipage}[c]{0.105\linewidth}
            \centering
            \includegraphics[width=\linewidth]{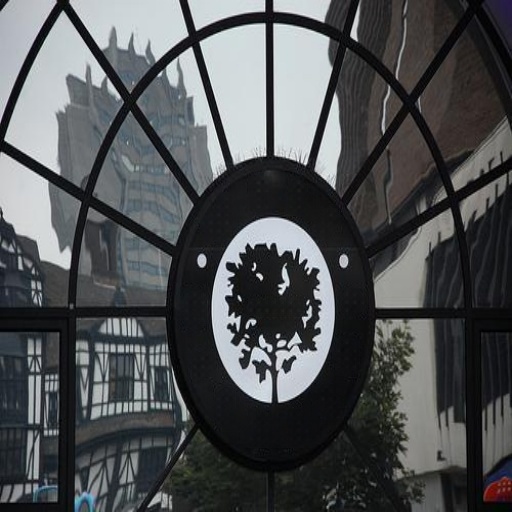}
        \end{minipage}%
    }
    \subfigure{
        \begin{minipage}[c]{0.105\linewidth}
            \centering
            \includegraphics[width=\linewidth]{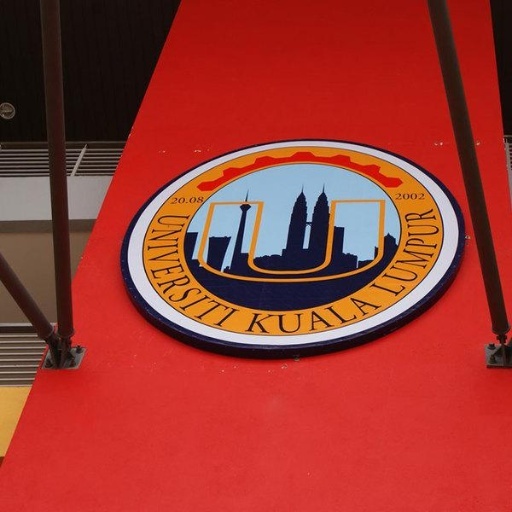}
        \end{minipage}%
    }%
    \subfigure{
        \begin{minipage}[c]{0.105\linewidth}
            \centering
            \includegraphics[width=\linewidth]{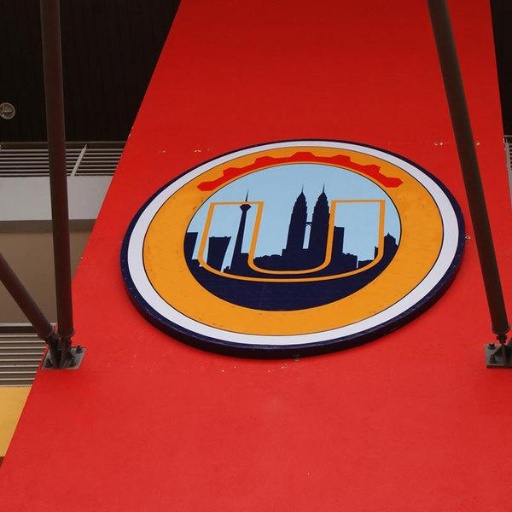}
        \end{minipage}%
    }%
    \subfigure{
        \begin{minipage}[c]{0.105\linewidth}
            \centering
            \includegraphics[width=\linewidth]{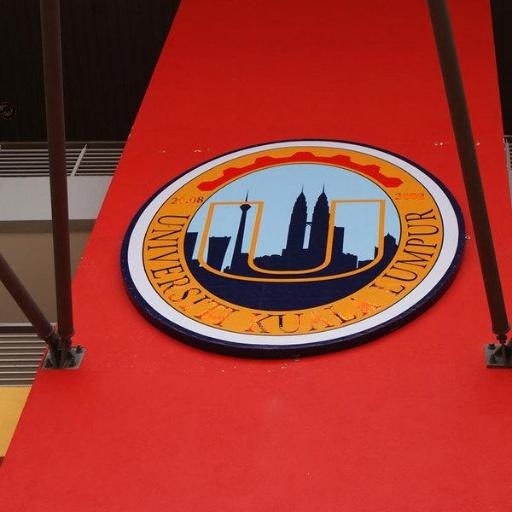}
        \end{minipage}%
    }%
    \subfigure{
        \begin{minipage}[c]{0.105\linewidth}
            \centering
            \includegraphics[width=\linewidth]{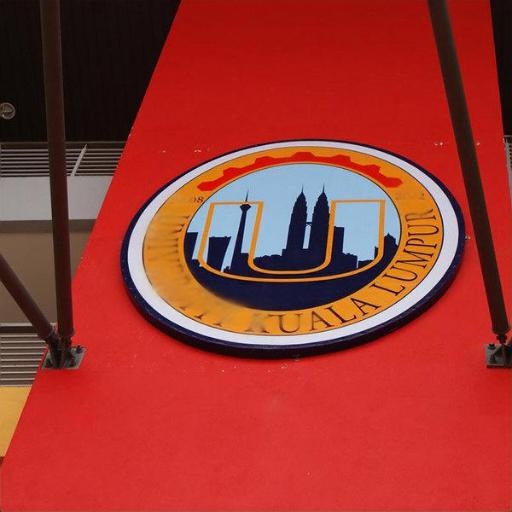}
        \end{minipage}%
    }%
    \subfigure{
        \begin{minipage}[c]{0.105\linewidth}
            \centering
            \includegraphics[width=\linewidth]{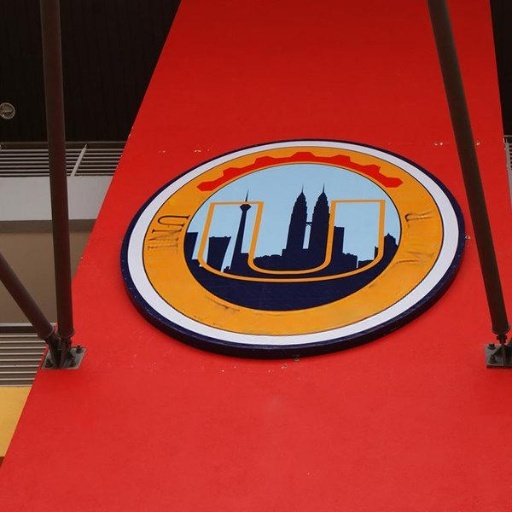}
        \end{minipage}%
    }%
    \subfigure{
        \begin{minipage}[c]{0.105\linewidth}
            \centering
            \includegraphics[width=\linewidth]{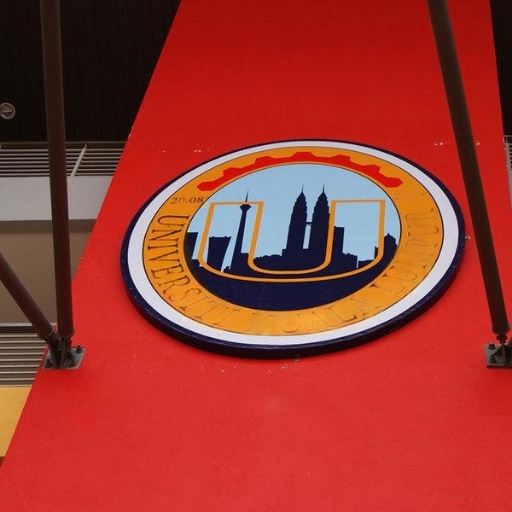}
        \end{minipage}%
    }%
    \subfigure{
        \begin{minipage}[c]{0.105\linewidth}
            \centering
            \includegraphics[width=\linewidth]{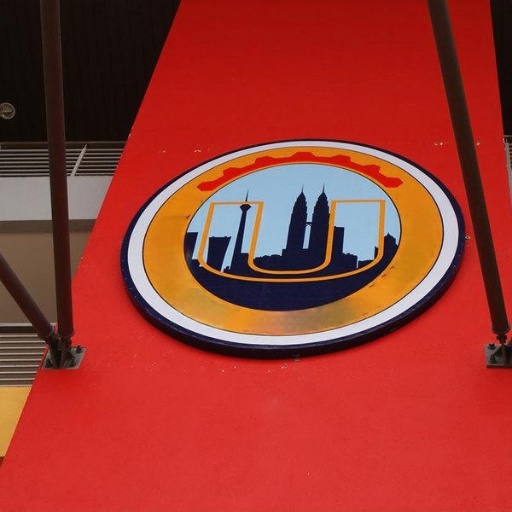}
        \end{minipage}%
    }%
    \subfigure{
        \begin{minipage}[c]{0.105\linewidth}
            \centering
            \includegraphics[width=\linewidth]{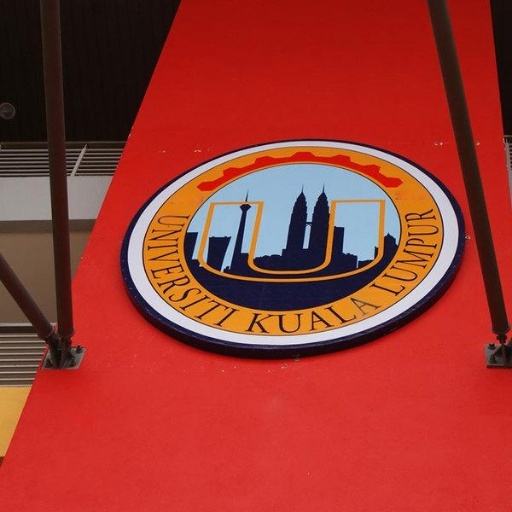}
        \end{minipage}%
    }%
    \subfigure{
        \begin{minipage}[c]{0.105\linewidth}
            \centering
            \includegraphics[width=\linewidth]{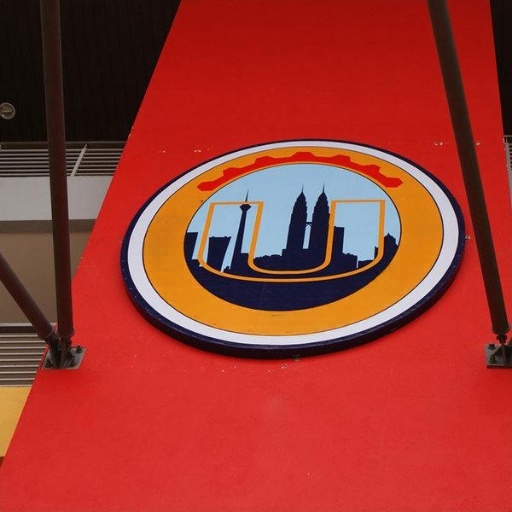}
        \end{minipage}%
    }
    \subfigure{
        \begin{minipage}[c]{0.105\linewidth}
            \centering
            \includegraphics[width=\linewidth]{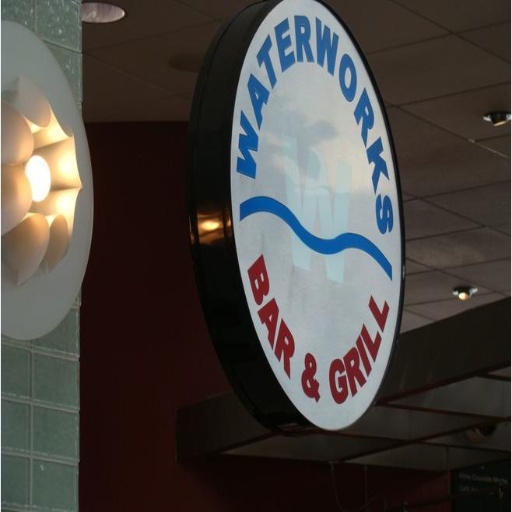}
        \end{minipage}%
    }%
    \subfigure{
        \begin{minipage}[c]{0.105\linewidth}
            \centering
            \includegraphics[width=\linewidth]{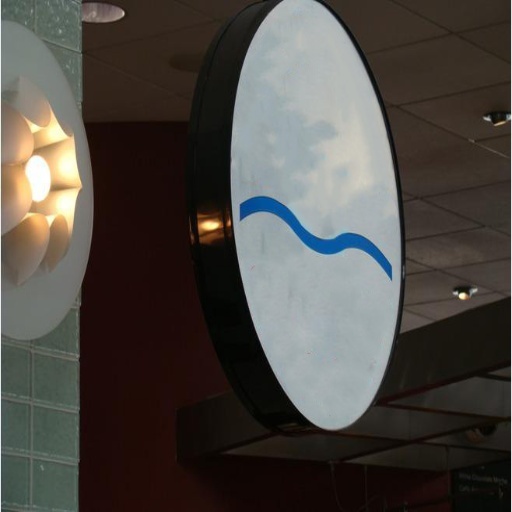}
        \end{minipage}%
    }%
    \subfigure{
        \begin{minipage}[c]{0.105\linewidth}
            \centering
            \includegraphics[width=\linewidth]{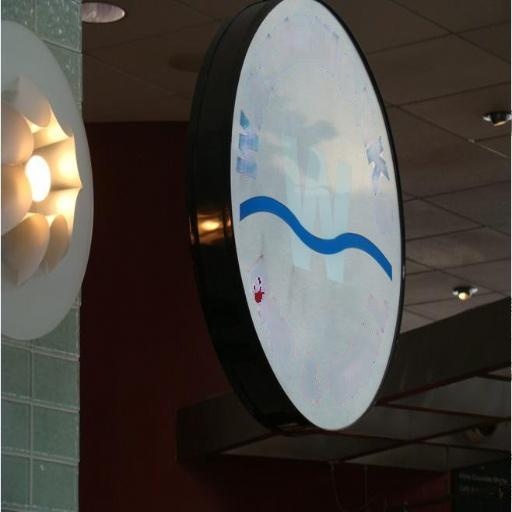}
        \end{minipage}%
    }%
    \subfigure{
        \begin{minipage}[c]{0.105\linewidth}
            \centering
            \includegraphics[width=\linewidth]{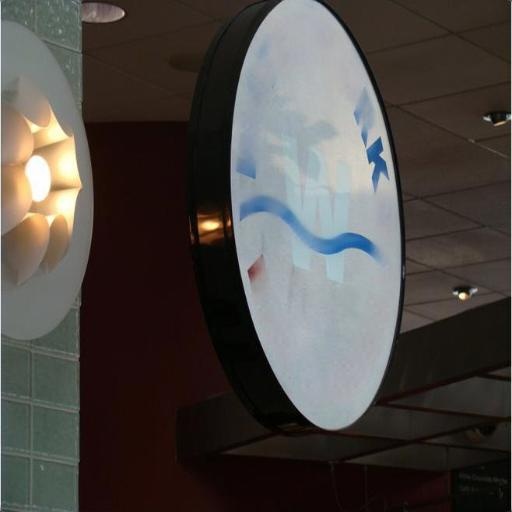}
        \end{minipage}%
    }%
    \subfigure{
        \begin{minipage}[c]{0.105\linewidth}
            \centering
            \includegraphics[width=\linewidth]{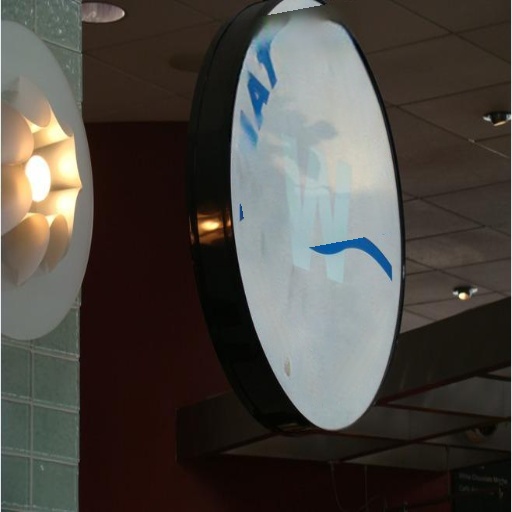}
        \end{minipage}%
    }%
    \subfigure{
        \begin{minipage}[c]{0.105\linewidth}
            \centering
            \includegraphics[width=\linewidth]{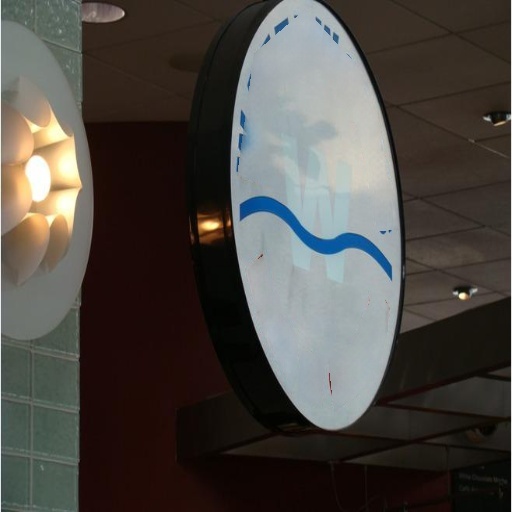}
        \end{minipage}%
    }%
    \subfigure{
        \begin{minipage}[c]{0.105\linewidth}
            \centering
            \includegraphics[width=\linewidth]{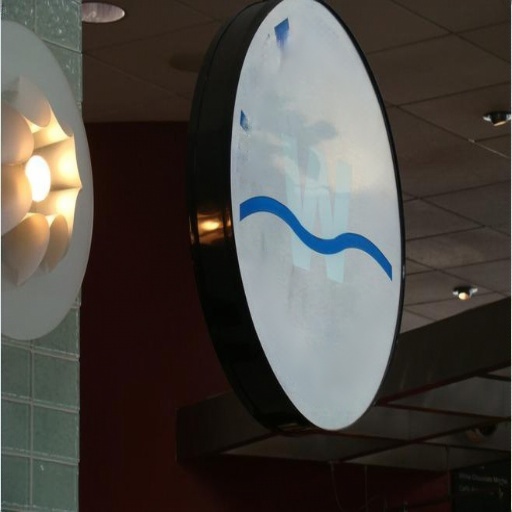}
        \end{minipage}%
    }%
    \subfigure{
        \begin{minipage}[c]{0.105\linewidth}
            \centering
            \includegraphics[width=\linewidth]{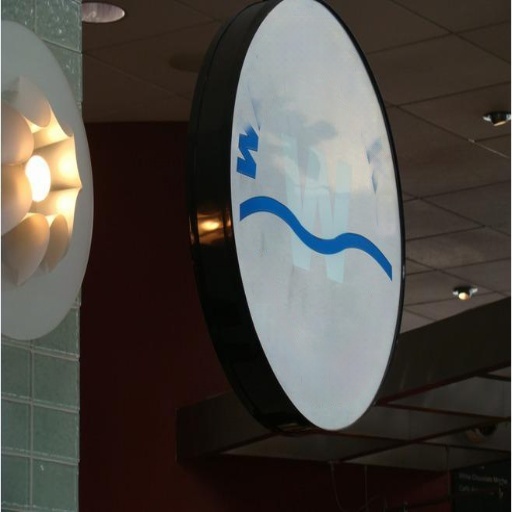}
        \end{minipage}%
    }%
    \subfigure{
        \begin{minipage}[c]{0.105\linewidth}
            \centering
            \includegraphics[width=\linewidth]{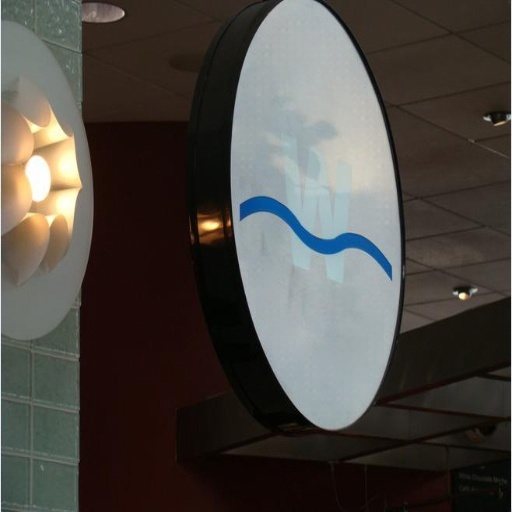}
        \end{minipage}%
    }
    \subfigure{
        \begin{minipage}[c]{0.105\linewidth}
            \centering
            \includegraphics[width=\linewidth]{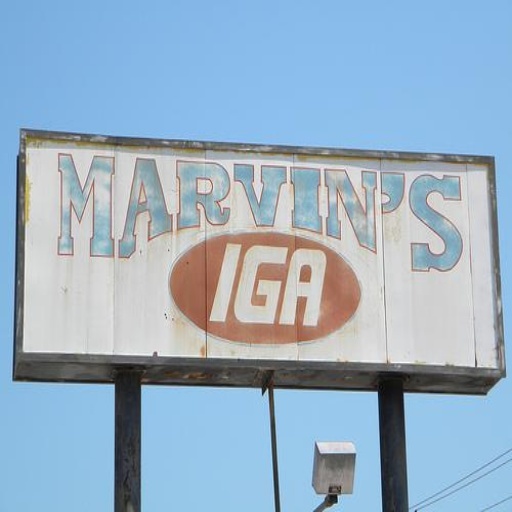}
        \end{minipage}%
    }%
    \subfigure{
        \begin{minipage}[c]{0.105\linewidth}
            \centering
            \includegraphics[width=\linewidth]{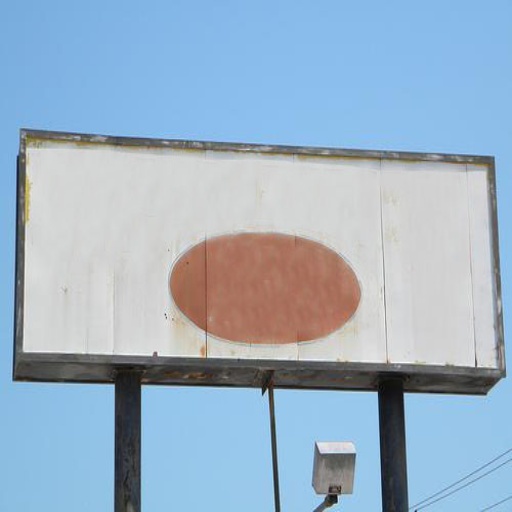}
        \end{minipage}%
    }%
    \subfigure{
        \begin{minipage}[c]{0.105\linewidth}
            \centering
            \includegraphics[width=\linewidth]{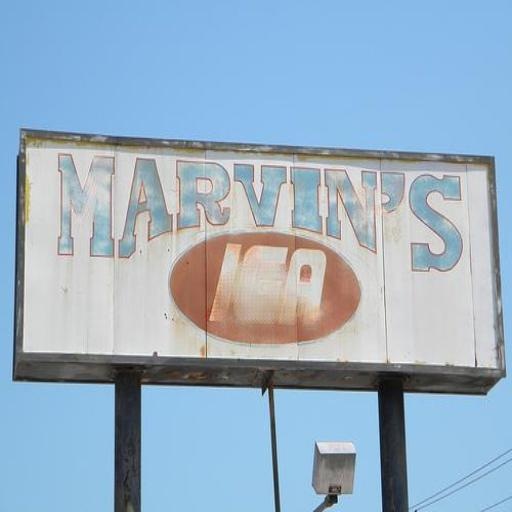}
        \end{minipage}%
    }%
    \subfigure{
        \begin{minipage}[c]{0.105\linewidth}
            \centering
            \includegraphics[width=\linewidth]{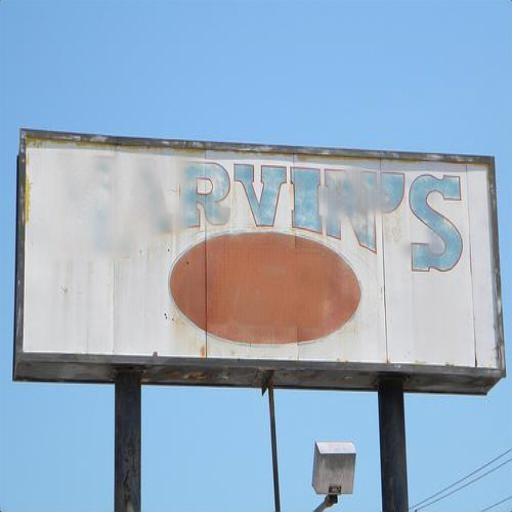}
        \end{minipage}%
    }%
    \subfigure{
        \begin{minipage}[c]{0.105\linewidth}
            \centering
            \includegraphics[width=\linewidth]{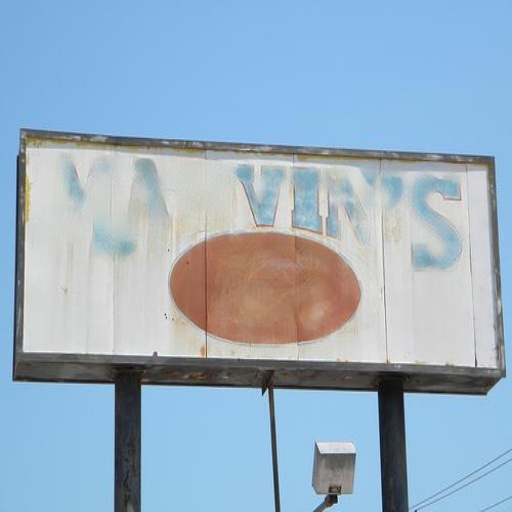}
        \end{minipage}%
    }%
    \subfigure{
        \begin{minipage}[c]{0.105\linewidth}
            \centering
            \includegraphics[width=\linewidth]{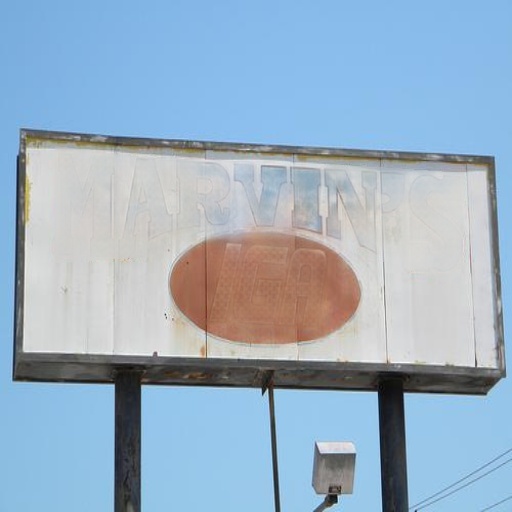}
        \end{minipage}%
    }%
    \subfigure{
        \begin{minipage}[c]{0.105\linewidth}
            \centering
            \includegraphics[width=\linewidth]{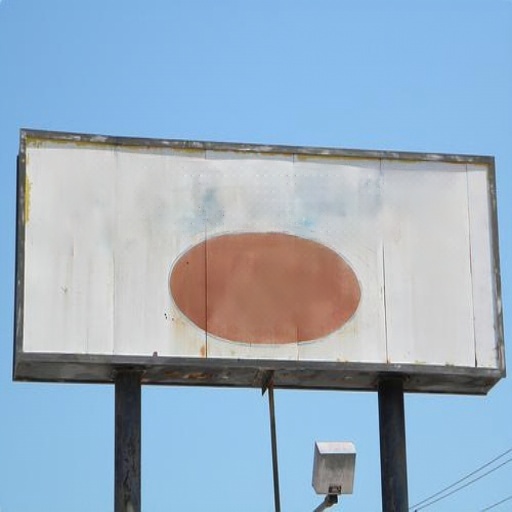}
        \end{minipage}%
    }%
    \subfigure{
        \begin{minipage}[c]{0.105\linewidth}
            \centering
            \includegraphics[width=\linewidth]{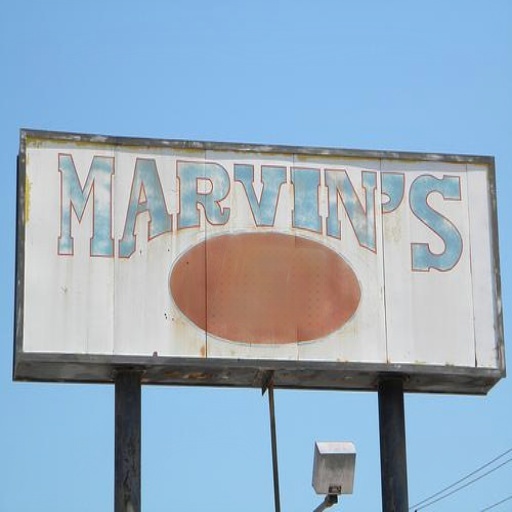}
        \end{minipage}%
    }%
    \subfigure{
        \begin{minipage}[c]{0.105\linewidth}
            \centering
            \includegraphics[width=\linewidth]{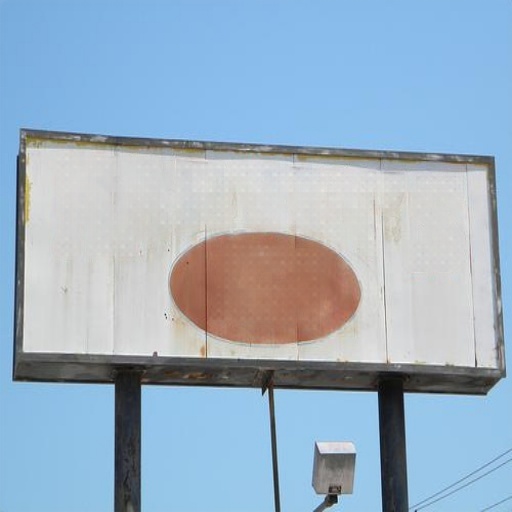}
        \end{minipage}%
    }
    \subfigure{
        \begin{minipage}[c]{0.105\linewidth}
            \centering
            \includegraphics[width=\linewidth]{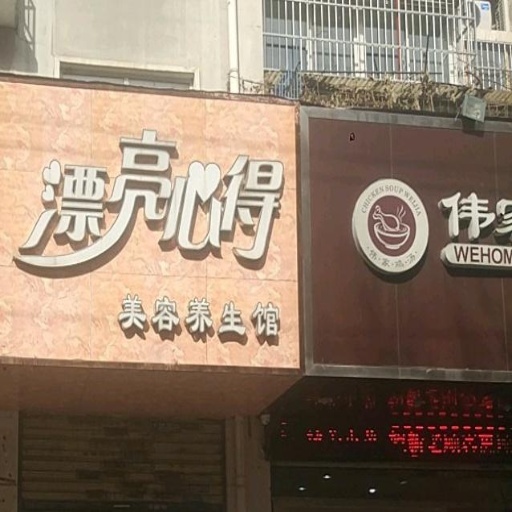}
        \end{minipage}%
    }%
    \subfigure{
        \begin{minipage}[c]{0.105\linewidth}
            \centering
            \includegraphics[width=\linewidth]{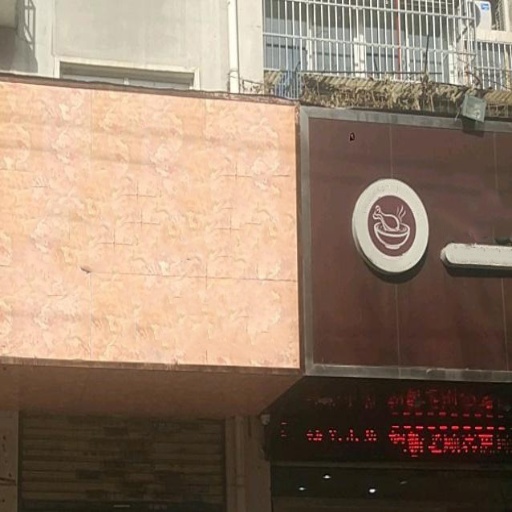}
        \end{minipage}%
    }%
    \subfigure{
        \begin{minipage}[c]{0.105\linewidth}
            \centering
            \includegraphics[width=\linewidth]{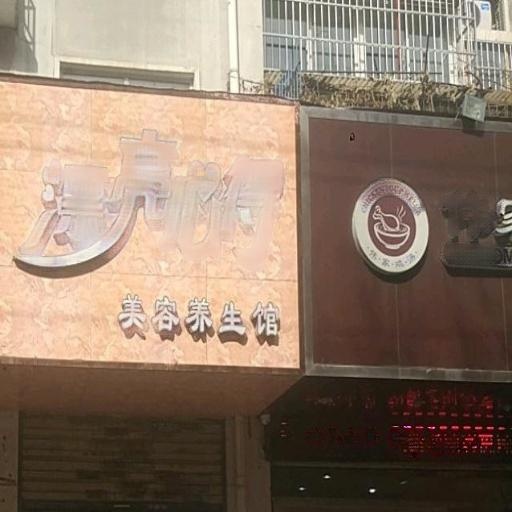}
        \end{minipage}%
    }%
    \subfigure{
        \begin{minipage}[c]{0.105\linewidth}
            \centering
            \includegraphics[width=\linewidth]{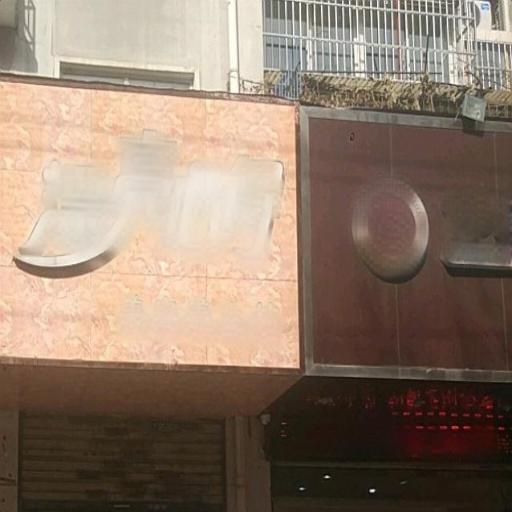}
        \end{minipage}%
    }%
    \subfigure{
        \begin{minipage}[c]{0.105\linewidth}
            \centering
            \includegraphics[width=\linewidth]{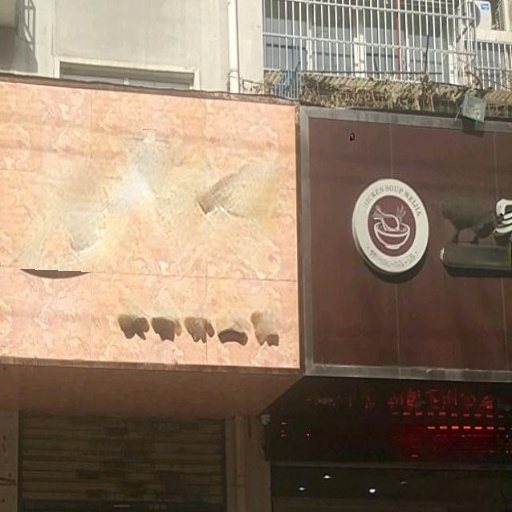}
        \end{minipage}%
    }%
    \subfigure{
        \begin{minipage}[c]{0.105\linewidth}
            \centering
            \includegraphics[width=\linewidth]{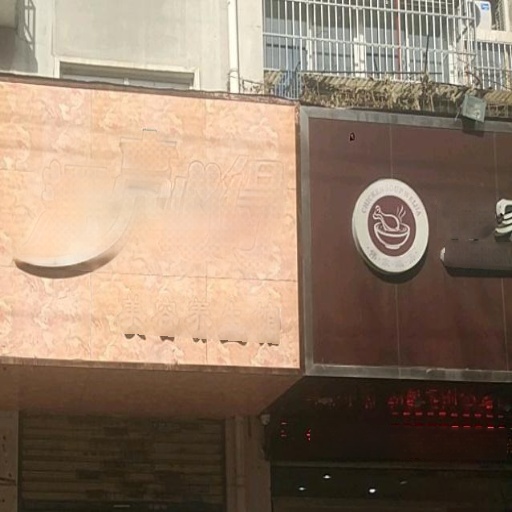}
        \end{minipage}%
    }%
    \subfigure{
        \begin{minipage}[c]{0.105\linewidth}
            \centering
            \includegraphics[width=\linewidth]{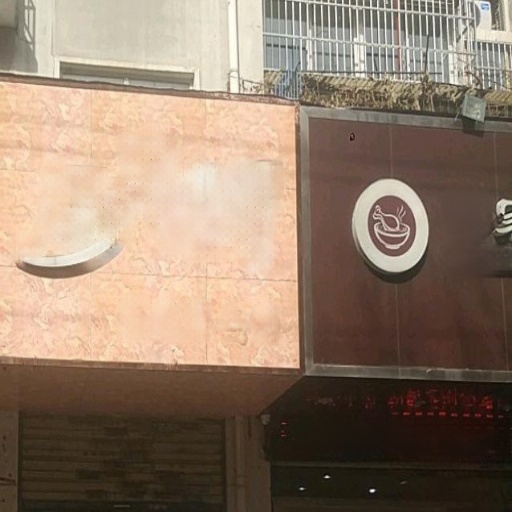}
        \end{minipage}%
    }%
    \subfigure{
        \begin{minipage}[c]{0.105\linewidth}
            \centering
            \includegraphics[width=\linewidth]{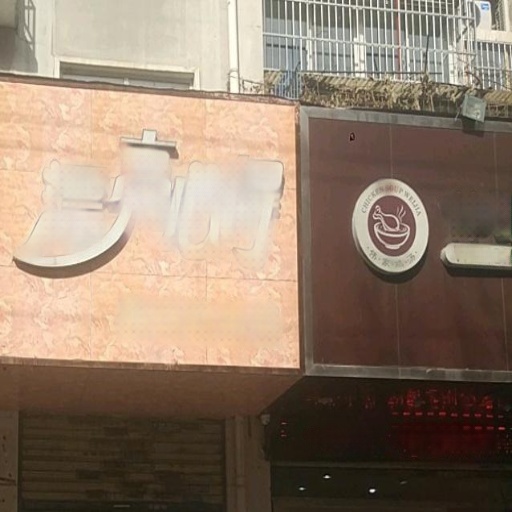}
        \end{minipage}%
    }%
    \subfigure{
        \begin{minipage}[c]{0.105\linewidth}
            \centering
            \includegraphics[width=\linewidth]{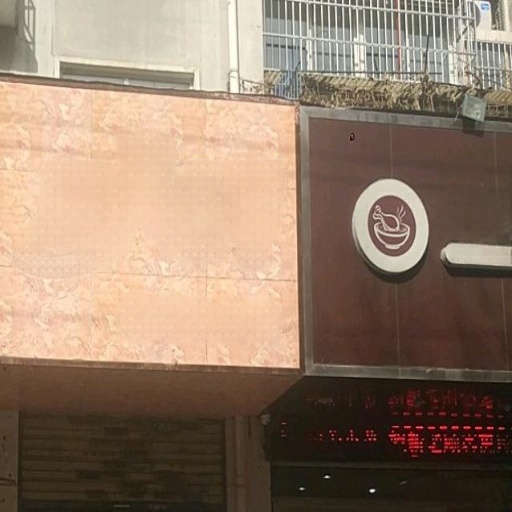}
        \end{minipage}%
    }
    \subfigure{
        \begin{minipage}[c]{0.105\linewidth}
            \centering
            \includegraphics[width=\linewidth]{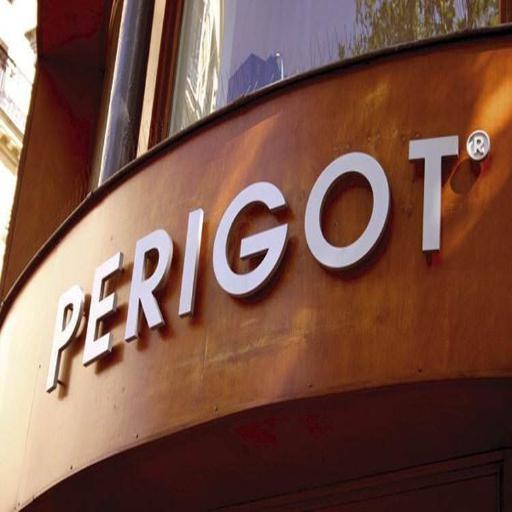}
        \end{minipage}%
    }%
    \subfigure{
        \begin{minipage}[c]{0.105\linewidth}
            \centering
            \includegraphics[width=\linewidth]{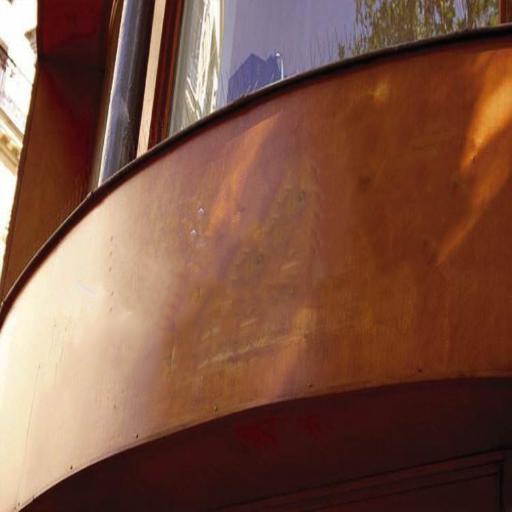}
        \end{minipage}%
    }%
    \subfigure{
        \begin{minipage}[c]{0.105\linewidth}
            \centering
            \includegraphics[width=\linewidth]{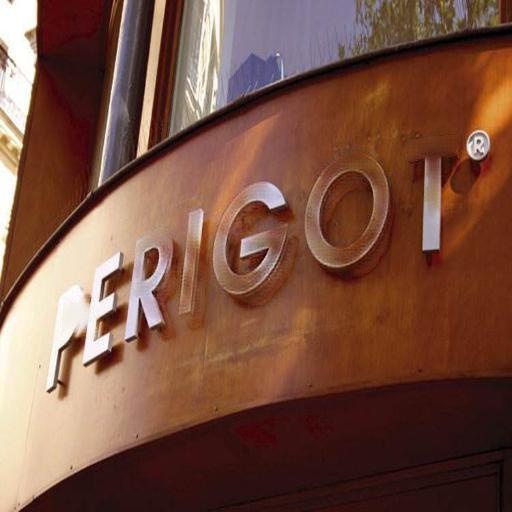}
        \end{minipage}%
    }%
    \subfigure{
        \begin{minipage}[c]{0.105\linewidth}
            \centering
            \includegraphics[width=\linewidth]{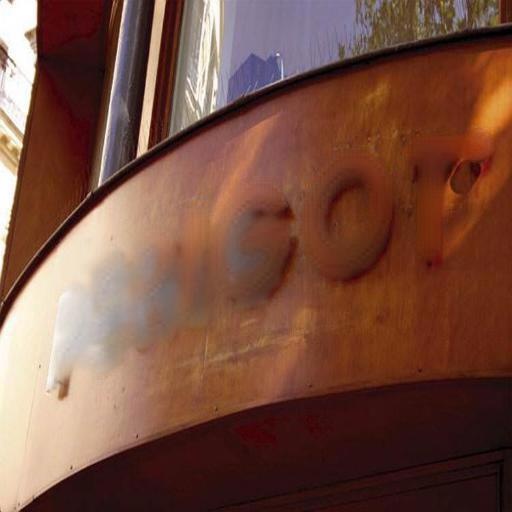}
        \end{minipage}%
    }%
    \subfigure{
        \begin{minipage}[c]{0.105\linewidth}
            \centering
            \includegraphics[width=\linewidth]{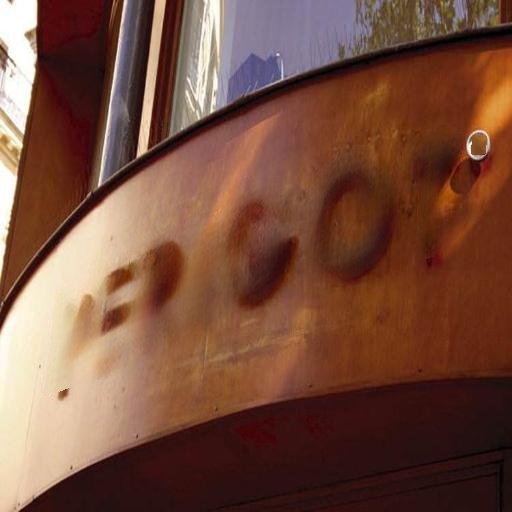}
        \end{minipage}%
    }%
    \subfigure{
        \begin{minipage}[c]{0.105\linewidth}
            \centering
            \includegraphics[width=\linewidth]{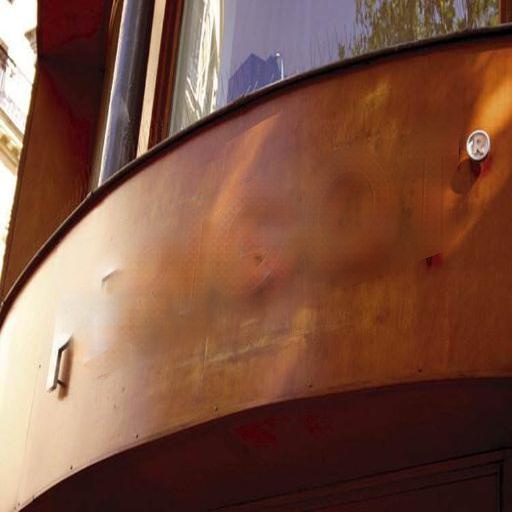}
        \end{minipage}%
    }%
    \subfigure{
        \begin{minipage}[c]{0.105\linewidth}
            \centering
            \includegraphics[width=\linewidth]{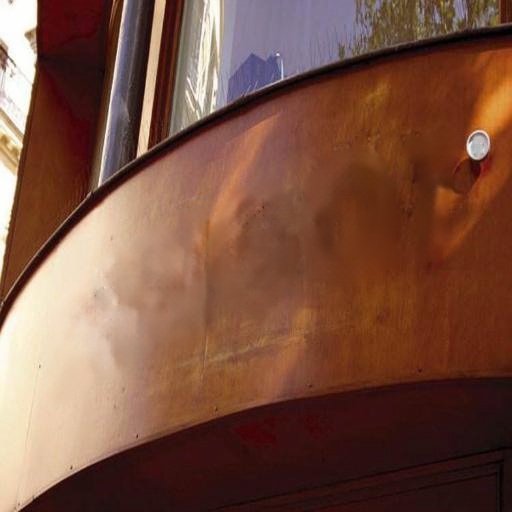}
        \end{minipage}%
    }%
    \subfigure{
        \begin{minipage}[c]{0.105\linewidth}
            \centering
            \includegraphics[width=\linewidth]{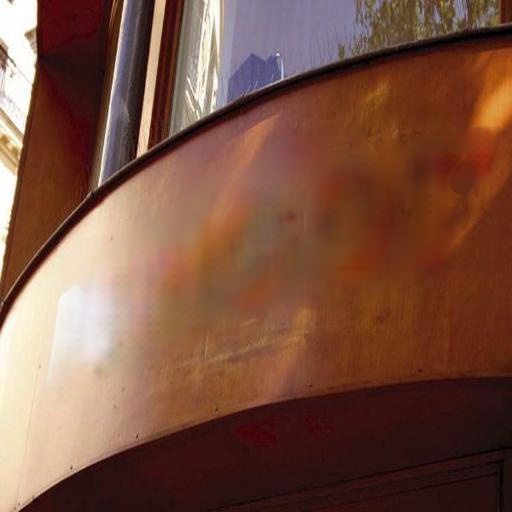}
        \end{minipage}%
    }%
    \subfigure{
        \begin{minipage}[c]{0.105\linewidth}
            \centering
            \includegraphics[width=\linewidth]{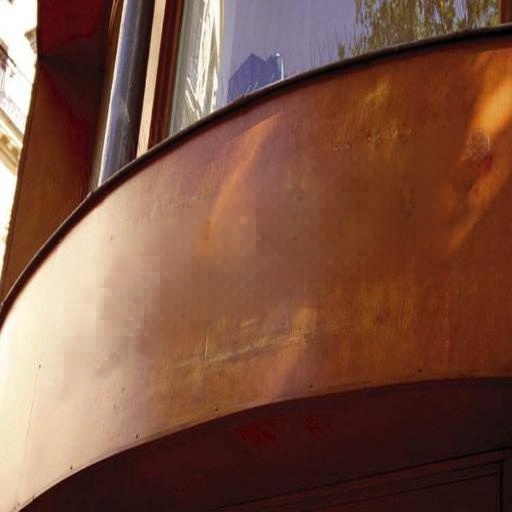}
        \end{minipage}%
    }
    \subfigure{
        \begin{minipage}[c]{0.105\linewidth}
            \centering
            \includegraphics[width=\linewidth]{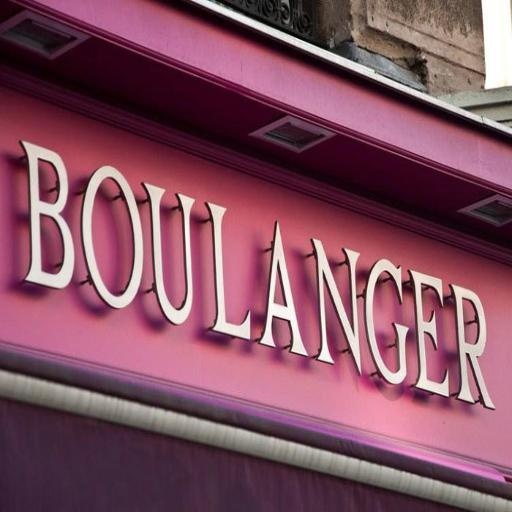}
        \end{minipage}%
    }%
    \subfigure{
        \begin{minipage}[c]{0.105\linewidth}
            \centering
            \includegraphics[width=\linewidth]{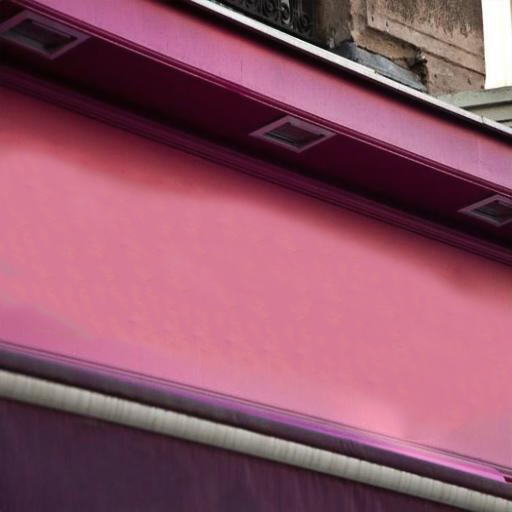}
        \end{minipage}%
    }%
    \subfigure{
        \begin{minipage}[c]{0.105\linewidth}
            \centering
            \includegraphics[width=\linewidth]{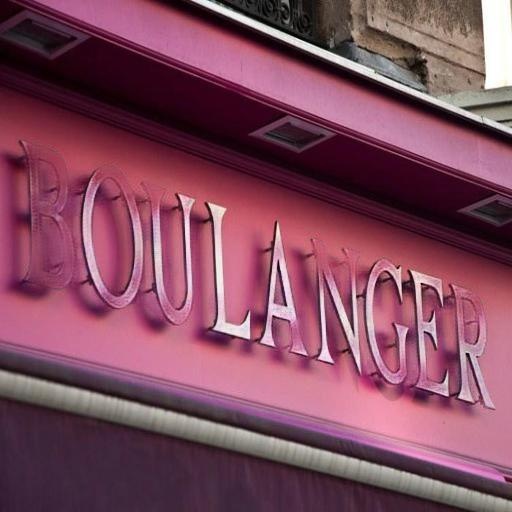}
        \end{minipage}%
    }%
    \subfigure{
        \begin{minipage}[c]{0.105\linewidth}
            \centering
            \includegraphics[width=\linewidth]{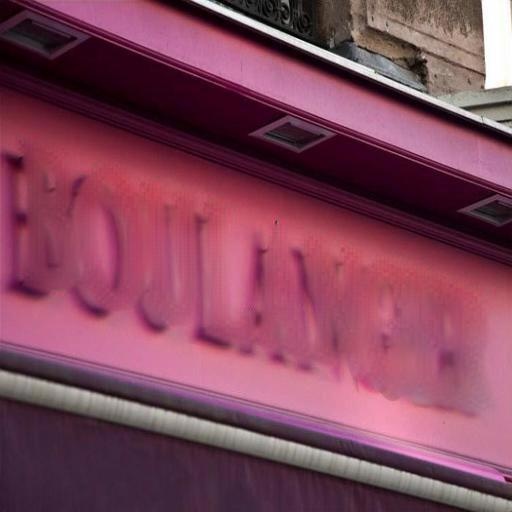}
        \end{minipage}%
    }%
    \subfigure{
        \begin{minipage}[c]{0.105\linewidth}
            \centering
            \includegraphics[width=\linewidth]{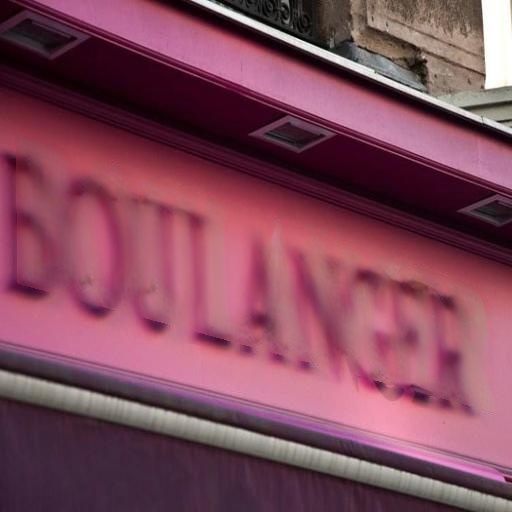}
        \end{minipage}%
    }%
    \subfigure{
        \begin{minipage}[c]{0.105\linewidth}
            \centering
            \includegraphics[width=\linewidth]{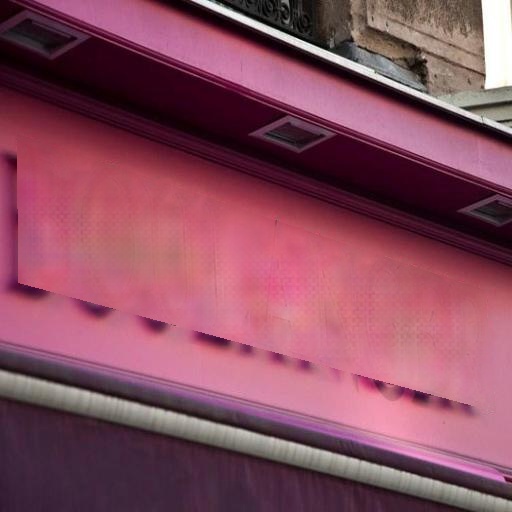}
        \end{minipage}%
    }%
    \subfigure{
        \begin{minipage}[c]{0.105\linewidth}
            \centering
            \includegraphics[width=\linewidth]{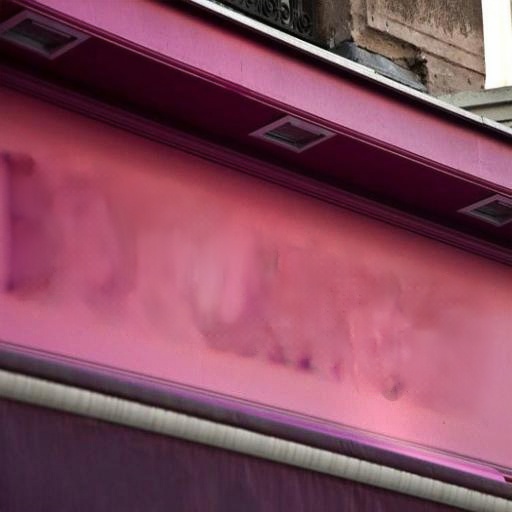}
        \end{minipage}%
    }%
    \subfigure{
        \begin{minipage}[c]{0.105\linewidth}
            \centering
            \includegraphics[width=\linewidth]{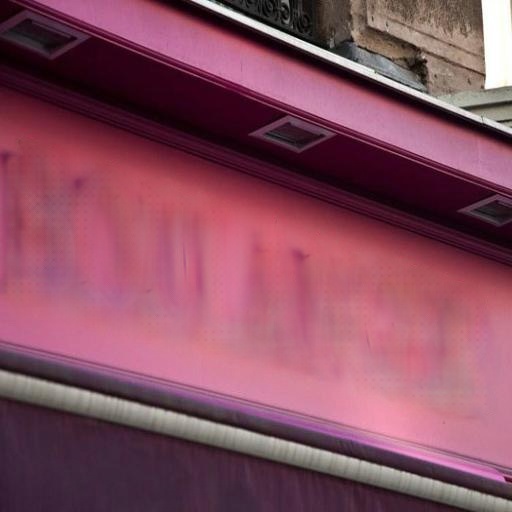}
        \end{minipage}%
    }%
    \subfigure{
        \begin{minipage}[c]{0.105\linewidth}
            \centering
            \includegraphics[width=\linewidth]{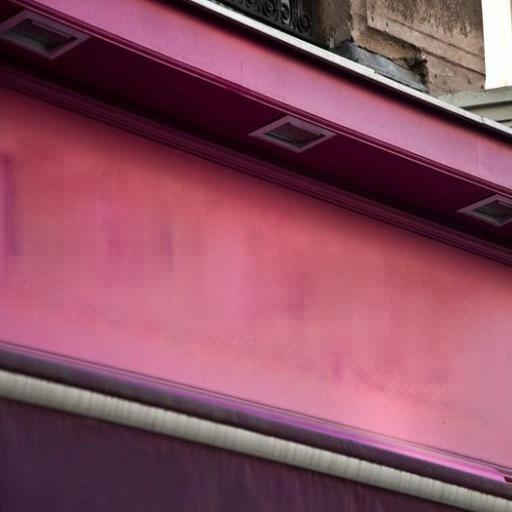}
        \end{minipage}%
    }
    \subfigure{
        \begin{minipage}[c]{0.105\linewidth}
            \centering
            \includegraphics[width=\linewidth]{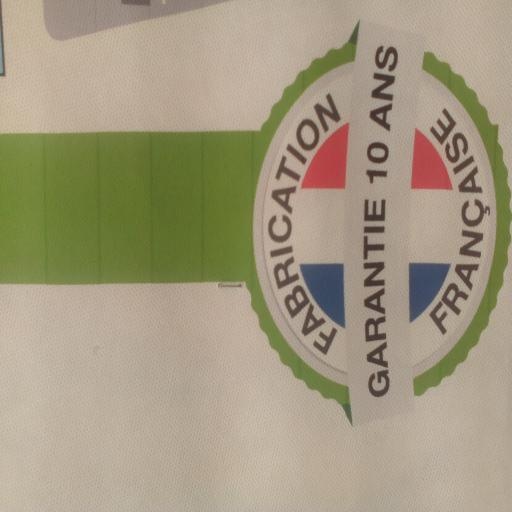}
        \end{minipage}%
    }%
    \subfigure{
        \begin{minipage}[c]{0.105\linewidth}
            \centering
            \includegraphics[width=\linewidth]{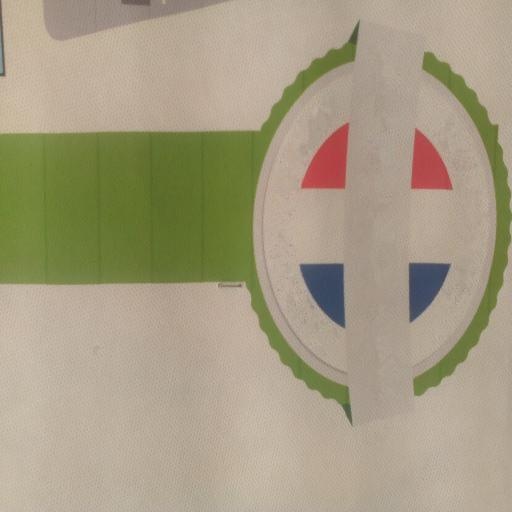}
        \end{minipage}%
    }%
    \subfigure{
        \begin{minipage}[c]{0.105\linewidth}
            \centering
            \includegraphics[width=\linewidth]{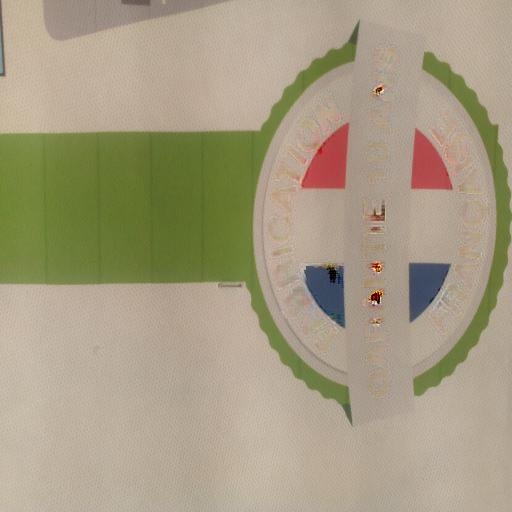}
        \end{minipage}%
    }%
    \subfigure{
        \begin{minipage}[c]{0.105\linewidth}
            \centering
            \includegraphics[width=\linewidth]{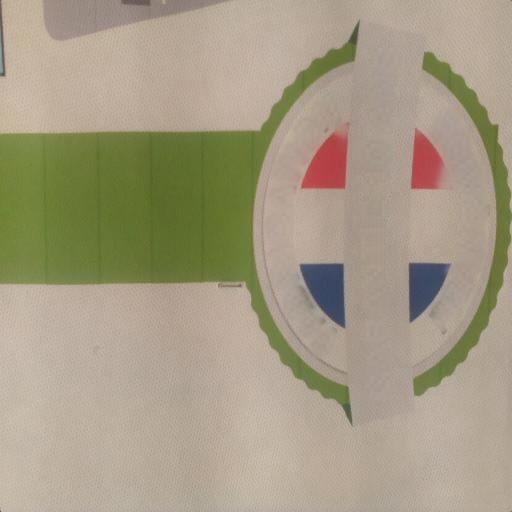}
        \end{minipage}%
    }%
    \subfigure{
        \begin{minipage}[c]{0.105\linewidth}
            \centering
            \includegraphics[width=\linewidth]{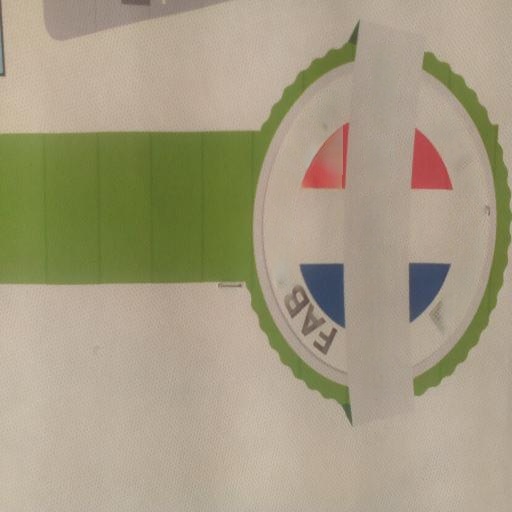}
        \end{minipage}%
    }%
    \subfigure{
        \begin{minipage}[c]{0.105\linewidth}
            \centering
            \includegraphics[width=\linewidth]{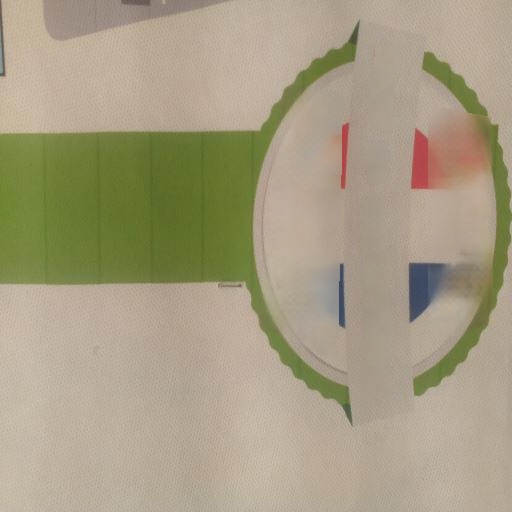}
        \end{minipage}%
    }%
    \subfigure{
        \begin{minipage}[c]{0.105\linewidth}
            \centering
            \includegraphics[width=\linewidth]{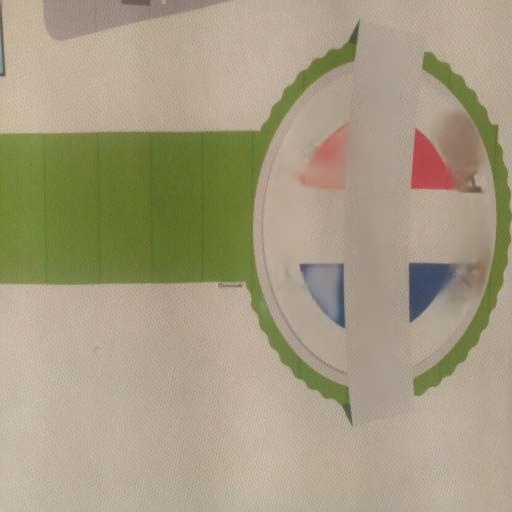}
        \end{minipage}%
    }%
    \subfigure{
        \begin{minipage}[c]{0.105\linewidth}
            \centering
            \includegraphics[width=\linewidth]{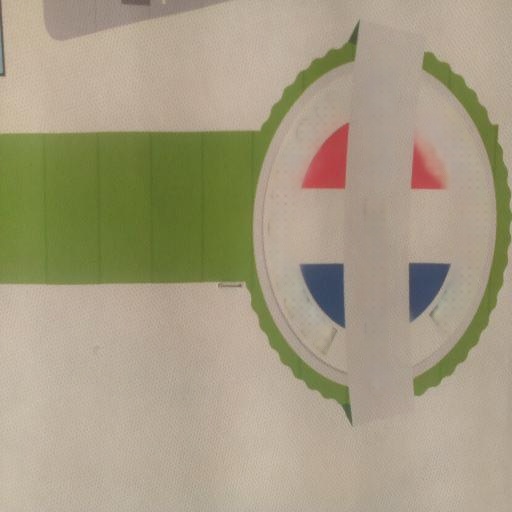}
        \end{minipage}%
    }%
    \subfigure{
        \begin{minipage}[c]{0.105\linewidth}
            \centering
            \includegraphics[width=\linewidth]{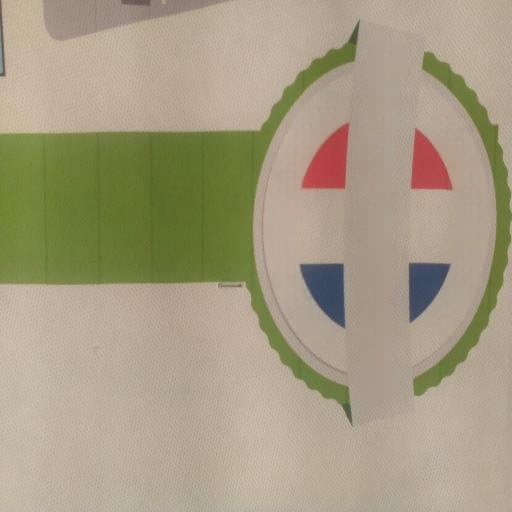}
        \end{minipage}%
    }
    \setcounter{subfigure}{0}
    \subfigure[Input]{
        \begin{minipage}[c]{0.105\linewidth}
            \centering
            \includegraphics[width=\linewidth]{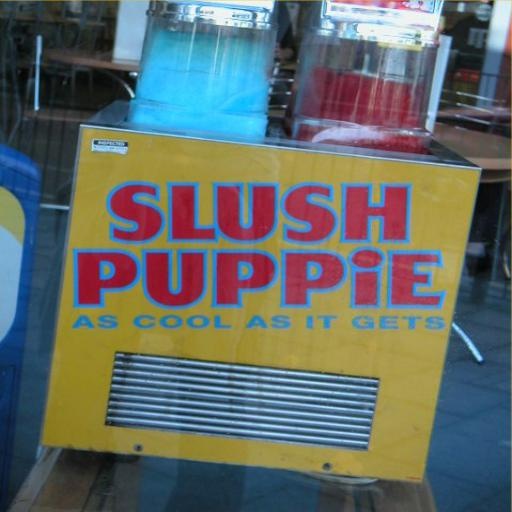}
        \end{minipage}%
    }%
    \subfigure[GT]{
        \begin{minipage}[c]{0.105\linewidth}
            \centering
            \includegraphics[width=\linewidth]{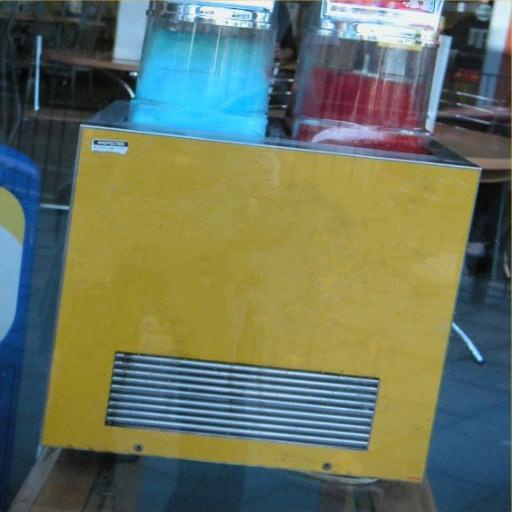}
        \end{minipage}%
    }%
    \subfigure[MTRNet++]{
        \begin{minipage}[c]{0.105\linewidth}
            \centering
            \includegraphics[width=\linewidth]{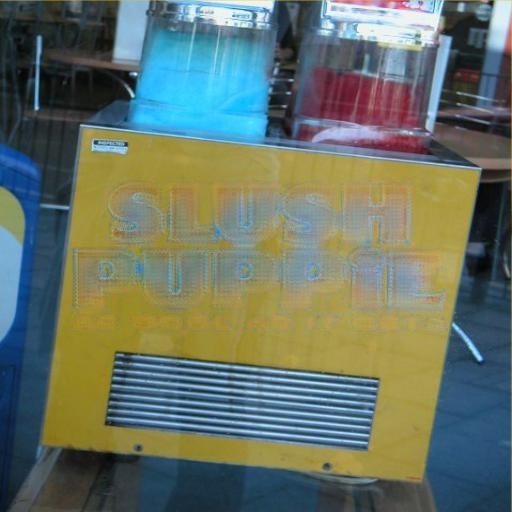}
        \end{minipage}%
    }%
    \subfigure[EraseNet]{
        \begin{minipage}[c]{0.105\linewidth}
            \centering
            \includegraphics[width=\linewidth]{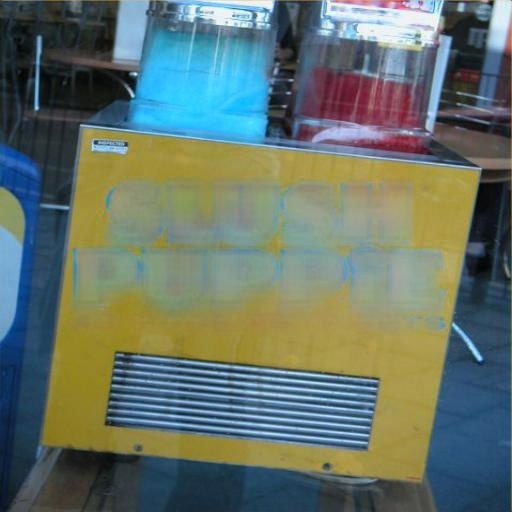}
        \end{minipage}%
    }%
    \subfigure[SSTE]{
        \begin{minipage}[c]{0.105\linewidth}
            \centering
            \includegraphics[width=\linewidth]{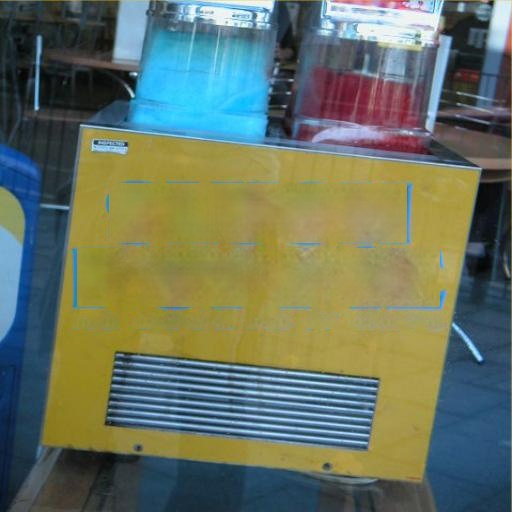}
        \end{minipage}%
    }%
    \subfigure[GaRNet]{
        \begin{minipage}[c]{0.105\linewidth}
            \centering
            \includegraphics[width=\linewidth]{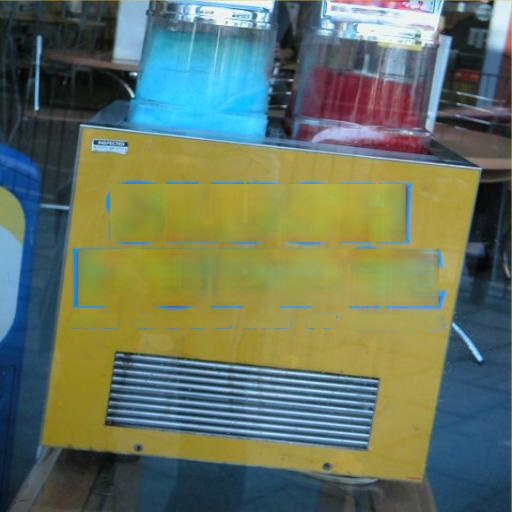}
        \end{minipage}%
    }%
    \subfigure[CTRNet]{
        \begin{minipage}[c]{0.105\linewidth}
            \centering
            \includegraphics[width=\linewidth]{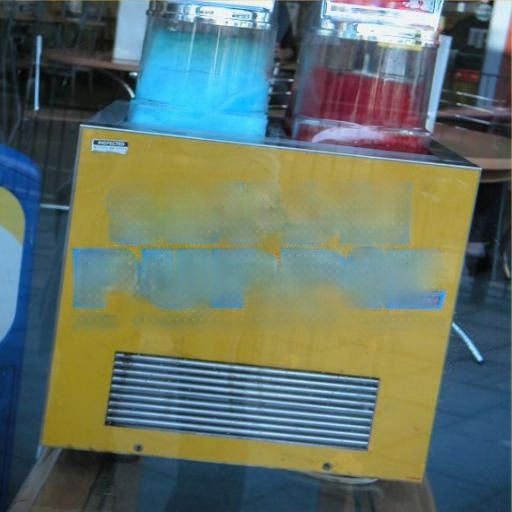}
        \end{minipage}%
    }%
    \subfigure[PERT]{
        \begin{minipage}[c]{0.105\linewidth}
            \centering
            \includegraphics[width=\linewidth]{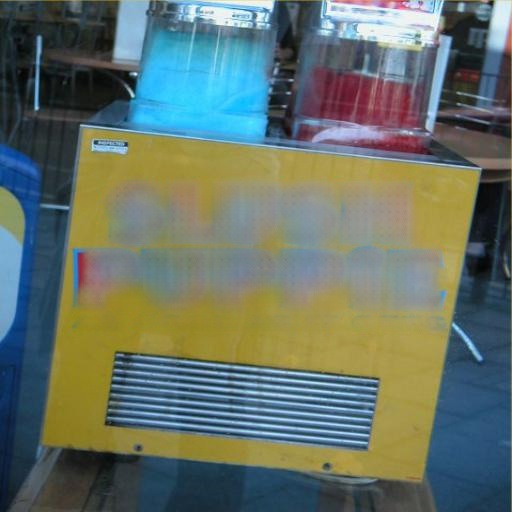}
        \end{minipage}%
    }%
    \subfigure[ViTEraser]{
        \begin{minipage}[c]{0.105\linewidth}
            \centering
            \includegraphics[width=\linewidth]{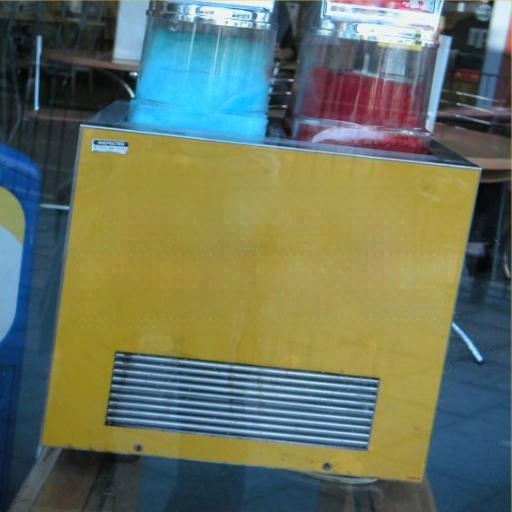}
        \end{minipage}%
    }
    \caption{More visualization results on SCUT-EnsText.
    From (c) to (i), the visualizations are obtained by MTRNet++ \cite{tursun2020mtrnet++}, EraseNet \cite{liu2020erasenet}, SSTE \cite{tang2021stroke}, GaRNet \cite{lee2022surprisingly}, CTRNet \cite{liu2022don}, PERT \cite{wang2023pert}, and ViTEraser-Swinv2-Small (with SegMIM), respectively. 
    Zoom in for a better view. (GT: Ground Truth)}
    \label{fig:vis_more_scutens}
\end{figure*}

We provide more visualization results on SCUT-EnsText \cite{liu2020erasenet} in Fig.~\ref{fig:vis_more_scutens}.
It can be seen that the proposed ViTEraser excels in handling large texts (1\textit{st} row), tiny texts (2\textit{nd} row), arbitrary-shaped texts (3\textit{rd} to 6\textit{th}), complex fonts (7\textit{th} row), 3D texts (8\textit{th} and 9\textit{th} rows).
Moreover, ViTEraser can generate more visually plausible backgrounds as demonstrated by the 10\textit{th} and 11\textit{th} rows.

\subsubsection{SCUT-Syn}

\begin{figure*}[t]
    \centering
    \subfigtopskip=0pt 
    \subfigbottomskip=2pt 
    \subfigcapskip=1pt 
    \subfigure{
        \begin{minipage}[c]{0.15\linewidth}
            \centering
            \includegraphics[width=\linewidth]{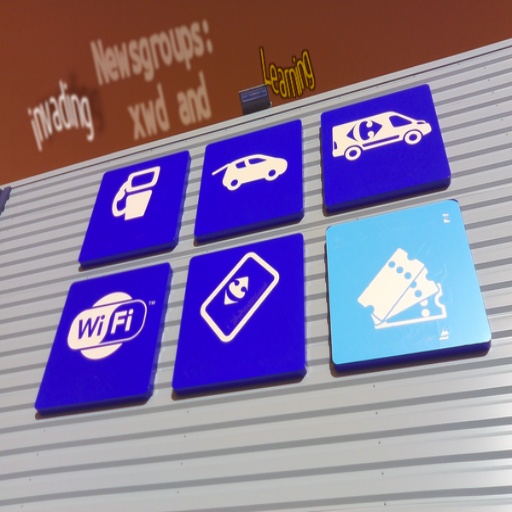}
        \end{minipage}%
    }%
    \subfigure{
        \begin{minipage}[c]{0.15\linewidth}
            \centering
            \includegraphics[width=\linewidth]{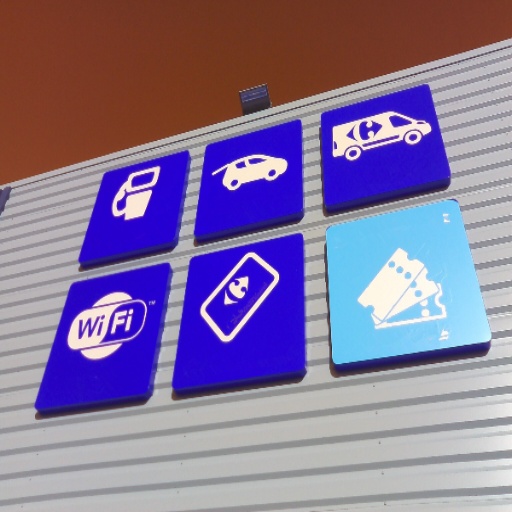}
        \end{minipage}%
    }%
    \subfigure{
        \begin{minipage}[c]{0.15\linewidth}
            \centering
            \includegraphics[width=\linewidth]{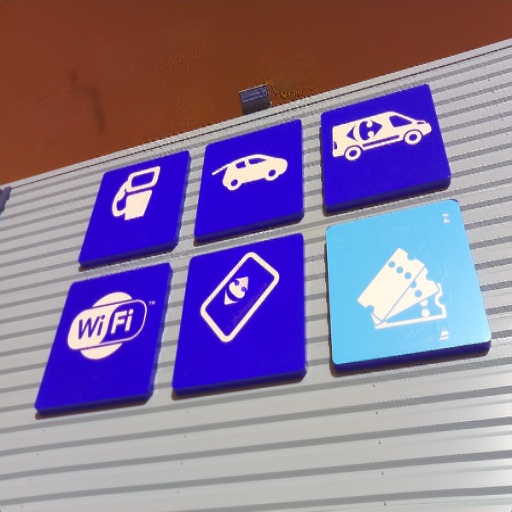}
        \end{minipage}%
    }%
    \subfigure{
        \begin{minipage}[c]{0.15\linewidth}
            \centering
            \includegraphics[width=\linewidth]{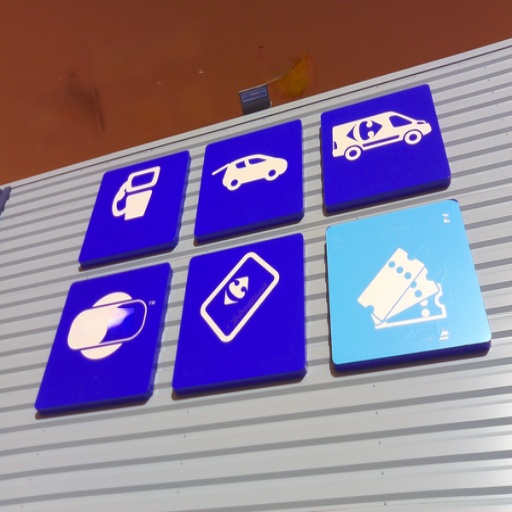}
        \end{minipage}%
    }%
    \subfigure{
        \begin{minipage}[c]{0.15\linewidth}
            \centering
            \includegraphics[width=\linewidth]{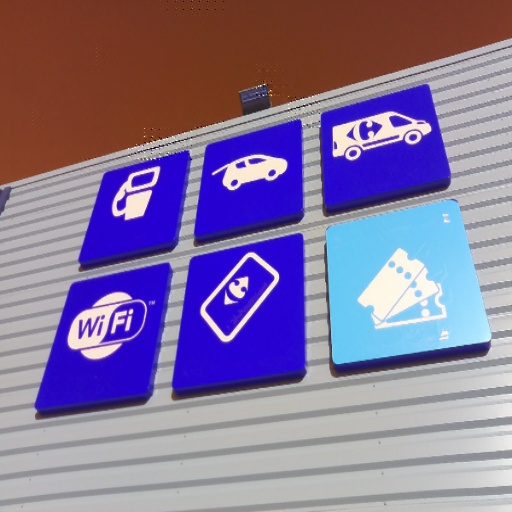}
        \end{minipage}%
    }
    \subfigure{
        \begin{minipage}[c]{0.15\linewidth}
            \centering
            \includegraphics[width=\linewidth]{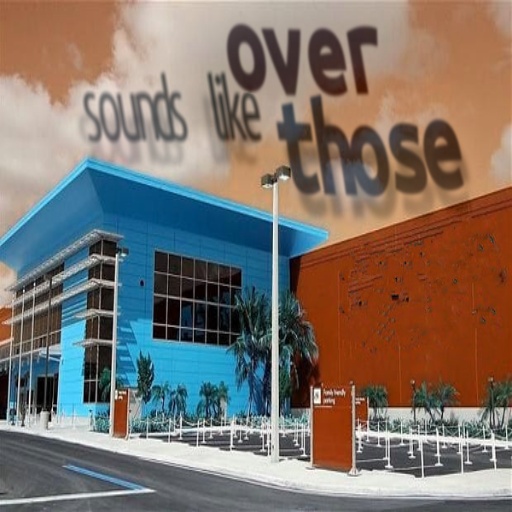}
        \end{minipage}%
    }%
    \subfigure{
        \begin{minipage}[c]{0.15\linewidth}
            \centering
            \includegraphics[width=\linewidth]{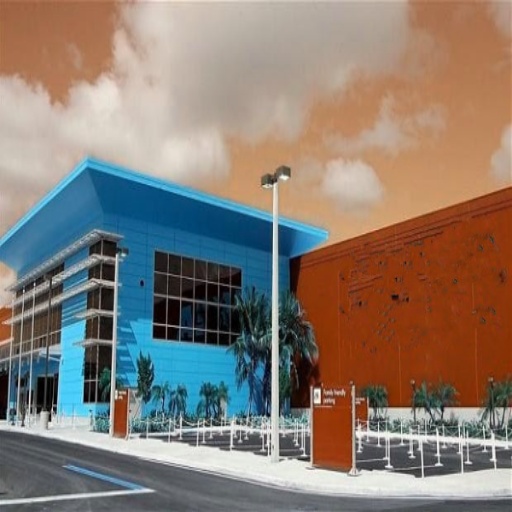}
        \end{minipage}%
    }%
    \subfigure{
        \begin{minipage}[c]{0.15\linewidth}
            \centering
            \includegraphics[width=\linewidth]{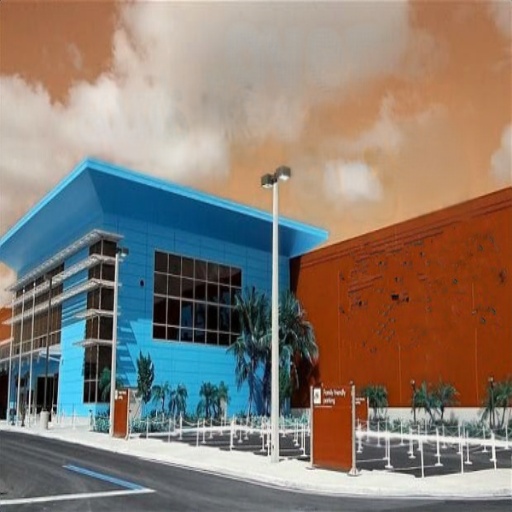}
        \end{minipage}%
    }%
    \subfigure{
        \begin{minipage}[c]{0.15\linewidth}
            \centering
            \includegraphics[width=\linewidth]{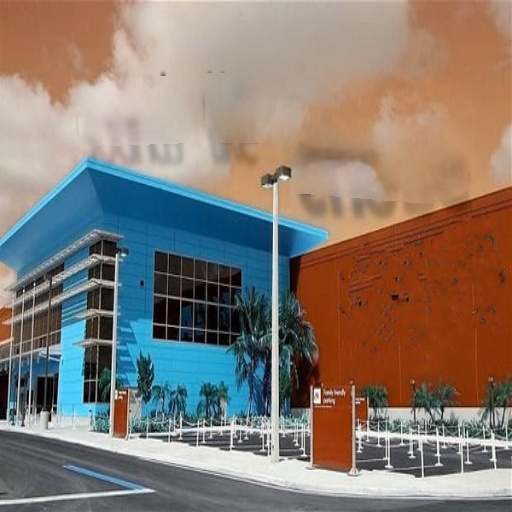}
        \end{minipage}%
    }%
    \subfigure{
        \begin{minipage}[c]{0.15\linewidth}
            \centering
            \includegraphics[width=\linewidth]{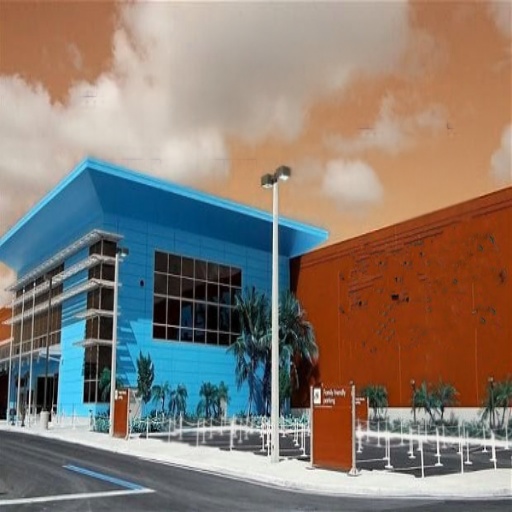}
        \end{minipage}%
    }
    \subfigure{
        \begin{minipage}[c]{0.15\linewidth}
            \centering
            \includegraphics[width=\linewidth]{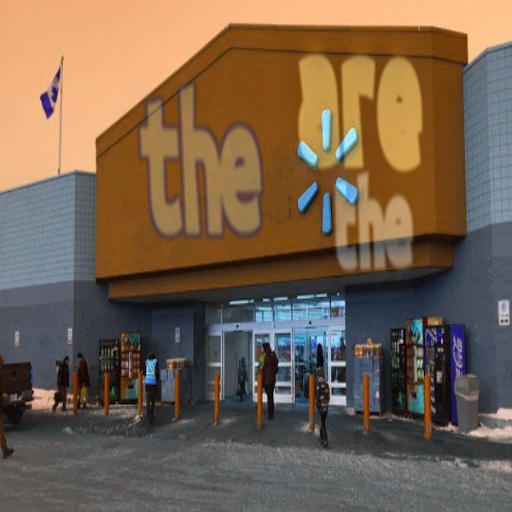}
        \end{minipage}%
    }%
    \subfigure{
        \begin{minipage}[c]{0.15\linewidth}
            \centering
            \includegraphics[width=\linewidth]{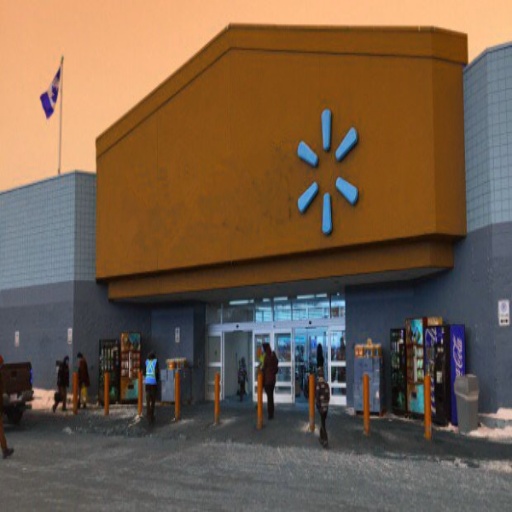}
        \end{minipage}%
    }%
    \subfigure{
        \begin{minipage}[c]{0.15\linewidth}
            \centering
            \includegraphics[width=\linewidth]{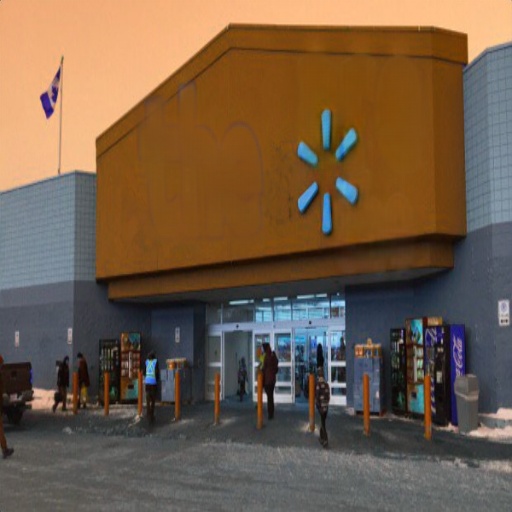}
        \end{minipage}%
    }%
    \subfigure{
        \begin{minipage}[c]{0.15\linewidth}
            \centering
            \includegraphics[width=\linewidth]{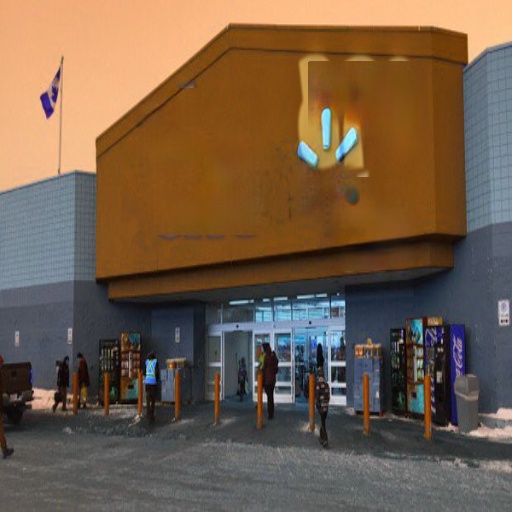}
        \end{minipage}%
    }%
    \subfigure{
        \begin{minipage}[c]{0.15\linewidth}
            \centering
            \includegraphics[width=\linewidth]{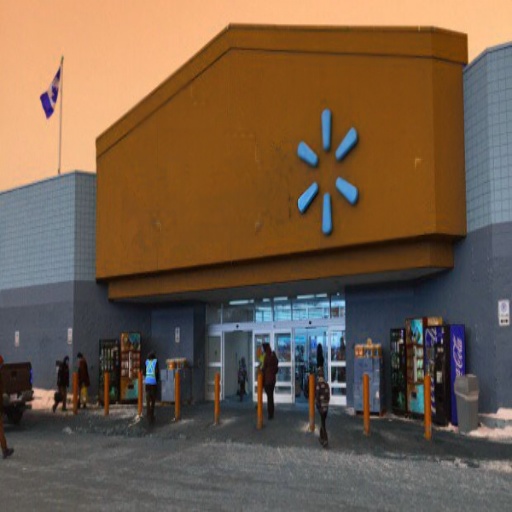}
        \end{minipage}%
    }
    \subfigure{
        \begin{minipage}[c]{0.15\linewidth}
            \centering
            \includegraphics[width=\linewidth]{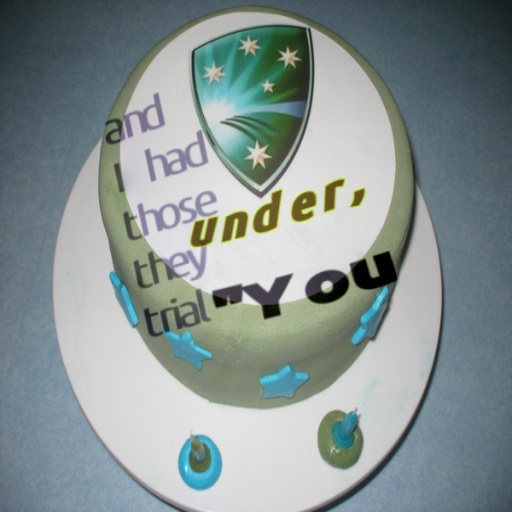}
        \end{minipage}%
    }%
    \subfigure{
        \begin{minipage}[c]{0.15\linewidth}
            \centering
            \includegraphics[width=\linewidth]{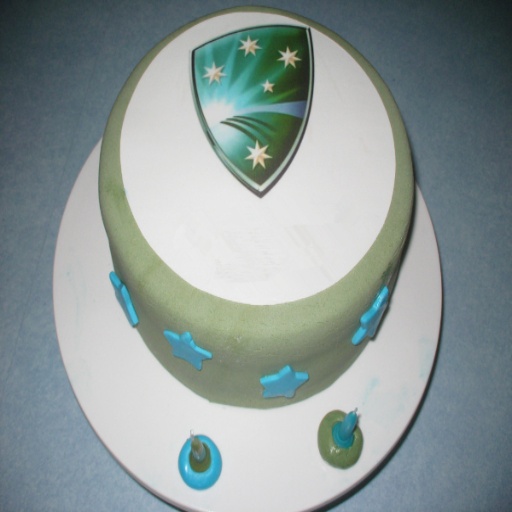}
        \end{minipage}%
    }%
    \subfigure{
        \begin{minipage}[c]{0.15\linewidth}
            \centering
            \includegraphics[width=\linewidth]{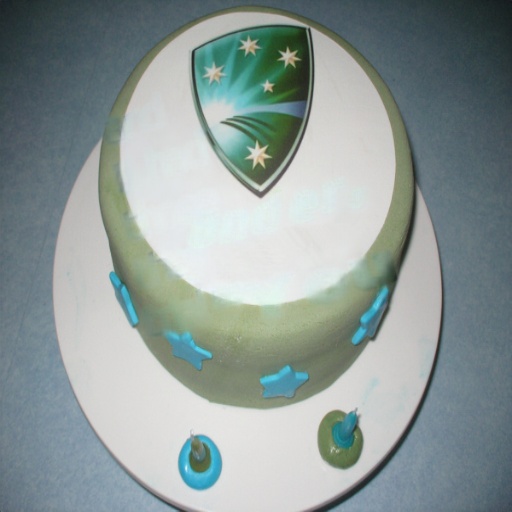}
        \end{minipage}%
    }%
    \subfigure{
        \begin{minipage}[c]{0.15\linewidth}
            \centering
            \includegraphics[width=\linewidth]{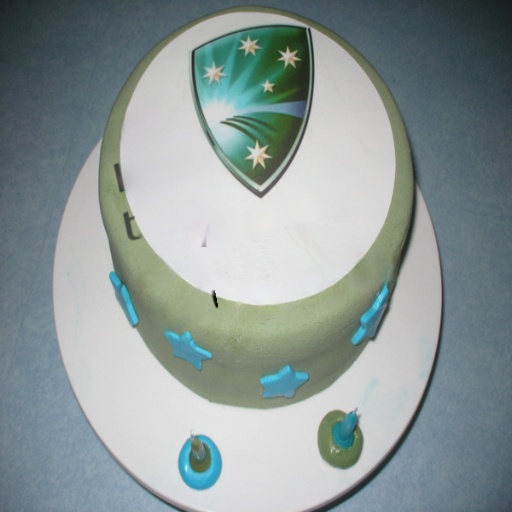}
        \end{minipage}%
    }%
    \subfigure{
        \begin{minipage}[c]{0.15\linewidth}
            \centering
            \includegraphics[width=\linewidth]{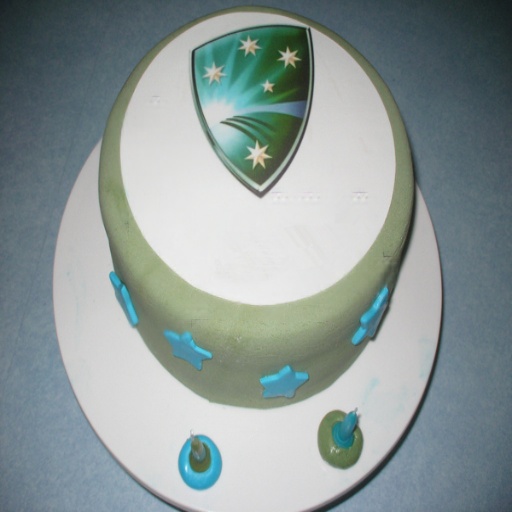}
        \end{minipage}%
    }
    \subfigure{
        \begin{minipage}[c]{0.15\linewidth}
            \centering
            \includegraphics[width=\linewidth]{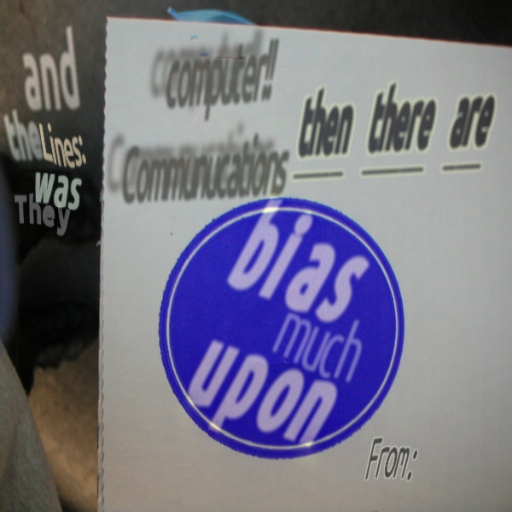}
        \end{minipage}%
    }%
    \subfigure{
        \begin{minipage}[c]{0.15\linewidth}
            \centering
            \includegraphics[width=\linewidth]{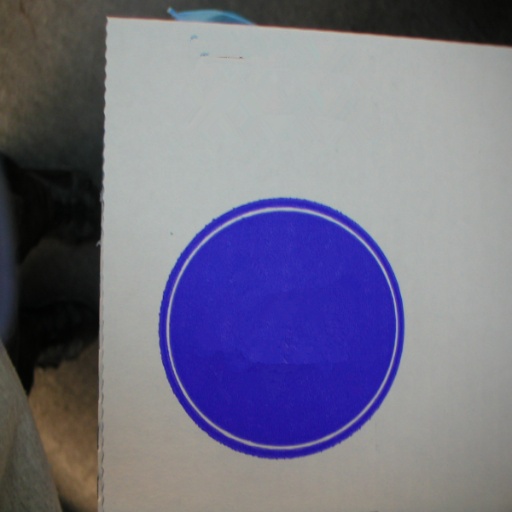}
        \end{minipage}%
    }%
    \subfigure{
        \begin{minipage}[c]{0.15\linewidth}
            \centering
            \includegraphics[width=\linewidth]{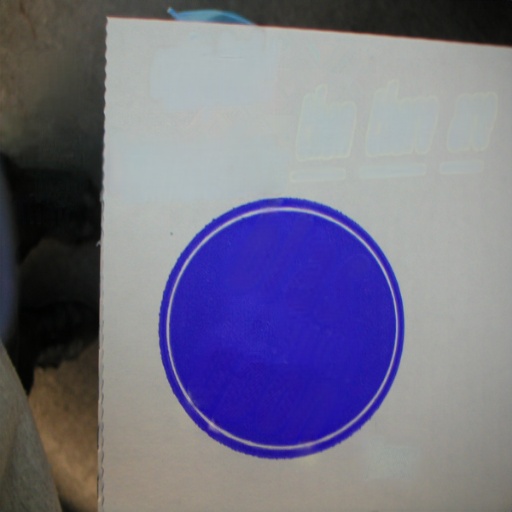}
        \end{minipage}%
    }%
    \subfigure{
        \begin{minipage}[c]{0.15\linewidth}
            \centering
            \includegraphics[width=\linewidth]{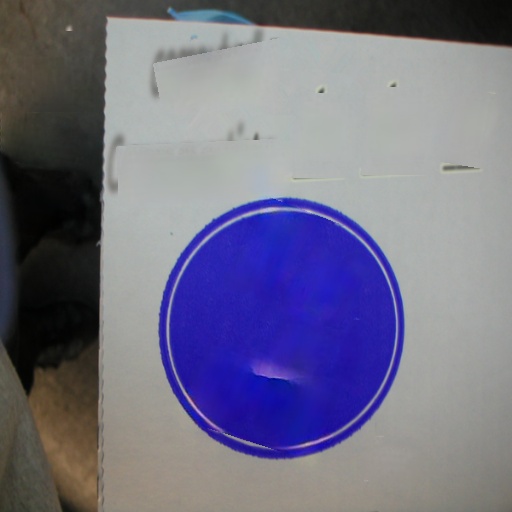}
        \end{minipage}%
    }%
    \subfigure{
        \begin{minipage}[c]{0.15\linewidth}
            \centering
            \includegraphics[width=\linewidth]{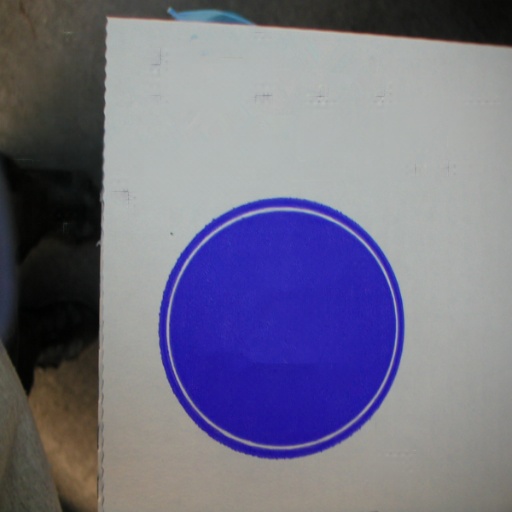}
        \end{minipage}%
    }
    \setcounter{subfigure}{0}
    \subfigure[Input]{
        \begin{minipage}[c]{0.15\linewidth}
            \centering
            \includegraphics[width=\linewidth]{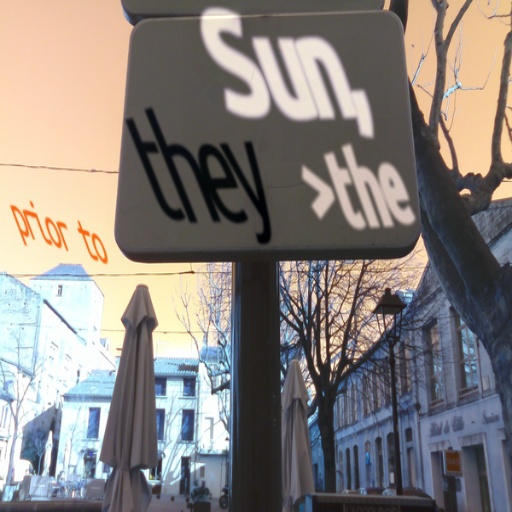}
        \end{minipage}%
    }%
    \subfigure[GT]{
        \begin{minipage}[c]{0.15\linewidth}
            \centering
            \includegraphics[width=\linewidth]{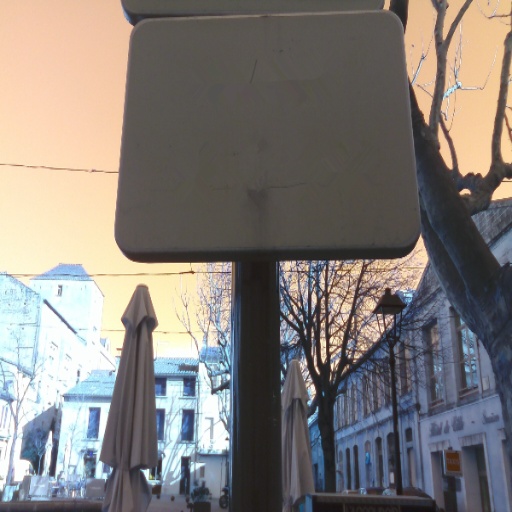}
        \end{minipage}%
    }%
    \subfigure[EraseNet]{
        \begin{minipage}[c]{0.15\linewidth}
            \centering
            \includegraphics[width=\linewidth]{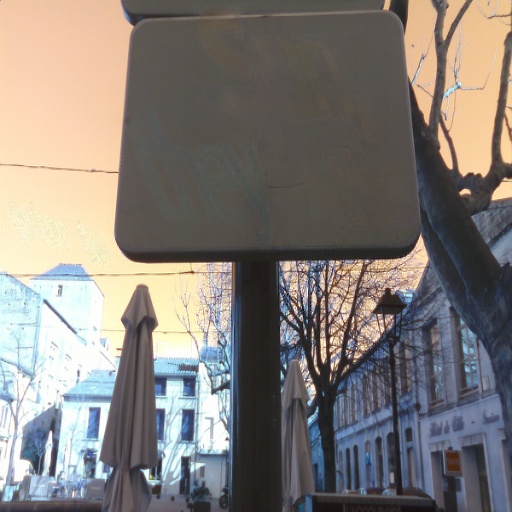}
        \end{minipage}%
    }%
    \subfigure[SSTE]{
        \begin{minipage}[c]{0.15\linewidth}
            \centering
            \includegraphics[width=\linewidth]{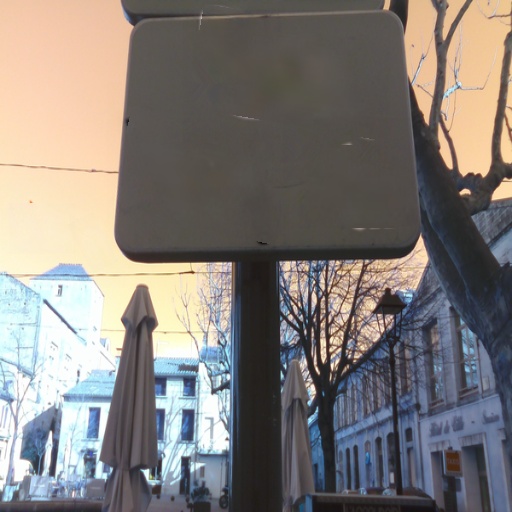}
        \end{minipage}%
    }%
    \subfigure[ViTEraser]{
        \begin{minipage}[c]{0.15\linewidth}
            \centering
            \includegraphics[width=\linewidth]{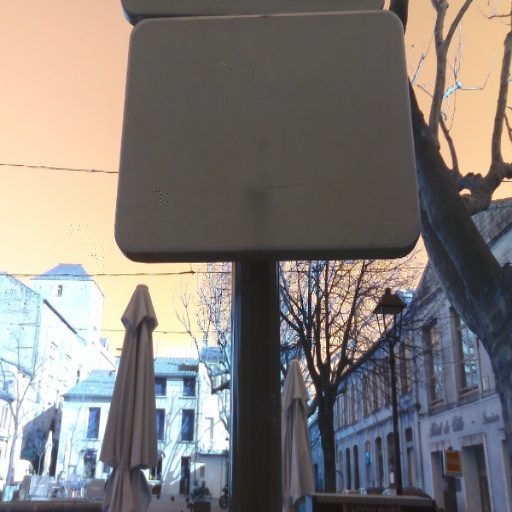}
        \end{minipage}%
    }
    \caption{More visualization results on SCUT-Syn.
    From (c) to (e), the visualizations are obtained by EraseNet \cite{liu2020erasenet}, SSTE \cite{tang2021stroke}, and ViTEraser-Swinv2-Base (with SegMIM), respectively.
    Zoom in for a better view.}
    \label{fig:vis_more_scutsyn}
\end{figure*}

We provide more visualization results on SCUT-Syn \cite{zhang2019ensnet} in Fig.~\ref{fig:vis_more_scutsyn}, demonstrating that the proposed ViTEraser can exhaustively erase the texts without remnant traces.

\end{document}